%% file: main.tex
\documentclass{article}
\usepackage{packages}

\icmltitlerunning{\bench}

\begin{document}

\twocolumn[\icmltitle{\bench: Evaluating AI Agents across \\ Diverse Real-World IT Automation Tasks}

\icmlsetsymbol{equal}{*}
\begin{icmlauthorlist}
\icmlauthor{Saurabh Jha}{equal,ibm}
\icmlauthor{Rohan Arora}{equal,ibm}
\icmlauthor{Yuji Watanabe}{equal,ibm}
\icmlauthor{Takumi Yanagawa}{ibm}
\icmlauthor{Yinfang Chen}{uiuc}
\icmlauthor{Jackson Clark}{uiuc}
\icmlauthor{Bhavya Bhavya}{ibm}
\icmlauthor{Mudit Verma}{ibm}
\icmlauthor{Harshit Kumar}{ibm}
\icmlauthor{Hirokuni Kitahara}{ibm}
\icmlauthor{Noah Zheutlin}{ibm}
\icmlauthor{Saki Takano}{ibm}
\icmlauthor{Divya Pathak}{ibm}
\icmlauthor{Felix George}{ibm}
\icmlauthor{Xinbo Wu}{uiuc}
\icmlauthor{Bekir O Turkkan}{ibm}
\icmlauthor{Gerard Vanloo}{ibm}
\icmlauthor{Michael Nidd}{ibm}
\icmlauthor{Ting Dai}{ibm}
\icmlauthor{Oishik Chatterjee}{ibm}
\icmlauthor{Pranjal Gupta}{ibm}
\icmlauthor{Suranjana Samanta}{ibm}
\icmlauthor{Pooja Aggarwal}{ibm}
\icmlauthor{Rong Lee}{ibm}
\icmlauthor{
Pavankumar Murali}{ibm}
\icmlauthor{Jae-wook Ahn}{ibm}
\icmlauthor{Debanjana Kar}{ibm}
\icmlauthor{Ameet Rahane}{ibm}
\icmlauthor{Carlos Fonseca}{ibm}
\icmlauthor{Amit Paradkar}{ibm}
\icmlauthor{Yu Deng}{ibm}
\icmlauthor{Pratibha Moogi}{ibm}
\icmlauthor{Prateeti Mohapatra}{ibm}
\icmlauthor{Naoki Abe}{ibm}
\icmlauthor{Chandrasekhar Narayanaswami}{ibm}
\icmlauthor{Tianyin Xu}{uiuc}
\icmlauthor{Lav R. Varshney}{uiuc}
\icmlauthor{Ruchi Mahindru}{ibm}
\icmlauthor{Anca Sailer}{ibm}
\icmlauthor{Laura Shwartz}{ibm}
\icmlauthor{Daby Sow}{ibm}
\icmlauthor{Nicholas C. M. Fuller}{ibm}
\icmlauthor{Ruchir Puri}{ibm}
\end{icmlauthorlist}
\icmlaffiliation{ibm}{IBM}
\icmlaffiliation{uiuc}{University of Illinois at Urbana-Champaign}

\icmlcorrespondingauthor{Saurabh Jha}{Saurabh.Jha@ibm.com}
\icmlkeywords{Machine Learning, ICML}

\vskip 0.3in
]

\printAffiliationsAndNotice{\icmlEqualContribution} %
\input{sections/00_abstract}

\input{sections/01_intro}

\input{sections/02-new-related-work}

\input{sections/03_bench}

\input{sections/04_experiments}

\input{sections/05_evaluation}

\input{sections/07_conclusion}

\input{sections/0x_statements}

\input{sections/0x_acknowledge}

\bibliography{main}
\bibliographystyle{abbrvnat}

\newpage
\appendix
\input{appx/00_table_of_contents}
\input{appx/01_related_work}
\input{appx/02_bench_details}

\input{appx/usecases/SRE/main}

\input{appx/usecases/compliance/main}
\input{appx/usecases/finops/main}

\end{document}

%% file: sections/00_abstract.tex
\begin{abstract}

Realizing the vision of using AI agents to automate critical IT tasks depends on the ability to measure and understand effectiveness of proposed solutions. %
We introduce \bench, a framework that offers a systematic methodology for benchmarking AI agents to address real-world IT automation tasks. %
Our initial release targets three key areas: Site Reliability Engineering (SRE),
Compliance and Security Operations (CISO),
and Financial Operations (FinOps). 
The design %
enables AI researchers to understand the challenges and opportunities of AI agents 
for IT automation with push-button workflows and interpretable metrics. 
\bench includes an initial set of 94 real-world scenarios, which can be easily extended by community contributions. Our results show that agents powered by state-of-the-art models resolve only 13.8\% of SRE scenarios, 25.2\% of CISO scenarios, and 0\% of FinOps scenarios. 
We expect \bench to be a key enabler of AI-driven IT automation that is correct, safe, and fast.

\end{abstract}

%% file: sections/01_intro.tex
\section{Introduction}
\label{sec:intro}
Modern IT systems are driving many facets of our economy. They have grown significantly in complexity with the adoption of cloud computing and agile development practices \cite{ITComplexity2022hbr, finops2024}. Effective management of these systems is becoming extremely challenging as corporations struggle to keep up with this growing complexity. Various IT personas ranging from Chief Information Officers to Site Reliability Engineers, Security and Compliance officers and IT engineers in general are struggling to ensure resiliency, reliability, security and cost effective operations of IT Systems.

The recent CrowdStrike outage highlighted these challenges as it brought down our society's most critical systems from hospital services to air travel and was estimated to cost US Fortune 500 companies a staggering \$5.4 billion \cite{csincident2024}. This incident underlined the critical need for intelligent IT incident resolution, compliance and risk management capabilities, a topic also addressed in the Digital Operational Resiliency Act (DORA) in Europe \cite{dora}. 

\begin{figure*}[ht]
    \centering
    \includegraphics[width=0.9\linewidth]{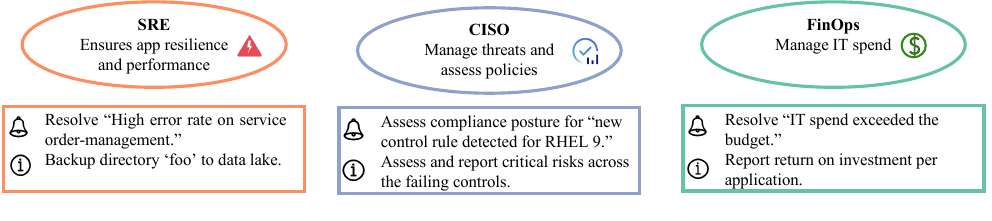}
    \caption{Sample personas and IT tasks. Bell icon represents event-triggered tasks. Information icon represents other tasks such as data analysis, preventive maintenance tasks, or continuous optimization.}
    \label{fig:motivation}
\end{figure*}

The rising popularity of AI agents and their projected ability to handle intricate tasks have increased the demand for AI agents managing IT systems \cite{WIRE19-2024, IDC2024, pujar2023automatedcodegenerationinformation}. Given the complexity of IT tasks, a major hurdle for this research is establishing systematic methods to assess the effectiveness of AI agents prior to their production deployment \cite{Bogin2024SUPEREA, kapoor2024aiagentsmatter}. Consequently, there is an urgency to develop methods for evaluation of AI agents based on real IT tasks and their corresponding environments. %

This paper addresses this critical need and presents \bench, a first of its kind framework that is both comprehensive and visionary for benchmarking real-life IT automation tasks. The goal of \bench is to measure the performance of AI agents across a wide variety of complex and real-life IT tasks across personas including, \textit{Site Reliability Engineering (SRE)} focusing on availability and resiliency, \textit{Compliance and Security Operations (CISO)} ensuring compliance and security of IT implementations, and \textit{Financial Operations (FinOps)} enforcing cost efficiencies and optimizing return on investment, among others (as shown in \cref{fig:motivation}).

\bench aims to advance innovation and establish new standards in the field. Our contributions can be summarized along the following three axes:

\begin{itemize}[left=0pt, topsep=0pt, partopsep=0pt, itemsep=0pt, parsep=0pt]
\item \textbf{Reflecting the real world:} \bench  addresses the IT automation requirements that are relevant and prevalent in production settings. SRE scenarios are based on real-world incidents observed in our own SaaS products, CISO’s are on CIS benchmark (for Internet Security, CIS). FinOps scenarios are identified by the FinOps Foundation \cite{finopsbench} through key business outcomes.

\item \textbf{Being open and extensible with comprehensive IT coverage:} 
We view \bench as a central hub for benchmarking AI-driven solutions across diverse IT automation use cases. To support this, we provide IT benchmark suites and a framework for vertical (i.e., adding more scenarios) and horizontal expansion (i.e., adding more personas), ensuring extensive coverage of IT tasks. 
\bench is an open-source framework built with open-source technologies, while allowing organizations with proprietary technologies to use it for developing and benchmarking their solutions.

\item \textbf{Enabling automated evaluation with partial scoring:} 
\bench is designed to provide constructive feedback to drive improvements in the design of agentic solutions for IT problems. It includes a comprehensive evaluation framework and leaderboard that provide feedback to users at various stages of their agents' reasoning process. %

\end{itemize}

\bench provides push-button 
deployment and tooling for setting up environment, runtime agent, guardrail engine, as well as authorization and authentication. It allows developers and researchers to build novel solutions for managing complex IT systems. 
Currently, \bench addresses reactive problems including incidents diagnosis and resolution, 
compliance assessments in regulated environments for new controls, and cost management events. 
In future, we plan to expand on benchmark evaluation capabilities and include new benchmarks for additional IT processes. 
Currently, \bench comprises of an initial set of 94 scenarios spanning across SRE (42), CISO (50), and FinOps (2), with respective successful scenario handling rate of 13.8\%, 25.2\%, and 0\% (refer to \Cref{sec:results}).

We believe that, similar to the highly influential SWEBench \cite{jimenez2024swebench}, our new \bench framework—which encapsulates and measures the ability of AI agents to automate complex, real-world IT tasks—will spur a comparable acceleration in the performance of real-world IT AI agents.

%% file: sections/02-new-related-work.tex
\section{Related Work}
\label{sec:background-rl}

\begin{table*}
    \footnotesize
    \centering
    \begin{threeparttable}
        \caption{Comparison of \bench\ with related benchmarks }
        \label{tab:aiops_benchmarks}
        \begin{tabular}{%
            l                 %
            c                 %
            m{0.21\textwidth} %
            c                 %
            m{0.11\textwidth} %
            l                 %
            l                 %
        }
            \toprule
            \textbf{Benchmark} & 
            \textbf{\#Scenarios} & 
            \textbf{Personas and Tasks} & 
            \textbf{Resolvable} & 
            \textbf{Automated Evaluation} & 
            \textbf{Environment} & 
            \textbf{Leaderboard} \\
            \midrule

            \bench (ours)
            & 94 
            & \begin{tabular}[l]{@{}l@{}}
            SRE -- Incident Resolution, \\
              CISO -- Compliance Assessment, \\
              FinOps -- Cost Management \\
                \end{tabular}
            & \cmark   
            & \cmark
            & Real Env.
            & \cmark\, (verified)\\
            
            TrainTicket%
            & 22
            & SRE -- Incident Diagnosis
            & \xmark
            & \xmark
            & Real Env.
            & \xmark \\
            
            AIOpsLab%
            & 10 
            & SRE -- Incident Resolution
            & \cmark
            & \xmark
            & Real Env.
            & \cmark\,(unverified) \\
            
            InsightBench%
            & 100
            & Ticket Data Analysis
            & \xmark
            & \xmark
            & Synthetic
            & \xmark \\
            
            TSB-AD%
            & 40
            & Anomaly Detection
            & \xmark
            & \cmark
            & Synthetic
            & \xmark \\
            
            CIS%
            & 1000+
            & Compliance/Security Focal
            & \cmark
            & \xmark
            & n/a (info. only)
            & \xmark \\
            
            \bottomrule
        \end{tabular}
        \begin{tablenotes}
        \item[1] \textbf{Note:} We are not aware of related benchmarks in the FinOps domain that go beyond scorecards.
        \end{tablenotes}
    \end{threeparttable}
\end{table*}

\bench targets a comprehensive set of tasks for a wide range of personas within IT automation. The initial release of \bench focuses on evaluating scenarios within IT Operations (ITOps). \cref{fig:motivation} illustrates currently targeted personas and exemplar tasks that they are routinely facing.
There is clearly a rising trend and interest in developing benchmarks to evaluate AI and ML techniques in ITOps with specific focus on SRE, CISO and FinOps.

TrainTicket~\cite{zhou2018fault} provides 22 scenarios
collected through an industrial survey of real-world incidents, using hardcoded faults in the TrainTicket application to focus on fault localization. AIOpsLab~\cite{aiopslab} provides 10 SRE-focused scenarios  
(referred to as problems) utilizing a real environment (system) integration that allows interactive access to text, time series, and tabular data. InsightBench \cite{sahu2024insightbench} provides 100 scenarios to analyze ticket data using static tabular data and synthetic scenarios. TSB-AD \cite{TSBench} focuses on anomaly detection with 40 synthetic scenarios. %

CIS-Benchmark~\cite{cis-b} provides best practices for securing IT infrastructure. Despite the name of ``benchmark'', it offers only 
recommendation policies; it provides no experimental platform. %
Recently, Cloud Native Compute Foundation (CNCF) Sandbox project
\cite{oscal-compass} released an SDK to support the translation of the CIS human readable formats into \cite{oscal} compliance as code standard of the National Institute of Standards and Technology for programmatic usage in compliance automation. \bench CISO automation leverages this technology to assess policy requirements. 

FinOps Foundation \cite{finopsbench}, provides benchmarks that compare cloud financial performance across organizations and departments, focusing on KPIs such as resource utilization efficiency, contract coverage, and cost apportionment. These benchmarks help assess cloud efficiency by evaluating internal and external metrics, fostering structured, collaborative approaches to cloud optimization. 

While existing benchmarks are valuable resources for specific tasks and use cases, and highlight the critical need for systematic benchmarking,  they are limited in reflecting real-world IT problems, covering broad IT landscape, and automating evaluation. These limitations are addressed in \bench, as shown in Table~\ref{tab:aiops_benchmarks}.

%% file: sections/03_bench.tex
\begin{figure*}
    \centering    
    \includegraphics[width=0.77\linewidth]{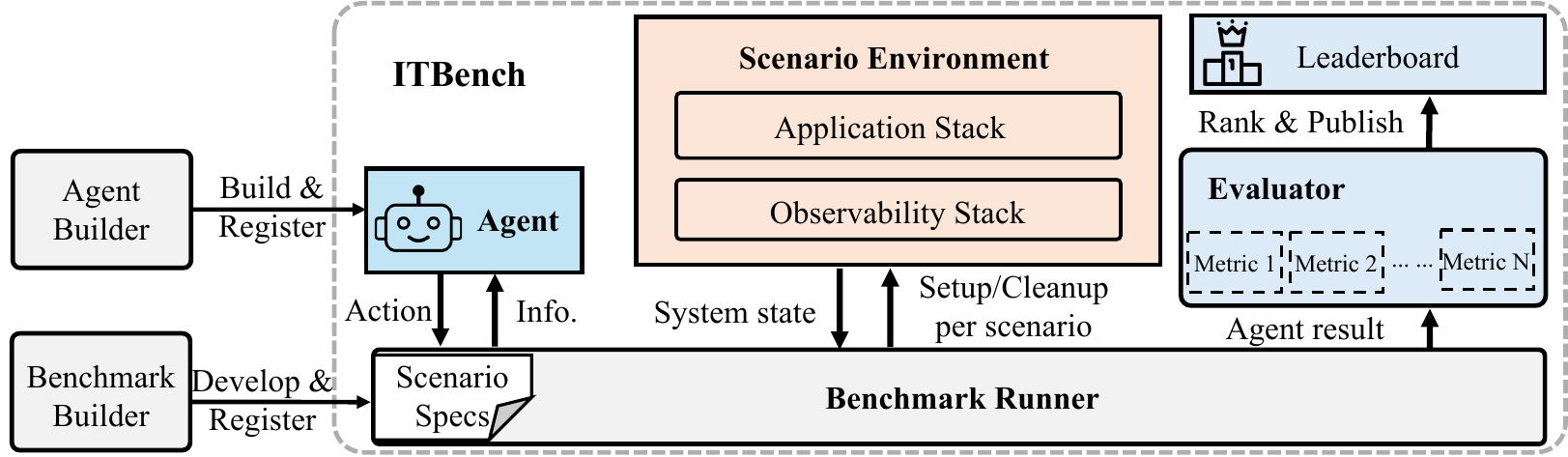}
    \vspace{-8pt}
    \caption{\bench automation framework.}
\label{fig:bench_design1}
\vspace{-10pt}
\end{figure*}

\section{\bench}
\label{sec:bench}

ITBench is a systematic benchmarking framework and runtime environment designed to evaluate AI agents tasked with automating IT operations, incorporating a robust architecture (see \Cref{fig:bench_design1}) comprising the AI Agent, Scenario Specification and Environment, Evaluator, and Leaderboard to facilitate comprehensive performance assessment.

Here, we present a brief overview of the key components: 1) Scenario Specification and Environment, 2) AI Agents, and 3) Leaderboard. More details are in Appendix \ref{appx:framework}.

\subsection{Scenario Specification and Environment}
\label{sec:scenarioenv}

The bench incorporates a collection of problems that we call \textit{scenarios}. For example, one of the problems in \bench is to resolve a ``High error rate on service order-management'' in a Kubernetes environment. Another example that is relevant for CISO persona involves assessing the compliance posture for a ``new control rule detected for RHEL 9.'' A fundamental challenge is to emulate such problems in a manageable testbed environment. A scenario environment is an operational testbed in which a specific problem(s) occurs. 

\begin{figure*}[t]
    \begin{subfigure}{0.23\linewidth}
        \centering
        \includegraphics[width=\linewidth]{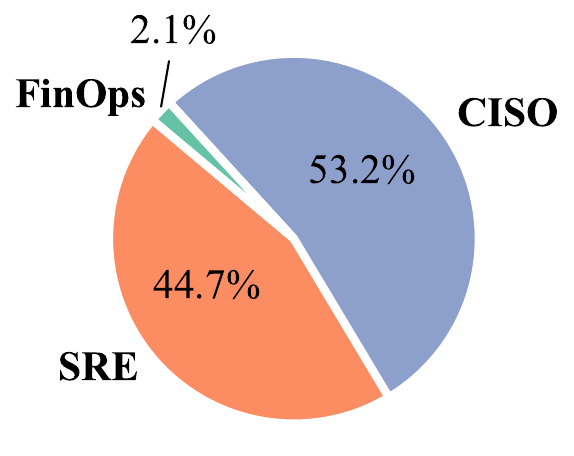}
        \caption{Scenarios by persona.}
        \label{fig:role-task-breakdown}
    \end{subfigure}
    \hfill
    \begin{subfigure}{0.23\linewidth}
        \centering
        \includegraphics[width=\linewidth]{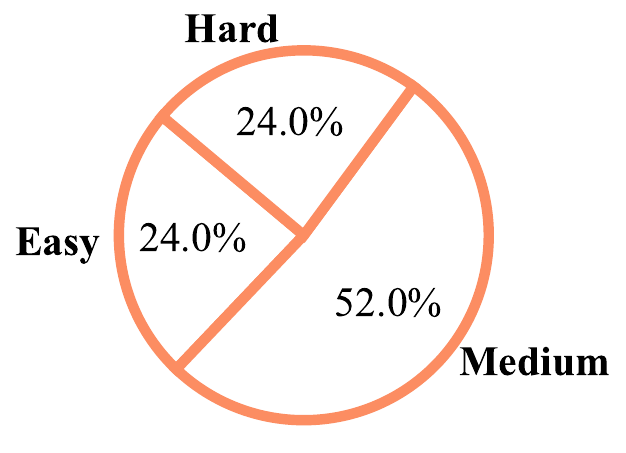}
        \caption{SRE scenario complexity.}
        \label{fig:sre-task-difficulty}
    \end{subfigure}
    \hfill
    \begin{subfigure}{0.23\linewidth}
        \centering
        \includegraphics[width=\linewidth]{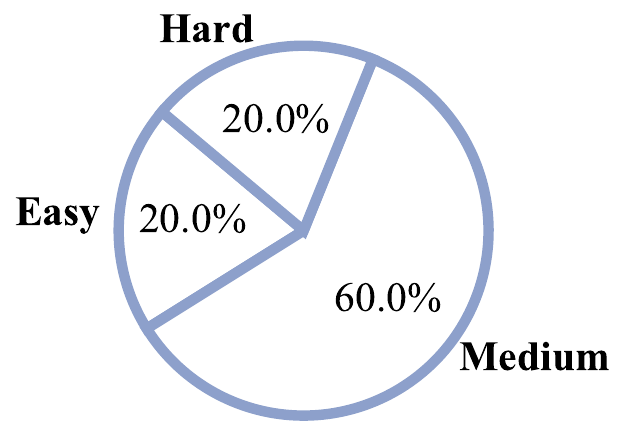}
        \caption{CISO scenario complexity.}
        \label{fig:ciso-task-difficulty}
    \end{subfigure}
      \hfill
    \begin{subfigure}{0.23\linewidth}
        \centering
        \includegraphics[width=0.78\linewidth]{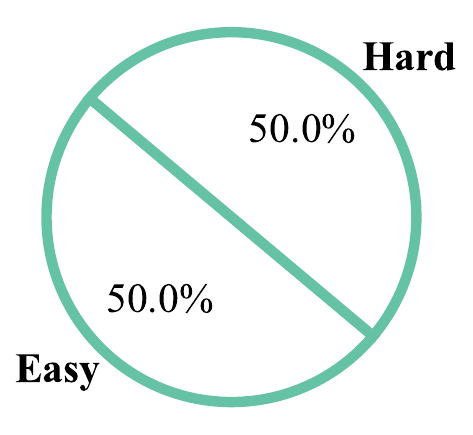}
        \caption{FinOps scenario complexity.}
        \label{fig:finops-task-difficulty}
    \end{subfigure}
    \caption{Characterization of \bench scenarios.}
    \label{fig:task-analysis}
\end{figure*}

A scenario $p$ generally corresponds to a problem to be solved in \bench. We formalize $p$ as a tuple $<M, E, T, D>$, where the variables are as follows:

\textbf{Scenario Specification.} $M$ represents metadata and deployment descriptors, for each scenario, which is stored in the Scenario Specs database in \bench (see \Cref{fig:bench_design1}). Exemplar metadata elements per scenario include \textit{scenario\_name}, \textit{scenario\_description}, \textit{scenario\_domain}, \textit{scenario\_class}, \textit{scenario\_complexity}, and \textit{scenario\_groundtruth} (see \Cref{tab:bench_scenarios}), which are defined below:
\begin{itemize}[left=0pt, topsep=0pt, partopsep=0pt, itemsep=0pt, parsep=0pt]
    \item \textit{scenario\_name} is name given to a scenario. For example, a scenario in \bench has name "Recommendation Service Cache."
    \item \textit{scenario\_description} describes the scenario. Example description of the scenario is "Recommendation Service in Astronomy Shop has a cache failure."
    \item \textit{scenario\_domain} represents different personas namely "SRE", "CISO", "FinOps" within IT automation. 
    \item \textit{scenario\_class} is used to group similar scenarios, such as "Kyverno-opa", "Kyverno-update", "CacheFailure", "HighCPU", and "CorruptImage".
    \item \textit{scenario\_complexity} captures the difficulty of a problem and is defined using domain knowledge. \Cref{fig:role-task-breakdown}, shows the breakdown of SRE, CISO, and FinOps scenarios in the bench. \Cref{fig:sre-task-difficulty}, \ref{fig:ciso-task-difficulty}, and \ref{fig:finops-task-difficulty} shows \textit{scenario\_complexity} distribution for SRE, CISO, and FinOps, respectively. SRE scenarios are developed based on real-world incidents observed in our own SaaS products. CISO scenarios are based on CIS benchmark \cite{CISBenchmarks}. FinOps scenarios are sparsely represented in \bench due to the lack of standard benchmarks. We based our scenario using ``Domains'' and ``Capabilities'' identified by the FinOps Foundation \cite{finopsbench} to describe key business outcomes. 
    \item \textit{scenario\_groundtruth} records task-specific outcomes that the Evaluator uses to compare against the agent's expected output. For instance, in incident resolution for SREs, the ground truth for the Diagnosis task includes a list of entities involved in the fault propagation chain, the actual fault propagation chain(s), and fault conditions, while for the Mitigation task, it captures plausible mitigation actions.
\end{itemize}

\textbf{Environment.} $E$ represents an an operational testbed where the problem occurs. Components within the environment expose APIs to observe and control the environment. When the Agent Builder registers the agent for benchmarking, the Benchmark Runner (see \Cref{fig:bench_design1}) randomly selects a set of scenarios, which may be optionally filtered based on the \textit{agent\_type} and \textit{agent\_level}. Next, the Benchmark Runner iterates through the set of scenarios and for each scenario it instantiates a testbed. An example of an environment includes Kubernetes cluster installed with OpenTelemetry Astronomy Shop Demo application \cite{otelastronomy}, observability stack including Grafana \cite{grafana}, Loki \cite{loki}, Jaeger \cite{jaeger}, and Prometheus \cite{prometheus}, along with mechanisms that induce problem(s) in the environment.

\textbf{Triggering Events.} $T$ is a set of triggering events that occur due to manifestation of a specific problem in the environment. Tools are configured to observe the environment and raise triggering events on problematic conditions. An example of a triggering event is "High Error Rate on adservice", which may be triggered in the environment due to cache failure problem.

\textbf{Desired Outcome.} $D$ defines the automation objective and represents the ultimate goal. For instanace, in case of SRE incident resolution, the ultimate goal is to clear $T$ in the $E$.

\begin{table*}[h]
    \centering
    \footnotesize
    \begin{threeparttable}
        \caption{Exemplar scenario classes and their complexity per scenario domain in \bench across 94 scenarios.}
        \label{tab:bench_scenarios}
        \begin{tabular}{m{0.08\textwidth}m{0.47\textwidth}m{0.15\textwidth}m{0.2\textwidth}}
            \toprule
            \textbf{Scenario Domain} & \textbf{Scenario Class} & \textbf{Scenario \mbox{Complexity}} & \textbf{Technologies} \\
            \midrule
             SRE & CacheFailure - Create a memory leak due to an exponentially growing cache            & Medium              & K8s, Redis, MongoDB \\
            & HighCPU - Trigger high CPU load in target service          & Medium                 & K8, Host, Pods \\
            & CorruptImage - Deployment uses wrong Docker image & Easy              & K8s, Image registry\\
            & HTTPRequestBodyTamperFault - Modify HTTP Post request between services     & Medium                & K8s, ingress/egress  \\
            & HTTPRequestAbortFault - Interrupt HTTP connection between services     & Medium                & K8ss, ingress/egress  \\
            & MemoryResourceLimit - Reduce memory limit on target service    & Easy                & K8s, Host, Pod  \\\midrule
            CISO & New K8s CIS-benchmarks on Kyverno & Easy            & K8, Kyverno  \\
            & New K8s CIS-benchmarks on OPA           & Medium            & K8s, OPA, Kubectl  \\
            & New RHEL9 CIS-benchmarks on Ansible-OPA         & Medium            & RHEL9, OPA, Ansible \\
            & Update K8s CIS-benchmarks on Kyverno               & Hard              & K8s, Kyverno   \\ \midrule
            FinOps & CostAlertMisconfiguration - Alert threshold is too low causing false alerts         &          Easy     & K8s, HPA  \\
            & AutoscalerMisconfiguration - Horizontal pod autoscaler thresholds are misconfigured creating excess pods           &    Hard          & K8s, HPA \\
            \bottomrule
        \end{tabular}
        \begin{tablenotes}
        \item[1] Scenario complexity  depends on the characteristics of the scenario, and is independent from agent capability. See appendices for details.
        \item[2] K8s refers to Kubernetes~\cite{kubernetes}.
        \item[3] Here, `technologies' refers to the set of tools and systems that a domain expert must understand to handle the task.
        \end{tablenotes}
    \end{threeparttable}
\end{table*}

\subsection{AI Agents}
\label{sec:baseline-agent}

\begin{figure}[h]
    \centering
    \includegraphics[width=\linewidth]{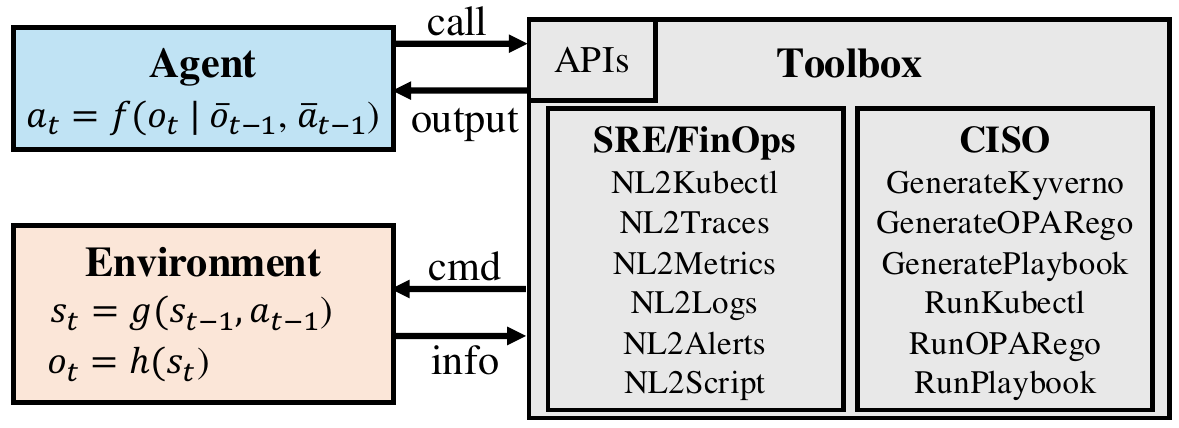}
    \caption{Agent and environment as a POMDP. Agents interact with the environment via the APIs exposed by \bench's toolbox.}
    \vspace{-10pt}
    \label{fig:bench-agent-POMDP}
\end{figure}

\begin{table}[h]
    \centering
    \small
    \begin{threeparttable}
        \caption{Personas, tasks, and metrics in \bench.}
        \label{tab:bench_tasks_metrics}
        \begin{tabular}{llp{2.7cm}}
            \toprule
            \textbf{Personas} &  
            \textbf{Tasks} &
            \textbf{Metrics} \\
            \midrule
             SRE & Diagnosis & pass@1, Fault Localization, Fault Propagation Chain, Mean Time to Diagnosis\\
              &    Mitigation & pass@1, Mean Time to Repair  \\
            \midrule
              CISO & Collect evidence & pass@1   \\
                   & Scan assessment posture & pass@1, Time to Process  \\
            \midrule
             FinOps & Identify inefficiency & pass@1 \\
             &    Mitigate inefficiency & pass@1, Hourly infra cost, Efficiency \\
            \bottomrule
        \end{tabular}
    \end{threeparttable}
\end{table}

%% file: sections/04_experiments.tex
In IT automation, the different personas are focused on specific desired outcome, which defines their automation goals. 
For SREs, incident resolution is the primary objective. Achieving this can involve multiple steps, such as diagnosing an incident, or a single step, like generating a diagnosis report. CISO persona focuses on the regulatory controls posture assessment process, including \textit{Collect evidence} and \textit{Scan assessment posture} tasks. FinOps persona focuses on the cost management, where sample tasks include \textit{Identify inefficiency} and \textit{Mitigate inefficiency}. During evaluation, each step (task) is assessed independently and is measured using well defined metrics, see \Cref{tab:bench_tasks_metrics}.

The goal of \bench is to evaluate AI agents on a broad range of real-world IT automation tasks that are otherwise performed by SREs, FinOps, CISO personas.

In this paper, an AI \textit{agent} is defined as an autonomous or semi-autonomous software program that uses an LLM to plan, make decisions, interact with the target environment, and execute actions to achieve goals.  An AI agent is expected to successfully handle any of the scenarios in the \bench, by interacting with the environment. 

As shown in \Cref{fig:bench-agent-POMDP}, agent and environment form a Partially Observed Markov Decision Process (POMDP), where the state is the snapshot of the environment. The state transitions are determined by the environment, which are then (partially) observed by the agent. %

Given a scenario $p$ instantiated in an environment $E$, an agent probes the environment via one of the tools and receives an observation $o_t \in \mathcal{O}$, based on which, it decides the next action: 
\begin{equation}
    a_t  = f(o_t|\bar{o}_{t-1};\bar{a}_{t-1})
\end{equation}
Here $f$ is the agent's decision function. 
$\bar{o}_{t-1}$ is the sequence of observations up to time $t-1$ and $\bar{a}_{t-1}$ is the sequence of actions taken up to $t-1$.  

At the beginning, $o_0$ may be a triggering event showing a problematic state $s_0$ of the environment. Given state $s_{t-1}$ and action $a_{t-1}$, the environment transitions to the next state: 
\begin{equation}
    s_{t} = g(s_{t-1},a_{t-1})
\end{equation} 
The observation $o_t$ is determined as a function of the state and is in general a proxy for the environment state $s_t$, hence the formulation can be thought of as a POMDP: 
\begin{equation}
    o_t = h(s_t)
\end{equation}

The set $\mathcal{A}$ of actions is defined as $\mathcal{Q} \bigcup \{\bot\}$, where $\mathcal{Q}$ is the set of tools and $\bot$ represents the `stop action' by the agent. 
We define $t^*$ as the time when agent stops: 
\begin{equation}
    t^*=\min \{t | a_t=\bot \} 
\end{equation}

An agent reflects on the result to guide its next action, continuing until the final goal is achieved. Given a set of scenarios that the agent works on, it targets to maximize the success defined as follows: 
\begin{equation}
    \mathbb{E}_{p \sim \pi_p}(\mathbb{I}(g(s^{p}_{t^*}, f(o_{t^*}|\bar{o}_{t^*-1},\bar{a}_{t^*-1}))=s^p_G))
\end{equation}
where $\mathbb{I}$ is an indicator function comparing the terminating state with goal state, $\pi$ is the distribution of scenarios.

\subsection{Baseline AI Agents}
We developed baseline agents: \lumyn for SRE, Compliance Assessment Agent for CISO, and FinOps-agent for FinOps.
Each of these agents uses state-of-the-art agentic techniques such as ReAct-based planning~\cite{yao2023react}, reflection~\cite{shinn2023reflexion}, and disaggregation~\cite{xu2023rewoodecouplingreasoningobservations}.
Reflection techniques vary from syntax checking/linting, semantic validation~\cite{xie2024travelplannerbenchmarkrealworldplanning}, and llm-as-a-judge~\cite{zheng2023judgingllmasajudgemtbenchchatbot}.

We open source two baseline agents (SRE\footnote{\url{https://github.com/IBM/itbench-sre-agent}} and CISO\footnote{\url{https://github.com/IBM/itbench-ciso-caa-agent}}) along with \bench. We use the open-source CrewAI framework \cite{crewai} to create and manage agents. The agents can be configured to use various LLMs either through watsonx, Azure, or vLLM. Each agent is initialized with a prompt that describes its goal, the context, the tasks, and the expected output format. In-context learning examples are included to guide the agent and demonstrate tool usage. Agents use natural language to access tools to interact with the environment for information gathering. 

Logs, traces, and metrics collected during the diagnosis process would overwhelm the context window of any LLM currently available due to large volume of data. Therefore, agent targeting the SRE or FinOps persona are equipped with specialized tools to interact with the environment (refer to \Cref{fig:bench-agent-POMDP}): 1) NL2Traces to extract trace data in a structured format, 2) NL2Metrics to analyze key system metrics, 3) NL2Logs to parse log data effectively, 4) NL2Kubectl to perform Kubernetes-specific operations, and a summarization tool to condense extensive data into actionable insights. For example, agent may use the NL2Kubectl tool to ``list all of the pods in the default namespace.'' In turn, NL2Kubectl tool uses an LLM to transform the utterance into an executable command, i.e.``kubectl get pods -n default''. 

Similarly, the compliance assessment required for new regulations and technologies, with the evidence and diverse policy languages would be overwhelming if submitted directly to LLMs. 
The compliance agents designed for CISO compliance assessment automation are equipped with specialized tools. 
These tools include capabilities to 1) generate policies such as Kyverno or OPA Rego Policy as Code starting from natural language specifications, 2) generate scripts for the collection of evidence, 3) access code repositories such as git to facilitate GitOps workflows for code management, and 4) deploy and execute the generated policies to accomplish the assessment task. %

\subsection{Leaderboard} 
\bench includes a leaderboard to promote reproducibility and comparative analysis, following the AI common task framework  \cite{Donoho2019,VarshneyKS2019}. %
The leaderboard offers a predefined, extensible set of performance metrics designed to provide clear insights into agent performance relative to the evaluation criteria.

\bench devises \textit{scoring methods for partially correct solutions} to provide meaningful feedback for summative assessments. %
This comprehensive approach establishes a new standard for evaluating and advancing AI-driven solutions in IT automation. For each scenario that an agent works on, upon task completion, the \bench
records the final system state, which is then used at the end of all scenario runs along with the pre-defined ground truth data to validate 
how well the agent performed across all the scenarios. For each scenario that an agent works on, upon task completion, the \bench
records the final system state, which is then used at the end of all scenario runs along with the pre-defined ground truth data to validate 
how well the agent performed across all the scenarios.

We are open-sourcing a small subset (11 out of 94) of scenarios \bench\footnote{\url{https://github.com/IBM/itbench-sample-scenarios}} along with the baseline agents to help the community familiarize with \bench through practical examples. We reserve the remaining scenarios in \bench to benchmark and evaluate the submitted agentic solutions.

%% file: sections/05_evaluation.tex
\begin{table*}[h]
\small
\centering
\setlength{\tabcolsep}{4pt}
\begin{threeparttable}
  \caption{Evaluation of \lumyn on SRE scenarios}
  \label{tab:sreagent-eval}
  \begin{tabular}{@{}lcccccc@{}}
    \toprule
    \multirow{2}{*}{\textbf{Models}}
      & \multicolumn{4}{c}{\textbf{Diagnosis}}
      & \multicolumn{2}{c}{\textbf{Mitigation}} \\
    \cmidrule(lr){2-5}\cmidrule(lr){6-7}
    & \textbf{pass@1 (\%)$\uparrow$}
    & \textbf{FL (NTAM)$\uparrow$}
    & \textbf{FPC (NTAM)$\uparrow$}
    & \textbf{MTTD (s)$\downarrow$}
    & \textbf{pass@1 (\%)$\uparrow$}
    & \textbf{MTTR (s)$\downarrow$}\\
    \midrule
    \textbf{granite-3.1-8B-instruct} &
    \cellcolor[gray]{0.97} $3.57 \pm 0.94$ &
    \cellcolor[gray]{0.96} $0.16 \pm 0.02$ &
    \cellcolor[gray]{0.94} $0.19 \pm 0.02$ &
    $259.92 \pm 65.01$ &
    $0.24 \pm 0.25$ &
    $845.50 \pm \text{---}$ \\
    \textbf{llama-3.1-8B-instruct} &
    $0.99 \pm 0.51$ &
    $0.07 \pm 0.01$ &
    $0.08 \pm 0.01$ &
    \cellcolor[gray]{0.85} $\textbf{57.50} \pm 2.05$ &
    \cellcolor[gray]{0.98} $1.98 \pm 0.68$ &
    \cellcolor[gray]{0.85} $\textbf{245.13} \pm 40.66$ \\
    \textbf{llama-3.3-70B-instruct} &
    \cellcolor[gray]{0.98} $3.10 \pm 0.84$ &
    \cellcolor[gray]{0.96} $0.16 \pm 0.02$ &
    \cellcolor[gray]{0.95} $0.16 \pm 0.02$ &
    \cellcolor[gray]{0.95} $191.85 \pm 31.34$ &
    \cellcolor[gray]{0.96} $3.33 \pm 0.90$ &
    \cellcolor[gray]{0.98} $776.27 \pm 252.87$ \\
    \textbf{gpt-4o} &
    \cellcolor[gray]{0.85} $\textbf{13.81} \pm 1.67$ &
    \cellcolor[gray]{0.85} $\textbf{0.39} \pm 0.05$ &
    \cellcolor[gray]{0.85} $\textbf{0.34} \pm 0.03$ &
    \cellcolor[gray]{0.86} $72.44 \pm 4.71$ &
    \cellcolor[gray]{0.85} $\textbf{11.43} \pm 1.52$ &
    \cellcolor[gray]{0.86} $282.47 \pm 30.04$ \\
    \bottomrule
  \end{tabular}
  \begin{tablenotes}
    \scriptsize
    \item[1] 42 scenarios (21 scenarios with traces and 21 without traces).
    \item[2] 10 runs per scenario per model.
    \item[3] pass@1 values are shown as percentages. `\text{---}' indicates missing data. 
    \item[4] std error for each metric is listed.
    \item[5] \textbf{FL (NTAM)} = Normalized topology-aware metric for root cause, 
          \textbf{FPC (NTAM)} = Normalized topology-aware metric for fault propagation chain (value between 0 and 1.0), 
          \textbf{MTTD} = Mean time to diagnosis (seconds), 
          \textbf{MTTR} = Mean time to repair (seconds). \textbf{Bold}: the best performance.
    \item[6] Details of NTAM are available in \Cref{appx:ntam}
  \end{tablenotes}
\end{threeparttable}
\end{table*}

\section{Results}
\label{sec:results}

\subsection{Evaluation Setup}

To understand the impact of reasoning and planning capabilities of LLMs on \bench scenarios, we instantiate our agents using different LLM models, both for natural language reasoning and code generation. 
Specifically, we employ GPT-4o\footnote{Checkpoint version 2024-11-20}, Llama-3.3-70B-instruct, Llama-3.1-8B-instruct, and Granite-3.1-8B-instruct for tasks that rely on natural language understanding and reasoning. For code-focused use cases, we utilize GPT-4o-mini, Llama-3.1-405b-instruct, and Mixtral-8x7b-instruct. 
All models use a context window of 128K tokens, enabling them to process more extensive input sequences.

We conduct our experiments primarily on AWS EC2 instances (m4.xlarge), although \bench can also be readily deployed on a consumer-grade laptop using a pseudo-cluster, thus making it easier to develop AI agents (Appendix \ref{appx:sre:exp_setup})

Below, we provide an overview of our baseline agents’ performance across \bench scenarios for SRE, CISO, and FinOps. Our findings indicate that both open-source and proprietary models often struggle with real-world tasks, underscoring the importance of benchmarks that push the limits of reasoning and planning in foundation models. For more comprehensive results and detailed scenario-level discussions, please refer to Appendix~\ref{appx:sre} (SRE), Appendix~\ref{appx:ciso} (CISO), and Appendix~\ref{appx:finops} (FinOps).

\begin{table*}[h]
\small
\centering
\setlength{\tabcolsep}{4pt}
\begin{threeparttable}
  \caption{Evaluation of CISO Compliance Assessment Agent on CISO scenarios}
  \label{tab:cisoagent-eval}
  \begin{tabular}{@{}lcccccc@{}}
    \toprule
    \multirow{2}{*}{\textbf{Models}}
      & \multicolumn{4}{c}{\textbf{Scenario pass@1 (\%) $\uparrow$}}
      & \multirow{2}{*}{\textbf{O/A pass@1 (\%) $\uparrow$}} 
      & \multirow{2}{*}{\textbf{TTP (s) $\downarrow$}} \\
    \cmidrule(lr){2-5}
    & \textbf{kyverno}
    & \textbf{k8s-opa}
    & \textbf{rhel-opa}
    & \textbf{kyverno-update} \\
    \midrule
    \textbf{granite-3.1-8B-instruct} &
    \cellcolor[gray]{0.99} $7.84 \pm 3.84$ &
    \cellcolor[gray]{1.00} $0.00 \pm 0.00$ &
    \cellcolor[gray]{1.00} $0.00 \pm 0.00$ &
    \cellcolor[gray]{1.00} $1.59 \pm 1.58$ &
    \cellcolor[gray]{1.00} $1.71 \pm 0.76$ &
    \cellcolor[gray]{1.00} $197.03 \pm 2.52$ \\
    \textbf{mixtral-8x7B-instruct} &
    \cellcolor[gray]{1.00} $7.35 \pm 3.19$ &
    \cellcolor[gray]{1.00} $1.43 \pm 1.42$ &
    \cellcolor[gray]{1.00} $0.00 \pm 0.00$ &
    \cellcolor[gray]{1.00} $1.29 \pm 4.34$ &
    \cellcolor[gray]{0.99} $3.94 \pm 1.03$ &
    \cellcolor[gray]{0.88} $120.63 \pm 3.77$ \\
    \textbf{llama-3.1-8B-instruct} &
    \cellcolor[gray]{0.99} $8.57 \pm 3.37$ &
    \cellcolor[gray]{1.00} $0.00 \pm 0.00$ &
    \cellcolor[gray]{1.00} $0.00 \pm 0.00$ &
    \cellcolor[gray]{0.94} $7.46 \pm 3.23$ &
    \cellcolor[gray]{0.99} $3.59 \pm 1.07$ &
    \cellcolor[gray]{0.88} $121.49 \pm 3.00$ \\
    \textbf{llama-3.3-70B-instruct} &
    \cellcolor[gray]{0.95} $18.46 \pm 4.94$ &
    \cellcolor[gray]{1.00} $0.00 \pm 0.00$ &
    \cellcolor[gray]{0.99} $1.43 \pm 2.88$ &
    \cellcolor[gray]{0.94} $8.06 \pm 3.50$ &
    \cellcolor[gray]{0.95} $9.32 \pm 1.67$ &
    \cellcolor[gray]{0.99} $189.61 \pm 2.71$ \\
    \textbf{mistral-large-2} &
    \cellcolor[gray]{1.00} $6.56 \pm 3.20$ &
    \cellcolor[gray]{0.92} $22.73 \pm 5.32$ &
    \cellcolor[gray]{0.96} $7.23 \pm 2.88$ &
    \cellcolor[gray]{0.92} $10.45 \pm 3.77$ &
    \cellcolor[gray]{0.94} $11.55 \pm 1.95$ &
    \cellcolor[gray]{0.95} $167.98 \pm 3.42$ \\
    \textbf{llama-3.1-405B-instruct} &
    \cellcolor[gray]{0.96} $16.22 \pm 4.32$ &
    \cellcolor[gray]{0.93} $20.83 \pm 4.86$ &
    \cellcolor[gray]{0.96} $8.75 \pm 3.26$ &
    \cellcolor[gray]{0.98} $3.17 \pm 2.22$ &
    \cellcolor[gray]{0.93} $12.46 \pm 1.98$ &
    \cellcolor[gray]{0.97} $178.89 \pm 3.37$ \\
    \textbf{gpt-4o-mini} &
    \cellcolor[gray]{0.96} $16.18 \pm 4.54$ &
    \cellcolor[gray]{0.85} $\textbf{43.10} \pm 6.99$ &
    \cellcolor[gray]{0.85} $\textbf{30.38} \pm 5.43$ &
    \cellcolor[gray]{0.93} $9.43 \pm 4.08$ &
    \cellcolor[gray]{0.85} $\textbf{25.19} \pm 2.80$ &
    \cellcolor[gray]{0.85} $102.40 \pm 3.70$ \\
    \textbf{gpt-4o} &
    \cellcolor[gray]{0.85} $\textbf{40.28} \pm 5.99$ &
    \cellcolor[gray]{0.86} $39.34 \pm 6.55$ &
    \cellcolor[gray]{0.96} $7.61 \pm 2.81$ &
    \cellcolor[gray]{0.85} $\textbf{17.74} \pm 4.92$ &
    \cellcolor[gray]{0.85} $24.74 \pm 2.64$ &
    \cellcolor[gray]{0.85} $\textbf{101.29} \pm 3.81$ \\
    \bottomrule
  \end{tabular}
  \begin{tablenotes}
    \scriptsize
    \item[1] 50 scenarios.
    \item[2] 8 runs per scenario per model.
    \item[3] pass@1 values are shown as percentages.
    \item[4] TTP Time to process (seconds).\\
    \item[5] \textbf{kyverno} = New K8s CIS-benchmarks on Kyverno, easy scenario class; 
          \textbf{k8s-opa} = New K8s CIS-benchmarks on OPA, medium scenario class;
          \textbf{rhel-opa} = New RHEL9 CIS-benchmarks on Ansible-OPA, medium scenario class;
          \textbf{kyverno-update} = Update K8s CIS-benchmarks on Kyverno, hard scenario class.
  \end{tablenotes}
  \vspace{-10pt}
\end{threeparttable}
\end{table*}

\begin{table*}[h]
\small
\centering
\begin{threeparttable}
  \caption{Evaluation of FinOpsAgent on FinOps scenarios.}
  \label{tab:finopsagent-eval}
  \begin{tabular}{@{}lccp{1.85cm}p{1.85cm}p{1.85cm}p{1.85cm}@{}}
    \toprule
    \multirow{2}{*}{\textbf{Models}} 
      & \multicolumn{1}{c}{\textbf{Diagnosis}}
      & \multicolumn{5}{c}{\textbf{Mitigation}} \\
    \cmidrule(lr){2-2}\cmidrule(lr){3-7}
     & \textbf{pass@1 (\%) $\uparrow$} 
     & \textbf{pass@1 (\%) $\uparrow$} 
     & \textbf{Proximity to Optimal CPU Cost $\uparrow$} 
     & \textbf{Proximity to Optimal Memory Cost $\uparrow$} 
     & \textbf{Proximity to Optimal CPU Efficiency $\uparrow$} 
     & \textbf{Proximity to Optimal Memory Efficiency $\uparrow$} \\
    \midrule
    \textbf{granite-3.1-8B-instruct} 
      & 0 
      & 0 
      & $0.47 \pm 0.01$ 
      & \cellcolor[gray]{0.94} $0.48 \pm 0.06$ 
      & $0.53 \pm 0.04$ 
      & \cellcolor[gray]{0.93} $0.94 \pm 0.01$ \\
    \textbf{llama-3.1-8B-instruct} 
      & 0 
      & 0 
      & \cellcolor[gray]{0.85} $\textbf{0.49} \pm 0.01$ 
      & $0.46 \pm 0.07$ 
      & \cellcolor[gray]{0.95} $0.56 \pm 0.08$ 
      & \cellcolor[gray]{0.85} $0.96 \pm 0.02$ \\
    \textbf{llama-3.3-70B-instruct} 
      & \cellcolor[gray]{0.92} 16.6
      & 0 
      & $0.47 \pm 0.01$ 
      & \cellcolor[gray]{0.91} $0.49 \pm 0.05$ 
      & $0.53 \pm 0.03$ 
      & \cellcolor[gray]{0.85} $0.96 \pm 0.02$ \\
    \textbf{gpt-4o} 
      & \cellcolor[gray]{0.85} \textbf{33} 
      & 0 
      & \cellcolor[gray]{0.93} $0.48 \pm 0.01$ 
      & \cellcolor[gray]{0.85} $0.51 \pm 0.02$ 
      & \cellcolor[gray]{0.85} $\textbf{0.63} \pm 0.07$ 
      & $0.92 \pm 0.08$ \\
    \bottomrule
  \end{tabular}
  \begin{tablenotes}
    \scriptsize
    \item pass@1 values are shown as percentages. 
    \item Proximity values shows how close the observed values to optimal values. 
    One represents achieving optimal and any deviations from 1 represents sub-optimal performance.
  \end{tablenotes}
\end{threeparttable}
\end{table*}

\subsection{Overall Results}
\Cref{tab:sreagent-eval}, \Cref{tab:cisoagent-eval} and \Cref{tab:finopsagent-eval} show the performance of SRE-agent, CISO-agent, and FinOps-agent respectively. 

\textbf{SRE.} 
We measure the efficiency of \lumyn on its ability to diagnose and mitigate production incidents (e.g., ``a high error rate on frontend service'').

Diagnosis efficiency is measured using pass@1\cite{chen2021evaluating} (i.e., identifying the cause as mentioned in ground truth), NTAM (Normalized Topology-Aware Metric) for root cause and fault propagation chain, and time to diagnosis\footnote{NTAM is Normalized topology-aware metric that measures the quality of the predicted root cause and fault propagation chains using a system and application topology. Refer to \Cref{appx:ntam}.}.
Mitigation efficiency is measured in terms of pass@1 (i.e., whether the alert was cleared) and mean time to repair.

As shown in \Cref{tab:sreagent-eval}, across all SRE scenarios, GPT-4o consistently outperforms the other models, achieving the highest pass@1 scores for diagnosis (13.81\%) and mitigation (11.43\%), as well as the highest score on NTAM (FL and FPC) metrics. 
Llama-3.3-70B ranks second overall, trailing GPT-4o on most metrics.
The 8B models have lower mitigation success rate. 
Surprisingly, Granite-3.1-8B (without any specialized finetuning) achieves higher accuracy than Llama-3.1-70B on the diagnosis task. 

Removing trace data can drastically reduce success rates (see \Cref{tab:appx:sre:traces} and \Cref{tab:appx:sre:disabled} in Appendix). For instance, GPT-4o's pass@1 in diagnosis falls from 13.81\% with traces to 9.52\% without them, and mitigation plummets to 2.86\%. This highlights the critical role of system observability in SRE, which \bench can evaluate under varying conditions. As there is no perfect observability in practice, how to guide SRE-agents to collect new observability data and to help SRE-agents reason about failures with incomplete observability is an important but open problem.

\textbf{CISO.}
We measure the efficacy of our agents across the four scenario classes introduced in \Cref{tab:bench_scenarios}. Each \textit{scenario\_class} imposes a distinct set of CIS-benchmarks requirements (e.g., ``minimize the admission of containers wishing to share the host network namespace''), each class has a specific level of complexity (e.g., Easy, Medium, or Hard), and generates scenario-specific code artifacts. 

The efficacy of CISO-agents is measured based on the ability to detect artifact misconfigurations (aka non-compliance, e.g., no minimum count of containers sharing namespace, or the count is above the threshold), or confirm proper configurations (aka compliance), within the varied environments of the scenario classes randomly injected with misconfigurations. 
Notably, GPT-based models dominate on both pass@1 and Time to Process metrics. The pass@1 is nearly 2x better than second-best models (alternating between llama-3.1-405b-instruct and mistral-large-2), while the TTP shows a handling of the scenarios in the minimal time across our scenario classes.

\textbf{FinOps.}
We measure the effectiveness of FinOps-agent on its ability to diagnose and mitigate the origin of cost alert (e.g., `increase in cost by 20\%'). 
Diagnosis effectiveness is measured using pass@1 (i.e., identifying the cause).
Mitigation effectiveness is measured in terms of proportional proximity to optimal cost of running, and efficiency that can be achieved for that workload.

GPT-4o consistently outperforms all other models, achieving a 33\% pass rate for diagnosing the origin of the cost increase alert. 
Performance on additional metrics related to cost and workload efficiency remains comparable across all models, with none attaining optimal CPU and memory cost or delivering high CPU efficiency.

\subsection{Impact of Scenario Complexity}
\textbf{SRE.}
    We categorize scenarios as Easy, Medium, or Hard based on factors such as fault propagation chain length, number of resolution steps, and the diversity of technologies involved, as described in \Cref{ss:bench-sre-eq-task-complexity}. 
    Our results show that success rates (pass@1) clearly decline as the \textit{scenario\_complexity} increases.
    For example, GPT-4o (the best performing model) diagnosed only 36\%, 7.73\% and 5.0\% of the Easy, Medium, and Hard scenarios, respectively (refer to \Cref{tab:sre:diag_pass1}).
    Similarly, GPT-4o (the best performing model) successfully mitigated only 21\%, 12.27\% and 0.0\% of Easy, Medium, and Hard scenarios, respectively 
    (refer to \Cref{tab:sre:repair_pass1}). 
    
    None of the models could mitigate the hard scenarios in any of the runs, whereas over half of the Easy scenarios see successful mitigation. 
    Notably, GPT-4o is the only model that successfully diagnosed multiple ``Hard'' scenarios.

\textbf{CISO.}
The complexity of the CISO scenarios is directly mapped to scenario classes. For example, \textit{scenario\_complexity} of Kyverno scenarios is Easy, \textit{scenario\_complexity} of k8s-opa and rhel-opa is Medium, while \textit{scenario\_complexity} of Kyverno-update scenarios is Hard. 
All models struggle, as expected, as the difficulty of the scenarios increases from the Easy \textit{kyverno} class to the Hard \textit{kyverno-upadate} class. 

\textbf{FinOps.}
Currently, \bench only has two FinOps scenarios, \textit{scenario\_complexity} of one is Easy and the other is Hard. None of the models, could diagnose (except for GPT-4o) or mitigate the hard scenario. 

This spectrum of complexity in \bench ensures that evaluations capture both straightforward and highly intricate problems across personas.

\subsection{Inherent Non-determinism in the Environment} 
GPT-4o remains the top performer across all evaluated personas (SRE, CISO, and FinOps), yet it still exhibits notable variability in scenario outcomes. 
For example, the SRE-agent with GPT-4o struggles to maintain deterministic behavior despite hyperparameter tuning aimed at ensuring consistency. 
SRE-agent with GPT-4o diagnosed the problem only in 6 out of 10 runs for scenario 13, 1 out of 10 runs for scenario 8, and 8 out of 10 runs for scenario 21, respectively (refer to \Cref{fig:sre:trace_on_diagnosis_pass1} for details on all scenarios).
Similarly, it mitigated 6 out of 10 runs for scenario 16, 2 out of 10 runs for scenario 8, and 5 out of 10 runs for scenario 21, respectively (refer to \Cref{fig:sre:trace_on_repair_pass1} for details on all scenarios).
This inherent non-determinism was observed with FinOps and CISO scenarios as well. 

These fluctuations arise from minor real-time telemetry changes, which can alter the large language model’s token generation. By tracking such dynamic behavior over multiple runs, \bench provides crucial insights into each agent’s robustness and reliability.

%% file: sections/07_conclusion.tex
\section{Discussion and Conclusion}
\label{sec:discuss}

We presented \bench, the first framework  and experimental platform to benchmark AI Agents for IT automation tasks. \bench strives to capture the complexity of modern IT systems and the diversity of IT tasks. The reproducibility of \bench ensures the community-driven effort despite inherent nondeterminism of large-scale IT systems. 

One of the key design principles of \bench is ensuring its flexibility to support diverse areas of different IT systems
and its extensibility to new scenarios. While current scope of \bench is comprehensive and representative, we plan to further enrich the benchmark suites by adding other important processes essential to modern IT automation. Furthermore, we plan to expand our benchmarking beyond event-triggered scenarios. 
We are actively working to expand scenario coverage for the supported processes and promote growth through open-community contributions.
 We invite the community to reproduce their real-world-inspired incidents in a synthetic sandboxed environment leveraging the \bench. We expect that everyone contributing can bring their expertise to the table.

We expect \bench to drive the innovations of AI agent-based techniques with a direct impact on the safety, efficiency, and intelligence of today’s IT infrastructures. 
With \bench, we are starting to explore many deep, exciting open problems: How to develop domain-specific AI agents that specialize in certain types of IT tasks? How to orchestrate multiple agents with various expertise to collaborate on bigger projects? How can we ensure safety of agent-driven solutions? How can we effectively use human-in-the-loop while developing diverse adaptive agents? We invite everyone to participate in answering these questions and realizing the vision of using AI agents to automate critical IT tasks.

%% file: sections/0x_statements.tex
\section{Statements}
\subsection{Ethics \& Broader Impacts}
\label{appx:statement:impacts}
This research presents a novel benchmarking framework to measure the performance of AI agents across a wide variety of complex and real-life IT tasks, which has the potential to be a key enabler for AI-driven IT automation that is correct, safe and fast. While the primary focus is on advancing the field of machine learning, as this effort is an open-source framework built with open-source technologies, it allows organizations with proprietary technologies to use it
for developing and benchmarking their solutions more effectively. 
It also encourages mindsharing in the community and lowers the barrier to innovate in IT domain.

Agents that interact with the system pose several risks.
We identify three main risks that could arise when building and using a \bench and associated agents, then discuss how we incorporates measures that mitigate such problems.

First is the security risks that come with executing LM-generated code/commands on the system. Examples include executing commands like \texttt{kubectl delete node} and  \texttt{rm -rf asset/}.
To defend against this, we containerize the agents, and also provide a self-contained Kubernetes environment to create various scenarios.  

Second, if the wider community develops interest for \bench and associated agents and builds upon it, it is also possible that illegitimate evaluation datasets or infrastructure can be used to inject testing devices with malicious code or instructions to generate malicious code.
For instance, an unofficial repository claiming to host an inference/evaluation harness for \bench and associated agents could include a task instance with an issue description that tells the LM agent to build key logging functionality and store it in a hidden folder.
To eliminate confusion and reduce the possibility of such an event, we provide clear guidelines listed on our GitHub repositories, data stores, and websites indicating the official repositories and channels that we actively maintain.
We also encourage third parties to incorporate any improvements into our codebase and help with integrating such contributions.

Lastly are the consequences of \bench agents being deployed in the real world.
Prior works have conceptualized and put forth prototypes of agents that can carry out offensive security measures.
It is also not difficult to imagine that a system like \lumyn can be incorporated into pipelines resulting in the production of malicious code and libraries.
The strong performance of agents on \bench implies that future AI systems will likely be increasingly adept in the aforementioned use cases.
Releasing \bench agents as open source agents can support research towards designing sound, effective constraints for what software engineering agents are permitted to do.
It can also serve as a system that legal experts and policy-making entities can experiment with to shape the future of what AI-driven end to end software engineering could look like.

\subsection{Reproducibility}
\label{appx:statement:reproducibility}
To help the greater community reproduce the results presented in this paper and build on the \bench, we open source all of our resources that were created for this project.
The source code for the interactive pipeline, context management logic, command implementations, interface design, and everything else is entirely available in a GitHub repository.
We provide extensive text and video documentation describing how to run and modify different parts of the codebase.
Practitioners should be able to easily recover our findings by running the agent with simple scripts.
The results presented in the main and supplementary parts of this paper can be fully obtained by following instructions in the repositories.
Finally, we also maintain an active online help forum to assist with any reproduction problems or questions about how to build on \bench.

%% file: sections/0x_acknowledge.tex
\section*{Acknowledgements}
We would like to thank everyone at IBM and University of Illinois at Urbana-Champaign not explicitly on the author list, who have shared excitement, given feedback on early prototype, and worked with or supported the core team on many aspects of this project. 

In addition, we acknowledge the support of our colleagues in IBM Instana: Marc Palaci-Olgun, Guangya Liu, Brad Blancett, Chad Holliday, Arthur De Magalhaes, Ragu Kattinakere, Chris Bailey, Isabell Sippili, and Danilo Florissi; IBM Granite and Data Model Factory: Hui Wu and Bing Zhang; IBM Emerging Technology Engineering: Aditya Gidh, Mike Sava, Bill Rippon, and Danny Barnett; IBM UX Research: James Sutton, Connor Leech, and Justin McNair, IBM Research: Michal Shmueli-Scheuer, Lilach Edelstein, and Roy Bar-Haim.

%% file: appx/00_table_of_contents.tex
\addcontentsline{toc}{section}{Appendix} 
\doparttoc                 
\faketableofcontents   
\part{Appendix} %
In the appendix, we provide additional analyses and more extensive discussions about \bench, individual personas (SRE, ComplianceOps, FinOps) and agent performance.

\parttoc %

%% file: appx/01_related_work.tex
\section{Related Work}
\textbf{LM and agents for resolving IT automation tasks.}
There is a surge in use of AI/ML for handling IT automation tasks.
We describe related work for each persona.

\subsection{Site Reliability Engineering}
IT scenario\footnote{We use the term scenario broadly to refer to failures, performance problems, compliance issues and, cost anomalies.} resolution encompasses tasks such as detection (e.g., identifying anomalies or outages)~\cite{10.1145/2829988.2787496, leners2011detecting, sigelman2010dapper, fonseca2007x}, diagnosis (e.g., pinpointing root causes through metrics and logs)~\cite{tan2019netbouncer,jha2020live, ma2014node, PyRCA}, and mitigation (e.g., operational fixes or code changes). 
These efforts often rely on supporting tasks like ticket analysis and routing~\cite{gao2020scouts,liu2023ticket, arzani2016taking}, anomaly detection~\cite{TSBench}, topology extraction~\cite{ashok2024traceweaver, chakraborty2023causil, pham2024root, yao2024chain}, causal~\cite{budhathoki2022causal, dowhy, ikram2022root, chakraborty2023causil} and interventional~\cite{wang2023fault, bagehorn2022fault} analysis using IT data.
Clearly, there is significant research in this area, fully automating incident resolution or providing actionable insights to humans remains elusive due to the complexity of real-world systems, the variability of incidents, and the challenge of incorporating contextual knowledge into AI systems~\cite{jha2020live}. 
Recent advances in language models (LMs) have led to their adoption of ticket data analysis and diagnosis tasks~\cite{roy2024exploringllmbasedagentsroot, 10172904,cloudatlas, chen2023empowering, rcaflash}. 
Most notable examples include Cloud Atlas use LLMs for causal graph construction~\cite{cloudatlas}, RCACopilot for ticket analysis~\cite{chen2023empowering} with the aim to diagnose and mitigate incidents.
However, they achieve poor performance compared to other techniques. For example, \cite{roy2024exploringllmbasedagentsroot} shows that chain-of-thought only achieves accuracy of 35\%. 
More recently, LMs are used in agentic workflows, engaging with real or virtual environments, using several tools at their disposal, for tasks like web navigation~\cite{workarena2024, workarena++, koh2024visualwebarena}, system control~\cite{sahu2024insightbench, rcaflash, aiopslab}, and code generation~\cite{yang2024swe}. 
However, the initial results of these works show a high variability in the success rate \textemdash 35\% in InsightBench~\cite{workarena++} and the ReAct-based agent for ticket data analysis~\cite{roy2024exploringllmbasedagentsroot} to 100\% in Flash~\cite{aiopslab} for incident resolution despite the fact that %
it is a much harder task than identifying planted insights in tabular and ticket data. 
Our own results in this work suggest that LLMs and agents struggle to consistently complete incident resolution tasks.
\textit{We assert that the variability in success rate exists because of difference in realism of these datasets.}
This highlights the urgent need for standardized and open source benchmarks to evaluate and improve the efficacy of AI methods on incident resolution tasks effectively.

\textbf{SRE-focused Benchmarks}
The benchmarking landscape for IT operations (ITOps) tasks is still in its early stages, with a few existing efforts addressing specific aspects of the domain. 
AIOpsLab~\cite{aiopslab} focuses on resolving IT incidents  \textit{only} for SRE personas, covering nine distinct problems created in a real environment. 
It does not follow SRE best practices for system and application observability, e.g., using an alert management system, lacks comprehensive coverage analysis, and a leaderboard for systematic automated evaluation. 

InsightBench~\cite{sahu2024insightbench} targets the analysis of ServiceNow ticket data, a critical supporting task for incident routing and finding relevant past incidents, but its reliance on synthetic data and the lack of a real environment limit its applicability to agentic workflows. 
Similarly, TSB-AD~\cite{liu2024elephant} is designed for univariate and multivariate anomaly detection, a core task for incident detection. However, it is limited to synthetic data and focuses only on anomaly detection.

\subsection{Compliance}
Compliance automation software is emerging to help businesses streamline and automate compliance processes, reducing the need for manual monitoring and tracking of regulations. This ensures continuous adherence to laws. In particular, compliance as code is a very recent development in the IT industry motivated by companies and audit agencies shifting from annual audits to expectations of continuous and automated measurement of compliance to maintain control of their regulated environments' posture and risks for cyberattacks. 

Recent works~\cite{survey, change-iaai2022} have applied AI/ML techniques to speed up these tasks, focusing on mapping regulatory requirements to standard control frameworks such as NIST 800-53~\cite{nist800-53catalog}. Our agentic automation in the current \bench solution pioneers this type of effort to author compliance artifacts through AI / ML by bridging compliance as code into policy as code. Policy engines have a longer history in the IT industry compared to compliance as code; however, emerging general usage policy engines such as~\cite{Intro:OPA} try to address the need for a common framework for continuous compliance. 
We are not aware of any effort -albeit critical and needed- related to benchmarking of compliance automation software, whether with or without agentic support.

\subsection{FinOps}
The area of IT cost management encompasses multiple disciplines, namely FinOps, IT Financial Management (ITFM), Technology Business Management (TBM) and Porfolio Businesss Management (PBM). At present, the FinOps domain typically deals with cloud costs \cite{cloud_finops_2nd, OptITfinops}, which includes compute nodes, memory, other storage, networking, etc., that are incurred with one of the hyperscalers.  ITFM includes on-prem infrastructure, licensing, labor, procured services, tech support, etc. The \href{https://www.tbmcouncil.org/}{TBM Council} provides a standard taxonomy to describe cost sources, technologies, IT resources (IT towers), applications, and services. In addition, there are industry-specific extensions to the taxonomy, such as for healthcare, banking, etc. In essence, this taxonomy provides a generally accepted way of categorizing and reporting IT costs and other metrics. PBM refers to the practice of managing a collection of projects and programs within an organization, ensuring alignment with the overall business strategy and maximizing their collective value by allocating resources efficiently. The FinOps Foundation has indicated that over time it will include elements from ITFM, TBM, and PBM.

Currently, for FinOps, there is no benchmark that fits the definition of benchmark that we are using in this paper. However, over the years, the FinOps Foundation \cite{finopsbench} has compiled several KPIs that can form the basis for use cases and scenarios for a FinOps benchmark. 
Current FinOps Foundation KPIs include:
\begin{itemize}[left=0pt, topsep=0pt, partopsep=0pt, itemsep=0pt, parsep=0pt]
\item Usage or Spend Apportionment Validation
\item Total Unpredicted Variance of Spend
\item Percent of Compute Spend Covered by Commitment Discounts
\item Effective Savings Rate Percentage 
\item Percentage of Commitment Discount Waste
\item Percent of Unused Resources
\item Auto-scaling Efficiency Rate 
\item Forecast Accuracy Rate (Usage, Spend) 
\item Percentage of Unallocated Shared CSP Cloud Cost
\item Percentage Variance of Budgeted vs. Forecasted CSP Cloud Spend
\item Percentage of CSP Cloud Costs that are Tagging Policy Compliant
\item Percent Storage on Frequent Access Tier  
\item Percentage of Legacy Resource
\end{itemize}

With the advent of the cloud, the academic and industrial research communities have also been active in investigating ways to optimize costs while balancing multiple objectives.
Recent works in the space of FinOps have focused on applying machine learning and mathematical optimization techniques \cite{containerfinops, OptITfinops} to better serve customers' cloud infrastructure needs while offering them insights and recommendations on how they could optimize their overall cloud spend. \cite{snapefinops} addresses the issue of helping customers make trade-offs between cost and resource availability in the presence of offerings such as spot VMs which are cheaper than on-demand VMs but have reduced availability. They propose a framework that uses constrained reinforcement learning to optimize cost and availability by identifying an optimal mix of on-demand VMs and spot VMs. %
Papers such as \cite{Resourcescalingfinops, forecastingfinops, autoscalingfinops} propose forecasting algorithms to scale cloud resources for service level objectives, contributing to the broader field of FinOps-driven cost optimization.
\cite{Resourcecostoptfinops} uses anomaly detection, machine learning, and particle swarm optimization to achieve a cost-optimal cloud resource configuration. \cite{Costoptfinops} analyze the process of using cloud storage to explore opportunities, motivations, and challenges of cost optimization from user perspectives. \cite{Inventoryfinops} focuses on finding the optimal combination of on-demand and reserved instances, such that the demand is satisfied and the costs minimized. They model this optimization problem as a stochastic inventory control problem. 

\cite{CCOfinops} introduces a scalable cost optimizer that determines the most cost-effective deployment strategy for workloads on public or hybrid clouds, considering resource requirements and constraints to minimize costs.
In FinOps, there is an urgent need to move beyond comparative scorecards and broad taxonomies to specific use cases that test the ability of automated agents to optimize IT investments and reduce resource waste. %
To our knowledge, no benchmarks exist for use cases like forecasting, anomaly detection, or cost optimization, nor are there standardized methods to evaluate these techniques with or without agentic support. We are confident that \bench will unite research and development communities to tackle real-world problems through the power of AI and optimization.

%% file: appx/02_bench_details.tex
\section{\bench}
\label{appx:framework}

\bench framework, as shown in \Cref{fig:IT Agent Bench Leaderboard workflow}, supports two main phases corresponding to two personas as follows: (i) \textbf{benchmark registration} phase, where the target is the Benchmark Submitter persona, and (ii) \textbf{agent registration} phase, focusing on the Agent Submitter persona and the actual runtime benchmarking execution and evaluation.

\subsection{Benchmark Registration} 
This phase comprises two main steps: (i) scenario development and registration, and (ii) tasks and evaluation metrics registration.

\textbf{Scenario Development and Registration}

Our scenarios are designed to instantiate real-world IT problems in realistic and manageable environments. Each scenario comprises of two core components: (i) an environment specification, and (ii) a scenario specification metadata. 
The Benchmark Submitter persona then registers these scenarios with \bench, which stores them in its database. 
Each scenario is described using the metadata shown in \cref{ss:bench:tab:scenario_fields}.

\begin{table}[htbp]
\centering
\begin{threeparttable}
\centering
\caption{Scenario Metadata and Examples.}
\label{ss:bench:tab:scenario_fields}
\begin{tabular}{@{}lp{5cm}}
\toprule
\textbf{Field} & \textbf{Example} \\
\midrule
Type & CISO, SRE, FinOps \\
Name & For CISO: k8s CIS-b Minimize containers w/ shared net namespace \\
Description & For CISO: Minimize the admission of containers wishing to share the host network namespace \\
Complexity & Easy, Medium, Hard \\
Class &  
                  For CISO, this is defined based on the technology (e.g., k8s w/ Kyverno; k8s w/ OPA; Rhel9 w/ OPA). 
                  \\
\bottomrule
\end{tabular}
\end{threeparttable}
\end{table}

\textbf{Tasks and Evaluation Metrics Registration} For each scenario type, the Benchmark Submitter registers a well-defined set of tasks that form the basis for the Agent performance evaluation. \Cref{tab:bench_scenarios} summarizes the \bench currently supported IT automation tasks. Moving forward, we plan to extend \bench to incorporate additional tasks (e.g., threat analysis and resource optimization) and to broaden its applicability to other domains (e.g., DevOps).

\subsection{Agent Registration}
During this phase, the Agent Submitter first registers as a user on the platform, then follows with the Agent Registration. 

\subsubsection{Agent Registration}
During Agent Registration, the Agent Submitter specifies the agent metadata as shown in \Cref{ss:bench:tab:agent_fields}.

\begin{table}[htbp]
\begin{threeparttable}
\centering
\caption{Agent Metadata and Examples.}
\label{ss:bench:tab:agent_fields}
\begin{tabular}{@{}lp{4cm}}
\toprule
\textbf{Field} & \textbf{Example} \\
\midrule
Agent Name & -- \\
Agent Type (predefined) & CISO, SRE, FinOps \ldots \\
Agent Level & Beginner, Intermediate, Expert \\
            & (maps to scenario complexity: Easy, Medium, Hard) \\
Scenario Class    %
                  & For CISO: rhel9 w/ OPA; Kubernetes w/ Kyverno; Kubernetes w/ OPA, Kyverno update \\
\end{tabular}
\end{threeparttable}
\end{table}

Once the agent has been registered, the Agent Submitter selects the agent, and the corresponding benchmarks are retrieved from the database using the \textit{agent\_type, agent\_level}, and \textit{scenario\_class} specified during registration for the Agent. 
The Agent Submitter subsequently receives the tasks that the agent must complete to meet the designated objective, each of which has pre-defined evaluation metrics.

\subsection{Leaderboard}

Effective benchmarking of IT automation tasks, especially when selecting LLMs tailored to an organization's specific needs, requires consistent tracking and comparison of agent performance. The Leaderboard facilitates this need by offering a predefined, extensible set of performance metrics that provide clear insights into agent performance relative to the evaluation criteria. 

The Leaderboard supports both API and UI interfaces, enabling a streamlined benchmarking workflow. Users must register the agent endpoint via the Leaderboard’s UI or API. The agent can then query the Leaderboard to retrieve and deploy benchmark scenarios before reporting their operational status. The scenarios can be deployed either automatically by the \bench, as described above, in its hosted environment, or manually outside the Leaderboard, in the user's hosted environment, in which case both agent and environment can still leverage the same Leaderboard API endpoint to publish status updates.

\begin{figure*}[ht]
    \centering
    \includegraphics[width=\linewidth]{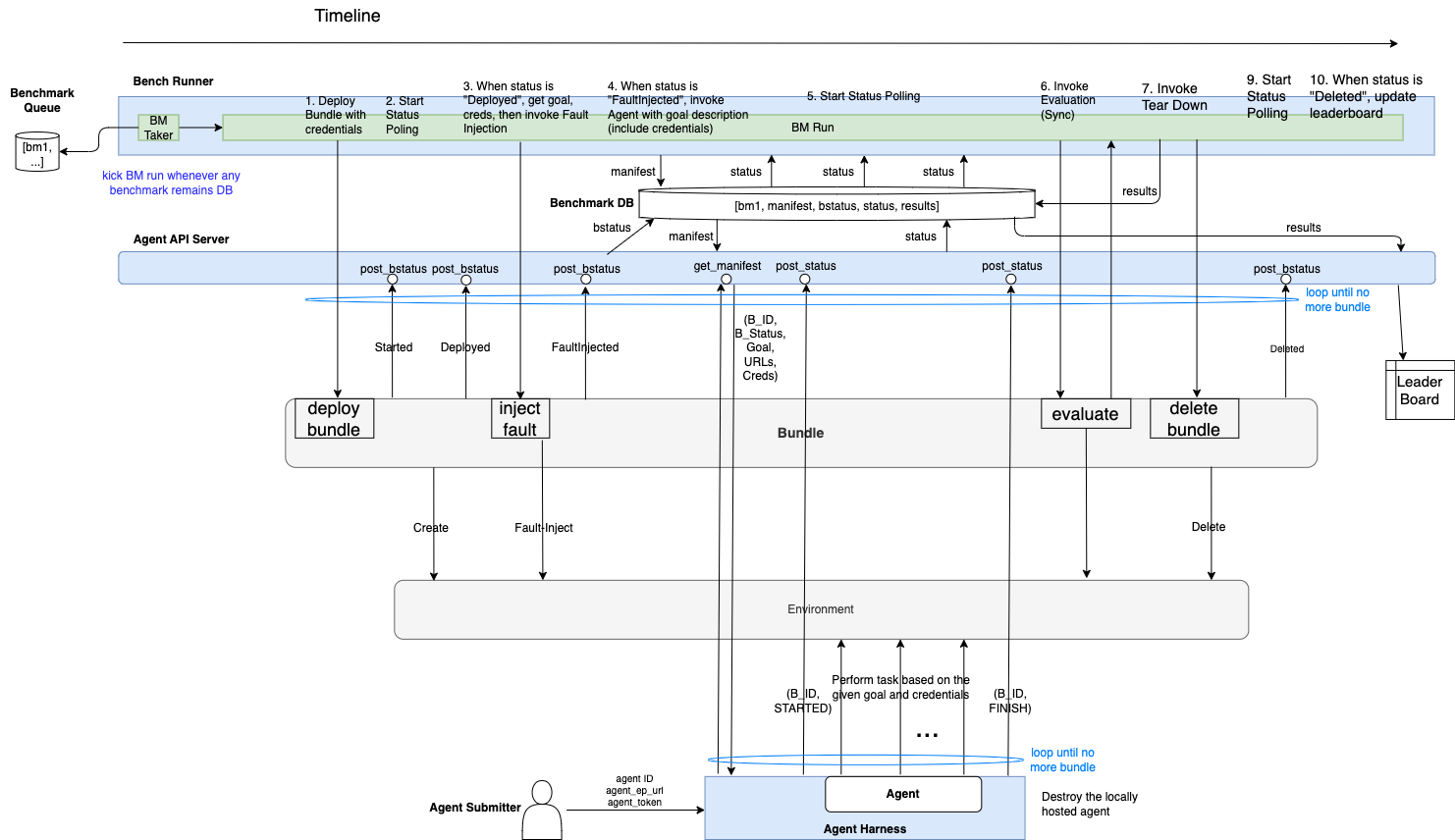}
    \caption{\bench leaderboard workflow.}
    \label{fig:IT Agent Bench Leaderboard workflow}
\end{figure*}

\begin{figure*}[ht]
    \centering
    \includegraphics[width=\linewidth]{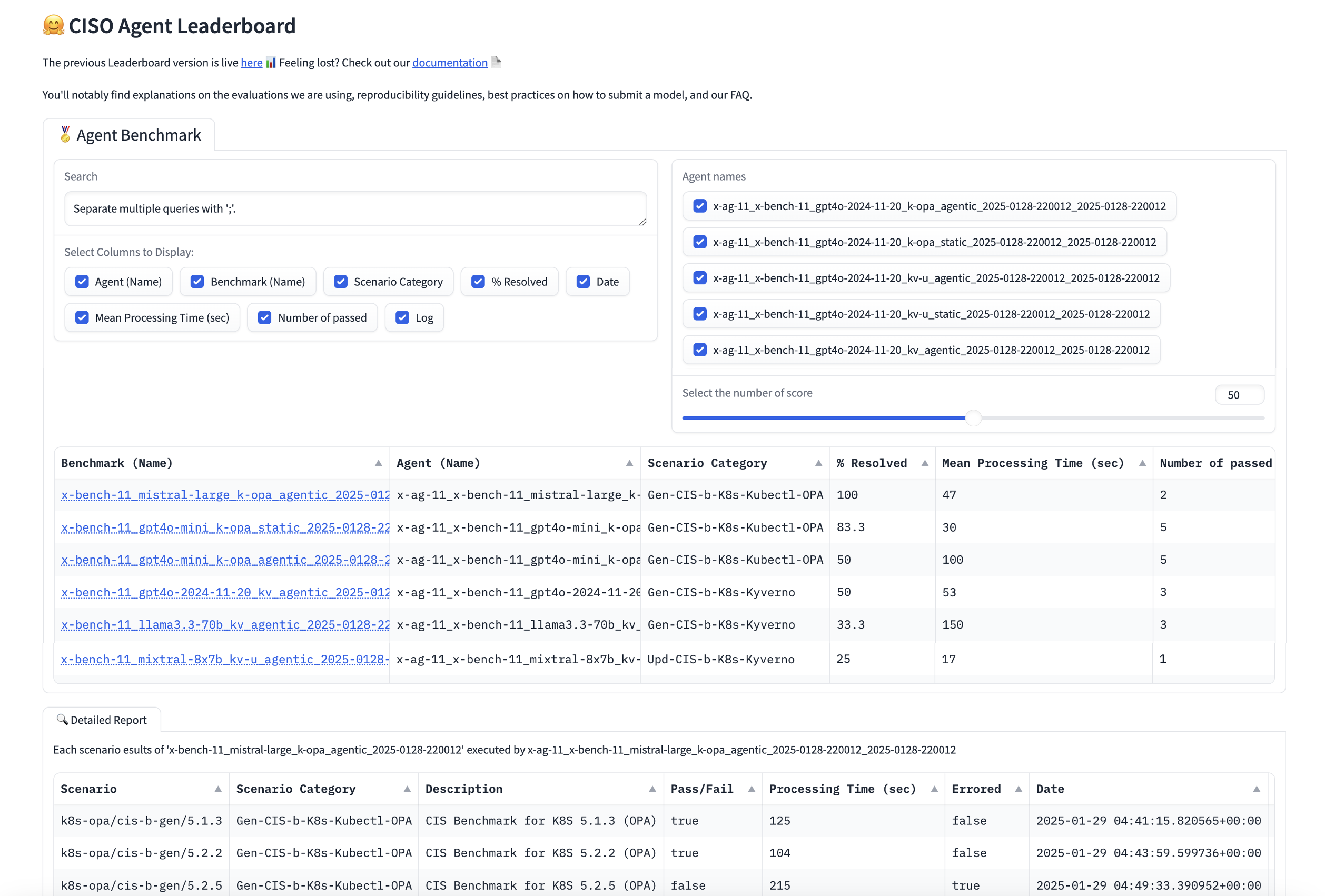}
    \caption{Example \bench leaderboard.}
    \label{fig:ciso-leaderboard}
\end{figure*}

The end-to-end workflow for the agent benchmarking process, after its registration by the Agent Submitter, is illustrated in \Cref{fig:IT Agent Bench Leaderboard workflow}, and summarized in the following.

\begin{enumerate}[left=0pt, topsep=0pt, partopsep=0pt, itemsep=0pt, parsep=0pt]
    \item New benchmark jobs are stored in the Benchmark Queue for processing.
    \item The Benchmark Runner fetches a benchmark scenario for a particular agent from the Benchmark Queue.
    \item The Benchmark Runner provisions the environment as per the benchmark scenario specification.
    The scenario's environment is the set of systems required for the execution of a specific IT task.%
    The Agent interacts with (and can potentially modify) the environment  to solve the given IT automation tasks. A benchmark evaluation measures the Agent's performance based on whether it successfully completes the tasks in the given environment. 
    The environment could be, for instance, a Kubernetes cluster running a target application or a RHEL 9 host with a specific configuration to be validated. 
    The environment is under the direct control of the Agent and therefore may be subject to destructive actions (in case of faulty performance), thus functioning as a sort of ``playground.''
    \item For each scenario included in the benchmark run, the Benchmark Runner and the Agent execute the following steps:
        \begin{enumerate}
            \item The Agent continuously polls the \textbf{get\_manifest} API to monitor when a new manifest enters the \textbf{Ready} state.
            \item Benchmark Runner deploys the scenario's environment by executing the \textbf{deploy\_scenario} function. Each environment reports its status to the Agent API Server using the \textbf{post\_bstatus} API.
            \item The Benchmark Runner monitors the environment's status via the Agent API Server's \textbf{get\_bstatus} API. Once the status becomes \textbf{Deployed}, it injects a fault into the environment by executing the \textbf{inject\_fault} function.
            \item The Benchmark Runner continues to monitor the environment's status using the \textbf{get\_bstatus} API. Once the status reaches \textbf{FaultInjected}, it updates the manifest's status in the Benchmark DB to \textbf{Ready}, including key details such as Benchmark ID, Scenario ID, cluster credentials, and URLs in the manifest. This allows the Agent to access and retrieve this manifest for working with the environment.
            \item Once the manifest status is \textbf{Ready}, the Agent retrieves it. The manifest contains URLs and credentials required to launch the Agent. Before starting the Agent, the Agent calls the \textbf{post\_status} API of the Agent API Server to report its status as \textbf{STARTED}.
            \item After the Agent completes its execution, the \textbf{post\_status} API is called again to report the Agent’s completion its status as \textbf{FINISH}.
            \item Benchmark Runner starts the evaluation and executes the \textbf{delete\_scenario} function.
        \end{enumerate}
    \item Once the evaluation results for all the scenarios in the benchmark are ready, Benchmark Runner aggregates them and publishes the results to the Leaderboard.
\end{enumerate}

We instantiated the Leaderboard evaluation metrics for a few IT automation tasks as detailed in \Cref{sec:scenarioenv}, \Cref{tab:bench_scenarios}. 
In \Cref{fig:ciso-leaderboard} shows the Leaderboard landing page displaying the benchmarking metrics and results for the CISO compliance assessment agent.

%% file: appx/usecases/SRE/main.tex
\section{Site Reliability Engineering}
\label{appx:sre}

\subsection{Background}
\label{ss:sre-background}
With the unprecedented growth in scale and complexity of modern IT systems and infrastructures, failures are the norm instead of exceptions~\cite{patterson:02,gunawi:16,kendrick:12,dimartinoDSN2014,veeraraghavan:osdi:18,Liu:2019,Ghosh:22}. First, {\it hardware failures} are frequent in large-scale IT infrastructures. For example, a new cluster at Google undergoes about a thousand individual machine failures and thousands of disk failures every year~\cite{Dean:2009}. Many of these failures further trigger correlated failures~\cite{ford:10}. New hardware fault models such as silent data corruptions in compute units~\cite{Hochschild:2021} and fail-slow storage~\cite{Gunawi:2018} further increase the challenges of detection and mitigation. In fact, in geo-distributed hyperscalar infrastructures, datacenter-level disasters are no longer rare events~\cite{veeraraghavan:osdi:18}.

Moreover, high velocity of software changes, {\it software failures} caused by code bugs~\cite{Gunawi:14} and misconfigurations~\cite{xu:13} have also become a major cause of IT system failures and service outages, significantly outnumbering hardware failures in recent years~\cite{maurer:15,Barroso:2018}.
For example, IT systems undertake hundreds to thousands of configuration changes daily, which introduces misconfigurations and triggers latent bugs~\cite{sun:osdi:20,tang:15}.
Recent trends in software architectures such as microservices and serverless computing~\cite{Jonas:19} are further enlarging IT reliability challenges by magnifying system complexity and dynamics with sophisticated interactions~\cite{tang:eurosys:23} and emergent behavior~\cite{Huang:2022}. 

The goal of Site Reliability Engineering (SRE) is to achieve high availability and serviceability of IT systems, in the presence of the aforementioned failures~\cite{srehandbook}. The essential job of SRE is failure management\footnote{\scriptsize We follow the classic Fault-Error-Failure model~\cite{aviz:04}, where a {\it fault} is a root cause such as a software bug, a hardware malfunction, or a misconfiguration. A fault can produce abnormal behaviors referred to as {\it errors}. However, some of these errors are transient and have no system-level effect. Only errors that propagate and become observable manifest as {\it failures}, such as crash, hang, incorrect result, or incomplete functionality, etc.}---detecting, diagnosing, and mitigating failures in production systems to prevent production {\it incidents} (the failures that cause user-perceived impacts) or to minimize the impacts and damages of incidents when incident {\it alerts} are triggered. Specifically:

\vspace{-10pt}
\begin{packed_itemize}
    \item {\bf Detection.} SRE must promptly detect production failures via logs, traces, and other telemetry data; detecting failures is the first step to prevent incidents or at least minimize their blast radius and impacts.  
    \item {\bf Diagnosis.} SRE must analyze the root causes of detected failures and localize the faults (e.g., the faulty component and the condition that triggers the fault). 
    \item {\bf Mitigation.} SRE must mitigate the failures to prevent propagation that leads to larger failures or incidents. Mitigation typically follows a resolution plan outlining a sequence of actions to restore the system to its expected state~\cite{chen:eurosys:24}.
\end{packed_itemize}

\bench currently focuses on diagnosis and mitigation tasks with plans to include more tasks 
    such as incident detection, prevention of similar failures/incidents (e.g., by regression testing).

Detection is simplified with golden-signal-based alerts, which observability tools provide natively. Though, the challenge intensifies during an event storm, requiring SREs to distinguish actionable alerts by suppressing false positives and prioritizing those that demand immediate attention — a daily struggle in incident resolution.
Both of these tasks are included in \bench by injecting multiple faults within certain scenarios, causing a flood of alerts. 
The agent must then determine which alerts to prioritize and in what order.

{\bf Urgent need of SRE automation.}
Currently, SRE is largely a human-based practice---SRE engineers are at the forefront of detecting, diagnosing, and mitigating failures and incidents daily~\cite{Beyer:2018,srehandbook}. 
However, IT systems are growing in scale and demand beyond what
human-based practice can reliably, continuously, and efficiently manage, and the cost of human resources and the limit of human reasoning has already become the bottleneck of failure and incident resolution. Today, SRE for IT systems has already become the major TCO (Total Cost of Ownership) of any cloud and software companies~\cite{Boulton2019, IDCStormClouds2024}.
\textit{Hence, SRE automation is no longer an optional enhancement, but an operational imperative.}

In fact, today's IT systems are already increasingly managed by operation programs that automate labor-intensive, human-based operations, known as {\it IT automation}. For example, 
modern cloud management platforms like Kubernetes~\cite{Burns:2016}, 
    Twine~\cite{Tang:2020}, and ECS~\cite{ecs} implement {\it operator} programs to automate a wide variety of operations such as software upgrades, configuration, autoscaling, etc.
However, so far, SRE has not yet become a common part of IT automation due to fundamental challenges of failure managements.

\subsection{Real-world Incident Example}
\label{ss:sre-tasks}
\Cref{tab:incident_details} shows a real-world incident report based on SREs' raw work notes. 
In this incident, SREs were notified of several alerts of type~\textemdash~high error rate (>1\% in last 10 minutes) on a service \textemdash~by Slack. 
The \textit{fault} occurred due to a ``node failure'' due to the accidental deletion of resources during a decommissioning process aimed at cutting IT costs.
The fault caused shard unavailability, leading to an Elasticsearch failure and an SLO violation due to the error rate SLI.
The fault propagated to cause unavailability of shards which in turn led to elasticsearch failure. 
The unavailability of elasticsearch caused SLO violation of error rate SLI.

The \textit{Ops resolution plan} included trial and error to finally arrive at a state which allowed SRE personnel to execute existing mitigation playbook\footnote{Playbook is a structured set of predefined procedures or automated scripts that outline the steps required to perform specific operational tasks or respond to incidents. Playbooks standardize responses, reduce errors, and enable automation of repetitive tasks, enhancing efficiency and reliability in IT operations.}.

As shown, such incidents provide valuable information in terms of:
\begin{inparaenum}[(i)]
    \item time to detection, diagnosis (post detection) and mitigation (post diagnosis), 
    \item symptoms and customer impact, 
    \item faulty condition, fault propagation path and depth,
    \item operation resolution plan, and 
    \item long term fix and improvements. 
\end{inparaenum}
Such real world insights into fault occurrence, propagation, and resolution are invaluable for fault prevention, and automating incident handling. 

\begin{table*}

    \centering
    \begin{threeparttable}
        \caption{An incident that occurred on a SaaS data platform. This incident shows the complex relationship between SRE and FinOps persona, as FinOps ensures that IT environment is cost optimized to meet the financial efficiency goals, while SREs focus is on minimizing service impact and resolving the issue.}
        \label{tab:incident_details}
        \begin{tabular}{m{0.3\textwidth}m{0.65\textwidth}}
            \toprule
            \textbf{Incident} & \textbf{Details} \\
            \midrule
            Triggering alert & Seven alerts of type - ``High error rate on service."\\
            Summary & Error was encountered due to unexpected node failures and EBS volume issues during the downscaling of the Elasticsearch (ES) cluster because of a human error. Downscaling of ES was initiated to save AWS costs associated with running the service.\\
            Incident duration & 180 minutes \\
            Time to detection & 60 minutes \\
            Time to diagnosis & 60 minutes \\
            Time to mitigate & 120 minutes \\
            Symptoms &  [\cmark] Traffic: $\downarrow$, [\cmark] Error: $\uparrow$, [\xmark] Saturation, [\xmark] Latency \\
            Customer impact & Yes.\\ %
            Fault propagation depth & six \\
            Fault propagation & \begin{itemize}[left=0pt, topsep=0pt, partopsep=0pt, itemsep=0pt, parsep=0pt]
                \item [$\downarrow$] Human error: accidental removal of healthy nodes during decommissioning process (maintenance window)
                \item [$\downarrow$] Primary failed while replica initializing (human extrapolation based on the context and manual validation)
                \item [$\downarrow$] Shard assignments failed (ES event: shard unassigned)
                \item [$\downarrow$] Elasticsearch became unhealthy (ES event: RED status)
                \item [$\downarrow$] Services unable to get data from ES (trace)
                \item [$\downarrow$] Increase in error rate on 7 services (events)
            \end{itemize} \\
            Faults & human error, failure during recovery \\
            Resolution plan & \includegraphics[width=0.3\textwidth]{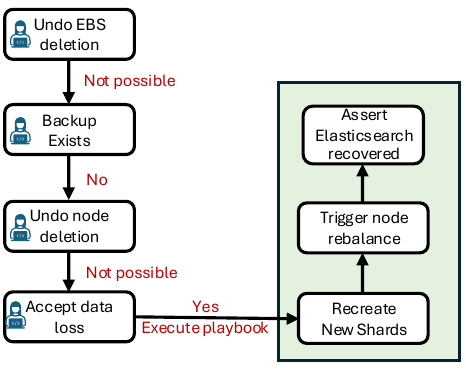}\\
            Resol. plan size & 5 (4 human + 1 automation via playbooks) \\
            Long term improvements & \begin{itemize}[left=0pt, topsep=0pt, partopsep=0pt, itemsep=0pt, parsep=0pt]
                \item [$\checkmark$] Maintain 24-hour gap between instance deletion and EBS deletion
                \item [$\checkmark$] Runbooks updated accordingly
            \end{itemize} \\
            \bottomrule
        \end{tabular}
    \end{threeparttable}
\end{table*}

\begin{table*}[h]
    \centering
    \begin{threeparttable}
        \caption{SRE tasks}
        \label{tab:bench_sre_tasks}
        \begin{tabular}{p{0.39\textwidth}p{0.55\textwidth}}
            \toprule
            \textbf{Task} & \textbf{Task Description} \\
            \midrule
            \textbf{Fault localization} & Identify the faulty entity (root cause) and fault condition. \\
            \textbf{Fault propagation analysis \mbox{(aka root cause analysis)}} & Identifying the causal chain from the root cause entity to the alert, including the identification of fault condition at each step of the chain. \\
            \textbf{Recommend mitigation actions} & Identifying corrective actions to resolve the incident (excluding the execution). \\
            \textbf{Mitigate incident} & Executing corrective actions to clear the alert. \\
            \bottomrule
        \end{tabular}
    \end{threeparttable}
\end{table*}

\subsection{\bench Architecture}
\label{ss:sre-bench}
\begin{figure*}[t!]
    \centering
    \includegraphics[width=0.8\linewidth]{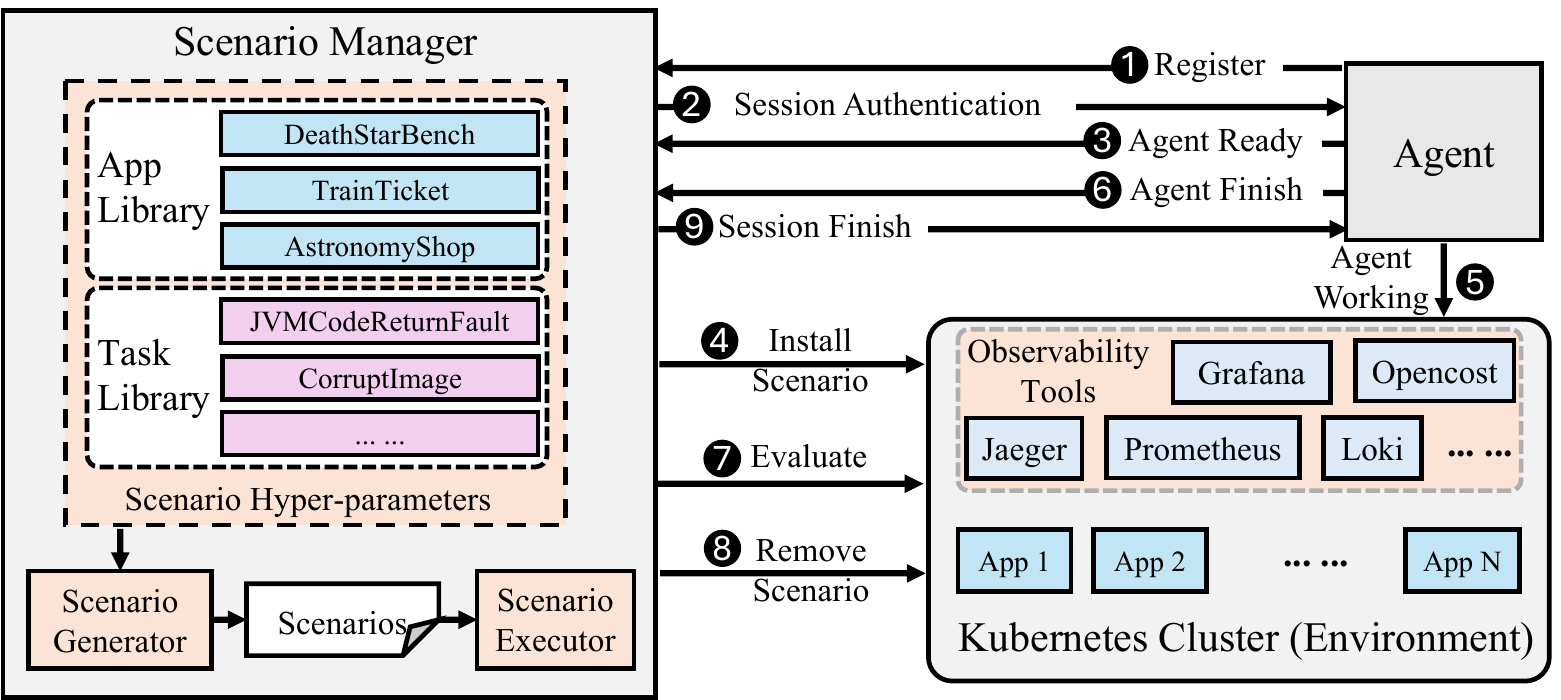}
    \caption{Architecture of \bench responsible for orchestrating SRE  scenarios.}
    \label{fig:bench_design}
\end{figure*}

\bench uses open source technologies to create completely repeatable and reproducible incidents (scenarios) on a Kubernetes platform as shown in \Cref{fig:bench_design}. 

\textbf{Orchestration.}
The core workflow involves a sequence of interactions between the \lumyn\footnote{Henceforth, we will refer to the agents handling SRE tasks as \lumyn} and various components of \bench. 
Initially, \lumyn (\circled{1}) enrolls in the benchmark leaderboard by sending the \texttt{enroll} command, which prompts the \bench to create a session (\circled{2}) and provide necessary credentials and details (e.g., Kubernetes access, time limits). 
Once ready, the agent sends the \texttt{ready} signal (\circled{3}), triggering the scenario executor to install a selected scenario from the scenarios database. 
This specification is used to set up the environment and  inject the fault, including installation of the observability tools (\circled{4}).
During the active phase (\circled{5}), the agent interacts with the environment using tools like \textit{NL2Alerts}, \textit{NL2Logs}, \textit{NL2Metrics}, and \textit{NL2Traces} to complete the task.
Upon task completion or time expiration (\circled{5}), \lumyn sends the \texttt{finish} command (\circled{6}), signaling \bench to evaluate the provided outputs and clean up the environment. 
The scenario executor validates the work of \lumyn (\circled{7}) restores the system to its baseline state (\circled{8}).
The interaction \circled{3} \textemdash \circled{8} continues until scenario manager sends session finish signal (\circled{9}). 

\subsubsection{Principles}
Following the bench principles indicated in the introduction, our \bench uses open-source technologies to construct completely repeatable and reproducible scenarios to simulate real-world incidents. 
\begin{itemize}[left=0pt, topsep=0pt, partopsep=0pt, itemsep=0pt, parsep=0pt]
    \item \textbf{Mimic SRE Best Practices.} \bench follows the guidelines outlined in SRE handbook~\cite{srehandbook} such as alerting on golden signals per application and enabling monitoring and observability. 
    Hence, in our current version, the detection is provided out-of-the-box using the approach outlined in ~\cite{srehandbook}.
    \item \textbf{Mimic Real-world Incidents.} 
We systematically examined 105 real-world incidents from our SaaS products to derive relevant incident patterns. 
Although we integrated several of these patterns into our \bench scenarios, not all were included due to the complexities of accurately reproducing these incidents and mirroring production-level characteristics. Nevertheless, our \bench will continuously evolve through the ongoing incorporation of additional incident patterns.
At the time of writing this paper, \bench supports
    24\% Easy, 24\% Medium, and 52\% Hard incidents, as shown in \Cref{fig:ss-bench-sre-task-complexity}.

    \item \textbf{Provide Observability.} 
    In real-world scenarios, SREs use observability tools alongside command-line access to monitor systems. These tools provide multiple data modalities such as traces, logs, metrics, and events—and support alerting for efficient anomaly detection, trend analysis, and automated troubleshooting. 
    \bench defaults to Grafana~\cite{grafana} but can support other tools including IBM Instana~\cite{instana}, Dynatrace~\cite{dynatrace}, and Datadog~\cite{datadog}. 
    \item \textbf{Model Data Variability.} 
    Depending on system criticality and budget, some data modalities may be missing; for instance, only about 20\% of applications have tracing enabled, complicating incident diagnosis. 
    \bench allows flexible control to enable, disable, or partially enable data modalities as needed.
    \item \textbf{{Manage} Scalability} Scenario hyperparameters consists of (i) environment specification and (ii) scenario specification. 
    Environment specification allows (i) application selection and their related infrastructure selection (e.g., replica count), and data censoring parameters.
    Scenario specification allows selection of hyper parameters (e.g., service name on which to inject fault on). 
    \bench creates a database of scenarios offline using the aforementioned hyper parameters.
    \item \textbf{{Ensure} Determinism.} \bench ensures that alerts are generated according to the scenario specifications before making the scenarios available in \bench. Moreover, \bench ensures that all the assertions (e.g., application is running correctly, alerts are fired correctly) are passed before sending the `READY' state signal to the agent.  
\end{itemize}

\subsubsection{Recreating Incidents in \bench using Real-world Scenarios}
By leveraging detailed incident reports from real-world outages, such as the one summarized in \Cref{tab:incident_details}, we systematically reconstruct similar failure scenarios in \bench. 
As outlined in \Cref{tab:testbed_setup}, this involves configuring a multi-node Elasticsearch cluster with EBS volumes and introducing targeted disruptions—ranging from altering network configurations (e.g., changing ports or IPs) to simulating node and volume deletions, or disabling write operations on specific shards. 
Each recreated scenario is designed to mirror the complexity of the observed production failures with small variations, including similar failure propagation paths, impact on metrics (such as error rates and latency), and the associated operational mitigation steps. 
This ensures that \bench incidents (scenarios) in \bench accurately replicate real-world technical details while also capturing the associated decision-making challenges, allowing for a realistic and representative evaluation of agents.

\begin{table*}[t!]
    \centering
    \begin{threeparttable}
        \caption{Recreated failure scenarios using the incident description described in \cref{tab:incident_details}.}
        \label{tab:testbed_setup}
        \begin{tabularx}{\linewidth}{lXXX}
            \toprule
            \multicolumn{4}{c}{\textbf{Testbed Setup}} \\
            \midrule
            \multicolumn{4}{p{\linewidth}}{
            \begin{itemize}[left=0pt, topsep=0pt, partopsep=0pt, itemsep=0pt, parsep=0pt]
                \item Develop an application that uses Elasticsearch for data storage and retrieval.
                \item A minimum of 3 nodes in the Elasticsearch cluster must be configured.
                \item Attach EBS volumes to each node to simulate the volume usage conditions as in the incident.
                \item Create an index with a sufficient number of documents to stress the system.
            \end{itemize}
            } \\
            \midrule
             & \textbf{Incident Scenario 1} & \textbf{Incident Scenario 2} & \textbf{Incident Scenario 3} \\
            \midrule
            Description &
            Make ES unavailable by changing port, IP address, etc. &
            (i) Identify a victim node: choose one of the nodes within the cluster and delete it, and (ii) delete the attached EBS volume. &
            Identify a victim shard and make it read-only (i.e., disable writes). \\
          
            Fault propagation &
            IP/Port changed $\rightarrow$ ES unavailable $\rightarrow$ Increased error rate in app &
            Similar to incident described in \cref{tab:incident_details} &
            Similar to incident described in \cref{tab:incident_details}, except caused by hardware failure \\
      
            Ops mitigation plan &
            Change the IP address/port to the correct value &
            Similar to incident described in \cref{tab:incident_details} &
            (i) Enable writes on the victim shard, or (ii) follow the procedure similar to incident described in \cref{tab:incident_details} \\
            \bottomrule
        \end{tabularx}
    \end{threeparttable}
\end{table*}

\subsection{Characterizing \bench incidents}
\label{ss:sre:scenario_complexity}

\Cref{tab:bench_sre_task_scenarios} summarizes the scenarios that are currently available in \bench. Beyond these 21 scenarios, \bench can easily produce a far larger range of fault patterns by parameterizing key dimensions such as the target application, the precise location of fault injection, and the number and types of concurrent faults. For instance, if the target application is HotelReservation, the fault of the PodFailure scenario alone can be applied to any of the 18 pods, effectively extending to another 18 scenarios. In this way, \bench can be used to systematically generate hundreds or even thousands of variations. In our evaluation, we focus on representative scenarios, while still enabling users to customize and scale their tests. 
\begin{table*}[htbp!]
\centering
    \centering
    \begin{threeparttable}
        \caption{Unique Scenarios available in \bench.}
        \label{tab:bench_sre_task_scenarios}
        \begin{tabular}{llcc}
            \toprule
             \textbf{Scenario Pattern} & \textbf{Technologies Impacted} & \textbf{\# Fault Propagation} & \textbf{\# Resolution Steps} \\
            \midrule
             CacheFailure               & nodejs                & 3 & 2 \\
             HighCPU &  Java, nodejs & 3 & 2 \\
             ServiceFailure & Java, nodejs & 4 & 3 \\
             ManualGarbageCollection & Java, nodejs & 3 & 2 \\
             MemoryLeak & python, Node.js, Go & 8 & 6 \\
             CorruptDeployment & Go, Java, nodejs & 8 & 6 \\
             CorruptDeployment & Java, Go, nodejs & 7 & 5 \\
             CorruptDeployment & Go, nodejs & 2 & 1 \\
             NetworkDelay &  Go, python, nodejs & 4 & 1 \\
             PodFault &  Go, nodejs & 2 & 2 \\
             NetworkPartition &  Tonic, Rust, Go, nodejs & 4 & 1 \\
             CorruptImage & Go, nodejs & 3 & 1 \\
             CorruptImage &  nodejs & 2 & 1 \\
             CPUStress &  python, nodejs & 2 & 2 \\
             HTTPRequestBodyTamperFault &  Ruby, Go & 3 & 1 \\
             HTTPRequestAbortFault &  PHP, Go, Tonic, Rust, nodejs & 4 & 1 \\
             HTTPRequestBodyTamperFault &  Ruby, Go, nodejs & 3 & 2 \\
             JVMCodeReturnFault &  Java, nodejs & 3 & 1 \\
             PodFailure &  Java, nodejs & 1 & 1 \\
             IncorrectAuthentication & .NET, Go, nodejs & 2 & 1 \\
             MemoryResourceLimit & Go,  nodejs & 1 & 2 \\
            \bottomrule
        \end{tabular}
    \end{threeparttable}
\end{table*}

\begin{figure*}[t!]
    \begin{subfigure}{0.33\linewidth}
        \centering
        \includegraphics[width=\linewidth]{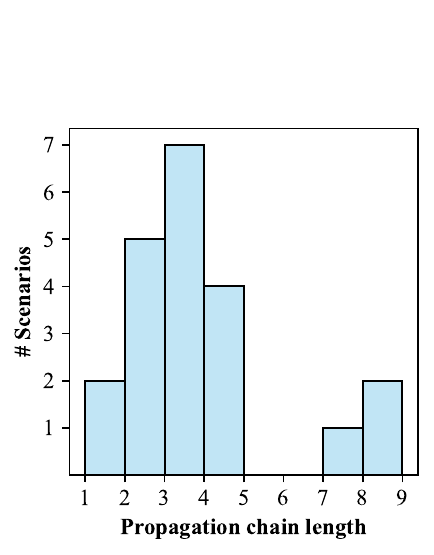}
        \caption{Fault propagation chain length. }
        \label{fig:bench-sre-task-causal-path}
    \end{subfigure}\hfill
    \begin{subfigure}{0.33\linewidth}
        \centering
        \includegraphics[width=\linewidth]{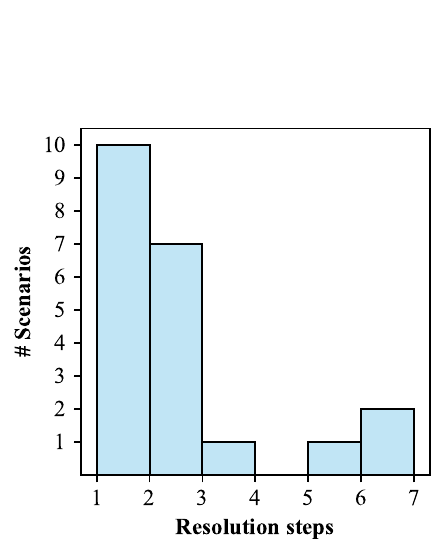}
        \caption{Mitigation plan size.}
        \label{fig:bench-sre-task-resolution-plan-size}
    \end{subfigure}\hfill
    \begin{subfigure}{0.33\linewidth}
        \centering
        \includegraphics[width=\linewidth]{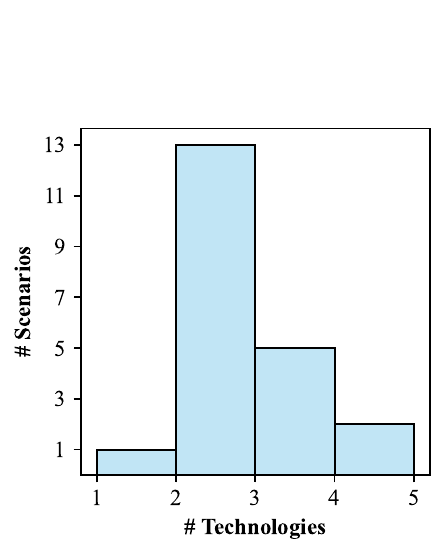}
        \caption{\# Technologies involved.}
        \label{fig:bench-sre-task-num-technologies}
    \end{subfigure}
    \caption{Characterizing \bench scenarios.}
    \label{fig:bench-sre-task-characterization}
\end{figure*}

\Cref{fig:bench-sre-task-characterization} illustrates key incident characteristics observed in our dataset, including the \textit{fault propagation chain length} (\Cref{fig:bench-sre-task-causal-path}), the \textit{resolution plan size} (\Cref{fig:bench-sre-task-resolution-plan-size}), and the \textit{number of distinct technologies} involved (\Cref{fig:bench-sre-task-num-technologies}). Intuitively, as the length of the fault propagation chain grows, the incident becomes more challenging to diagnose. Similarly, a longer resolution plan suggests that restoring service health requires multiple steps and interventions. The involvement of various technologies introduces additional complexity due to the diversity of tools, data sources, and failure modes.

Since \textit{fault propagation length}, \textit{resolution plan size}, and \textit{technology heterogeneity} all influence the difficulty of incident resolution, we define overall task complexity as their geometric mean. \Cref{ss:bench-sre-eq-task-complexity} captures this relationship:

\begin{equation}
\resizebox{\linewidth}{!}{
    $\text{Complexity} = {\sqrt[3]{(\text{propagation path length}\times\text{\# resolution steps}\times\text{\# technologies})}}
    $}
    \label{ss:bench-sre-eq-task-complexity}
\end{equation}

This formulation offers a balanced complexity measure, where the geometric mean ensures that all three factors contribute proportionally, rather than allowing one dominant factor to skew the assessment. 
While factors like required skill sets or the number and type of diagnostic interactions (e.g., tool invocations or queries) could further refine our complexity measure, these factors are often highly dependent on the observability platform, domain expertise, and team-specific processes. 
As discussed, LMs can potentially mitigate skill gaps through targeted fine-tuning and knowledge integration, thereby reducing the variability introduced by differences in human expertise and diagnostic strategies. 
Thus, we focus on the three core factors that are more consistent and inherent to the complexity of the incident itself.

\begin{figure}[t!]
    \centering
    \includegraphics[width=0.45\linewidth]{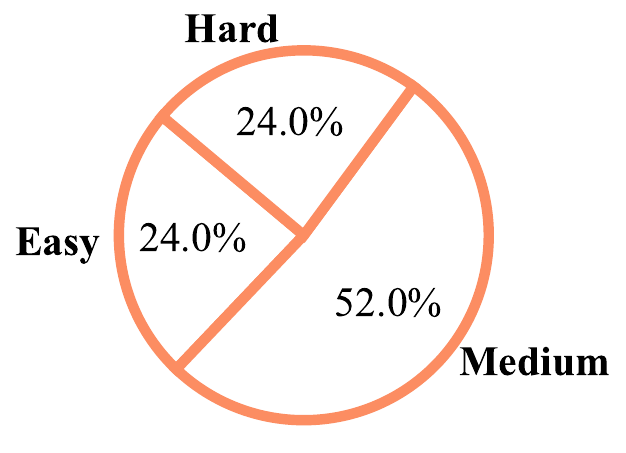}
    \caption{SRE scenario complexity.} 
    \label{fig:ss-bench-sre-task-complexity}
\end{figure}

\Cref{fig:ss-bench-sre-task-complexity} presents the distribution of task complexity values across our incident dataset using the above geometric mean formulation. 
The results show a diverse range of scenarios, with varying degrees of difficulty reflected in the natural interplay among propagation depth, resolution steps, and multi-technology integration. 
This complexity quantification provides a foundation for future analyses, including evaluating how automated reasoning tools, enriched observability stacks, or improved operator training might shift the distribution toward easier, more manageable tasks.

\subsubsection{Experimental Setup}
\label{appx:sre:exp_setup}
These tasks are implemented as Ansible playbooks to benefit from automation pipelines such as Ansible AWX. Below, we present one of our fault injection implementations, which utilizes Kubernetes network policies to simulate port blocking for a target service. We use roles to define different actions related to both fault injection and fault removal respectively. Our fault injections can be reconfigured using the variables to target different services to create additional scenarios. Each scenario has been validated to produce a relevant alert in Grafana, which provides important context to an agent working on a scenario.

\begin{lstlisting}[language=YAML]
---
- name: Define Network Policy to block port 8080
  set_fact:
    network_policy_spec: |
      apiVersion: networking.k8s.io/v1
      kind: NetworkPolicy
      metadata:
        name: "deny-{{ target_service }}-{{ target_port }}" 
        namespace: "{{ target_namespace_project_name }}"
      spec:
        podSelector:
          matchLabels:
            app.kubernetes.io/name: "{{ target_service }}"
        policyTypes:
        - Ingress
        ingress:
        - ports:
          - protocol: TCP
            port: {{ target_port }}
          from: []
  when:
    - is_custom
    - is_fault_injection or is_fault_removal
    - is_network_policy_service_block

- name: Apply Network Policy
  kubernetes.core.k8s:
    kubeconfig: "{{ kubeconfig }}"
    state: present
    definition: "{{ network_policy_spec }}"
  register: network_policy_apply_result
  when:
    - is_custom
    - is_fault_injection
    - is_network_policy_service_block

- name: Remove Network Policy
  kubernetes.core.k8s:
    kubeconfig: "{{ kubeconfig }}"
    state: absent
    api_version: v1
    kind: NetworkPolicy
    name: "deny-{{ target_service }}-{{ target_port }}" 
    namespace: "{{ target_namespace_project_name }}"
  register: network_policy_removal_result
  when:
    - is_custom
    - is_fault_removal
    - is_network_policy_service_block
\end{lstlisting}

For our experiments, we utilized an AWS m4 xlarge cluster configured with 1 control-plane node and 3 worker nodes. The worker nodes had 12 cores and 48 GiB of RAM, with 16 cores and 64 GiB of RAM being used in total. To gain insights into the resource demands imposed by our scenarios, we analyzed the cluster’s performance during a one-hour test period. The key metrics include Persistent Volume Claim (PVC) usage, CPU consumption, and memory utilization, as summarized in \cref{tab:resource-usage}.

\begin{table}[h]
    \centering
    \small
    \begin{threeparttable}
        \caption{Cluster resource usage during fault injection.}
        \begin{tabular}{lccc}
            \toprule
           \textbf{Resources} & \textbf{Usage}  & \textbf{Requests}  & \textbf{Limits} \\ 
 \midrule
CPU       & 2.06571 cores         & 8.19 cores        & 6.16 cores      \\ \hline
Memory    & 13.84 GiB             & 12.89 GiB         & 16.93 GiB       \\ \hline
PVC       & 62.21 GiB             & -                 & 160 GiB \\
\bottomrule 
\end{tabular}
\label{tab:resource-usage}
\end{threeparttable}
\end{table}

\bench also supports experiments on Kind clusters, offering a lightweight and portable option for local testing. We validated this capability on a machine with the following configuration: 1 control-plane node, Intel(R) Xeon(R) Gold 6248 CPU @ 2.50GHz, 12 CPU cores, and 16 GB RAM, running Red Hat Enterprise Linux. This setup allows researchers to efficiently simulate fault scenarios, such as observability stack deployment, OpenTelemetry application deployment, and fault injection tasks, with minimal infrastructure overhead. For example, Incident 22 demonstrated an average CPU usage of 361.71\% and memory consumption of 93.53\%, confirming the feasibility of Kind clusters for reproducible testing.

\subsection{SRE-Agent}
\label{ss:sre-agent}
As described in Section~\ref{sec:baseline-agent}, agents interact with the target environment, collect observability data, and execute action to accomplish its goals. For SRE, the goal is to diagnose and mitigate incidents. Below, we describe the observability data collected by the SRE-Agent and our LM-based, multi-agent system implementation. 

\subsubsection{Observability Data}
\begin{figure}[htbp!]
    \centering
    \includegraphics[width=\linewidth]{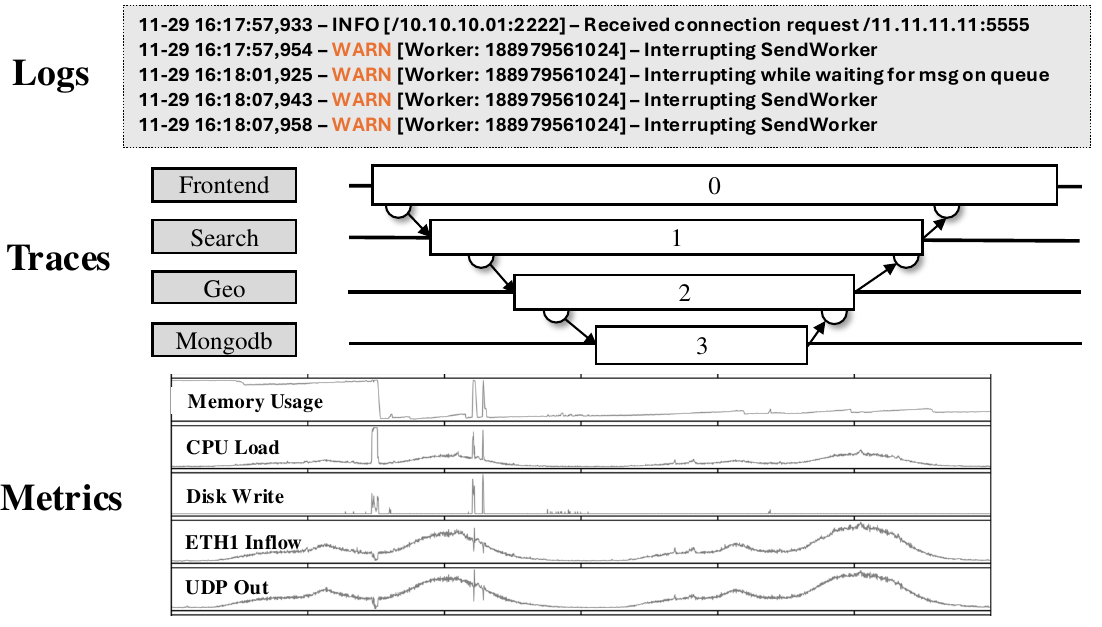}
    \caption{Multi-modality data for SRE task.} 
    \label{fig:ss-bench-sre-multi-modality}
\end{figure} 
As shown in ~\Cref{fig:ss-bench-sre-multi-modality}, SRE tasks involve analyzing multi-modal observability data: logs, traces, and metrics. 

\textbf{Logs.} Logs are semi-structured text records that capture hardware and software events. They are often categorized by severity levels, such as INFO, WARN, and ERROR, to reflect the system's runtime status and the seriousness of its behavior.

\textbf{Traces.} Request traces describe the execution flow of user requests as they traverse through various service instances in a distributed system. They provide a hierarchical representation of service invocations, where each operation is referred to as a span. A span records information about a single service invocation, such as its start time, duration, and associated metadata, including tags and logs. Spans are linked together to form a trace, capturing the complete execution path of the request. Additionally, program exception traces capture program crashes, providing valuable insights for developers during debugging.

\textbf{Metrics.} Metrics provide time-series data monitoring system performance and user-perceived indicators, such as latency, error rates, and resource utilization.

\begin{table*}[t!]
   
    \centering
    \begin{threeparttable}
        \caption{\label{tab:tools}List of the tools used by \lumyn}
        \begin{tabular}{m{0.2\textwidth}m{0.45\textwidth}m{0.10\textwidth}}
            \toprule
            \textbf{Name} & \textbf{Description} & \textbf{Supports Reflection}\\
            \midrule
            NL2Kubectl & Interacts directly with Kubernetes & yes \\
            NL2Traces & Interacts with Grafana API for traces & yes \\
            NL2Metrics & Interacts with Grafana API for fetching metrics stored in Prometheus & yes \\
            NL2Logs & Interacts with Grafana API for fetching logs stored in Loki & yes \\
            NL2Alerts & Interacts with Grafana API for fetching alerts & yes \\
            Mitigation & Generates mitigation plans & no \\
            Wait & Pauses execution for the specified seconds & no \\
            Summarization & Summarizes the input content & no \\
            DiagnosisJsonReport & Generates JSON Report of the diagnosis & no \\
            MitigationJsonReport & Generates JSON Report of the mitigation plan & no \\
            \bottomrule
        \end{tabular}
    \end{threeparttable}
\end{table*}

\subsubsection{\lumyn Architecture and Implementation}
\begin{figure}[h]
    \centering
    \includegraphics[width=\linewidth]{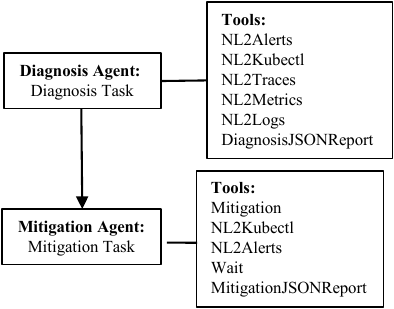}
    \caption{\lumyn architecture}
    \label{fig:agent_arch}
\end{figure}

The \lumyn architecture consists of two LM-based agents, a Diagnosis Agent and a Resolution Agent as shown in~\Cref{fig:agent_arch}. We first define the following basic components used in our implementation:
\begin{itemize}
    \item \textit{Agent.} An agent is an autonomous or semi-autonomous software program that uses a LM to plan, make decisions, interact with the target environment, and execute actions to accomplish goals.
    \item \textit{Task.} A task is a specific goal that the agent must accomplish before its execution terminates. In our implementation, a task is a complex multi-step process (e.g. diagnosing the cause of an incident). Tasks also have tools associated with them that the agent can use to achieve the goal.
    \item \textit{Tool.} A tool is a function or API call that the agent can use to perform a specific sub-task, such as, interact with the target environment to collect observability data. 
\end{itemize}

We now describe our implementation of each of the above components.

\textbf{Tools.} Table~\ref{tab:tools} lists all the tools available to \lumyn. All our tools are also LM-based, where the LM is prompted with an utterance from the agent instructing it to perform the required sub-task. The tools are of two types based on whether they generate natural language (e.g., Mitigation) or function calls (e.g., NL2Kubectl). Further, to potentially improve the accuracy and usability of our tools, we equip them with the following features.

\begin{itemize}
\item \textit{Reflection.} To enable automatic correction of wrong LM responses, they are provided with external feedback~\cite{pan2023automatically, huang2023large} from \textit{linters}. Specifically, for tools that generate function calls, linters are developed to validate the syntax and semantics of the output. If the linter finds a problem with the generated function call, the LM is re-prompted with the linter's feedback so that it can attempt to fix the problem. Similarly, if the generated function call passes linting, but causes an error upon execution, the error message and the failing function call are used to re-prompt the LM for a fixed function call. 

\item \textit{Summarizer.} For some tools, such as NL2Logs and NL2Traces, the output is not directly returned to the agent because it is very long and contains extraneous information. These tools utilize an additional step that prompts a LM with the output and asks it to provide a detailed summary with only relevant information.
\end{itemize}

\textbf{Tasks:} We define the following two tasks to be completed by \lumyn. Each task includes a description of its completion process and the expected output upon completion.  Each task also has tools associated with it that the agent can use to execute sub-tasks, gather information or interact with the environment.

\begin{itemize}
\item  \textit{Diagnosis Task.} For diagnosis, the goal is to identify the entire fault propagation chain, i.e., \textit{fault propagation chain} (FPC) analysis, and identify the exact cause of the problem within the chain, i.e., \textit{fault localization} (FL).

\item \textit{Mitigation Task.} For mitigation, the goal is to provide natural-language mitigation plans, and execute them to successfully clear the triggering alert. The mitigation plans increase agent explanability and help SREs in understanding why the agent executed certain commands.
\end{itemize}

\textbf{Agents.}  Overall, \lumyn consists of two agents, namely, diagnosis and mitigation agents. Each agent is assigned tasks that it must complete. In general, multi-agent systems can be \textit{hierarchical} or \textit{sequential}. Sequential execution allows tasks to be completed in a fixed, linear order. In hierarchical execution, a ``manager'' agent determines the task execution order and co-ordinates with the other agents. We adopt sequential execution because it is well suited for the SRE use case, where an incident must be diagnosed before it can be resolved. Although, the order of task execution is fixed, the sub-tasks or steps within each task may be completed in any order as determined by the agent itself. We describe the overall workflow of both our agents below.
\begin{itemize}
\item \textit{Diagnosis Agent.} First, the Diagnosis Agent uses the NL2Alerts tool to retrieve the active alerts in the environment. The agent then flexibly and iteratively uses observability tools to gather traces, logs, and metrics from the affected entity mentioned in the alert, and entities associated with the affected entity. It may also use NL2Kubectl commands to investigate the environment. Once the agent determines that it has sufficient information to provide a diagnosis, it proceeds to generate a structured diagnosis report in JSON format with its findings to facilitate evaluation. After the report is generated, the Mitigation Agent takes over.  

\item \textit{Mitigation Agent.} The Mitigation Agent ingests the diagnosis report to create mitigation plans and then utilizes the available tools to implement the plan. This involves using NL2Kubectl commands. To ensure that the executed commands mitigated the incident, it can also use the NL2Alerts tools to check whether the alerts in environment have been cleared. Further, since alerts could sometimes temporarily appear to get cleared due to fluctuations in a live environment, the agent can use the Wait tool to check whether the alerts \textit{stay} cleared even after some time. Finally, upon completion of the execution, the agent generates a JSON explaining the mitigation steps that it took.
\end{itemize}

\subsection{\bench Evaluation}
\label{ss:sre-evaluation}

\subsubsection{Experimental details}

We evaluate the \lumyn agent on a set of 42 SRE scenarios in the \bench. For the agent’s LM-based planning component, we consider four distinct models: gpt-4o, granite-3.1-8b-instruct, llama-3.3-70b-instruct, and llama-3.1-8b-instruct. None of these models are fine-tuned.

\Cref{tab:sre-model-param} shows the main hyper-parameter values used in our experiments. These values were chosen to ensure as deterministic results as possible. $decoding\_method$ is applicable for all models except gpt-4o.

\begin{table}[h]
    \centering
    \small
    \begin{threeparttable}
        \caption{Model hyper-parameters.}
        \begin{tabular}{p{3.5cm}l}
            \toprule
           \textbf{Hyper-parameter} &  \textbf{Value} \\ 
 \midrule
$temperature$ & 0 \\
$top\_p$ & 1e-7\\
$seed$ & 42 \\
$decoding\_method$ & greedy \\ 
\bottomrule
\end{tabular}
\label{tab:sre-model-param}
\end{threeparttable}
\end{table}

\begin{table*}[htp]
\centering
\begin{threeparttable}
\caption{Experimental details}
\label{tab:exp-setup-sre}
\begin{tabular}{@{}l c c c c@{}}
\toprule
\multirow{2}{*}{\textbf{Models}} 
 & \multirow{2}{*}{\textbf{Scenarios}}
 & \multicolumn{3}{c}{\textbf{Experiment Setup}} \\
\cmidrule(lr){3-5}
 & 
 & \textbf{\#Repeats} 
 & \textbf{\#Total} 
 & \textbf{\%Agent Submission} \\
\midrule
\textbf{granite-3.1-8B-instruct}         & 42 & 10 & 420 & 98.76\% \\
\textbf{llama-3.1-8B-instruct} & 42 & 10 & 420  & 100.0\% \\
\textbf{llama-3.3-70B-instruct} & 42 & 10 & 420 & 100\% \\
\textbf{gpt-4o}                       & 42 & 10 & 420 & 99.75\% \\
\bottomrule
\end{tabular}
\begin{tablenotes}
\footnotesize
\item \textit{Note:} “\%Agent submission” is the percentage of all trials completed in which the agent returned results. 
\end{tablenotes}
\end{threeparttable}
\end{table*}

\subsubsection{Evaluation Metrics}
\label{sss:eval-metrics}

We evaluate each LM-based agents on two primary tasks: (i) Diagnosis and (ii) Mitigation.

\textbf{Diagnosis.}
The agent is evaluated for diagnosis based on its ability to provide accurate \textit{fault localization} and \textit{fault propagation chains}.
Fault localization allows SREs to identify the exact resource \textit{causing} the problem, whereas fault propagation chain allows SREs to understand how the fault is cascading across the application stack and impacting the application. 
Fault propagation chain can be further used for other important tasks such as blast radius analysis.
\begin{itemize}[left=0pt, topsep=0pt, partopsep=0pt, itemsep=0pt, parsep=0pt]
    \item 	\textit{Fault localization} performance is measured using pass@1 and Normalized Topology-Aware Match (NTAM).
	\item \textit{Fault propagation chain} is assessed with NTAM.
	Additionally, we track Mean Time to Diagnosis (MTTD) to gauge overall diagnostic efficiency.
\end{itemize}

\textbf{Mitigation.}
For mitigation, we evaluate how effectively the agent resolves incidents (i.e., clears alerts).
\begin{itemize}[left=0pt, topsep=0pt, partopsep=0pt, itemsep=0pt, parsep=0pt]
    \item Success rate is quantified using pass@1.
    \item Efficiency is captured through Mean Time to Resolution (MTTR).
\end{itemize}
At the time of writing of this paper, \bench lacks the ability to automatically measure the natural language-based unstructured outputs fault condition (i.e., what is wrong with the identified resource) but have plans to extend to this task using LM-as-a-judge~\cite{zheng2023judgingllmasajudgemtbenchchatbot}.

\subsubsection{Metric definitions}
\label{appx:ntam}
\textbf{pass@1.}
We evaluate both fault localization and mitigation using the pass@1 metric~\cite{chen2021evaluating}, which is defined as follows:
$$\text{pass@}k := \mathbb{E}_{\text{Scenarios}}\left[1 - \frac{\binom{n-c}{k}}{\binom{n}{k}}\right] $$
It is an unbiased estimator of correctness in \textit{k}=1 trials across all scenarios. For \textit{fault localization}, correctness means whether the predicted root cause exactly matches the ground truth root cause. For \textit{mitigation}, correctness means whether the alerts are cleared.

\textbf{Normalized Topology-aware Matching.}
Existing approaches for evaluating \textit{fault propagation chains} and \textit{fault localization} focus on exact matches with  the ground truth ~\cite{ahmed2023recommending, zhu2024hemirca, chen2024automatic}, which overlooks topology and finer-grained analysis of propagation chains. For example, existing approaches cannot effectively differentiate agents and models when predicted propagation chains or root causes do not exactly match the ground truth, as they fail to measure how close the predictions are to the actual faults. %
Hence, we propose a new metric \textit{Normalized Topology-Aware Match} (NTAM), which measures agent performance compared to ground truth via topology-aware distance calculation.

NTAM requires a topology graph, where the nodes are the entities of the system, and edges indicate various types of connections between them (e.g., Deployment owns ReplicaSet). Given such a topology, it can be used to evaluate both the set of entities in the fault propagation chains, and the set of root cause entities for fault localization. NTAM is inspired by topology-based distance metrics and information retrieval concepts, such as BM25~\cite{fang2011diagnostic}, that down-weight less discriminative features. It is a flexible, general function with configurable components for fine-grained evaluation of predicted output quality.%

Specifically, it consists of the following main components:
\begin{itemize}
    \item \textit{Topology-based distance scoring} functions, which consider both the edge-type and sub-tree size, rewarding predicted entities closer to the ground truth. %
    Further, nodes with fewer connections (smaller sub-trees) receive higher scores, as they are more discriminative for fault localization.%
    
    \item A \textit{node importance factor} based on the position of the ground truth entity in the propagation chain. This captures the intuition that predicting the ground truth root-cause entity correctly should be rewarded more than getting another entity on the chain correct.
    \item \textit{Penalization terms for length mismatch} between the predicted and ground-truth entities. This is to ensure that predictions having too many or too few entities get lower scores.
\end{itemize}

All the components have corresponding hyper-parameters that can be tuned to adjust their contributions to the overall score. The final score is normalized to be between 0 and 1, where 1 indicates a perfect match. For fault localization, instead of evaluating the set of all entities, only the ground-truth and predicted root-cause entities are considered.

\textbf{Mean Time to Diagnosis.}
For the scenarios where an agent finishes diagnosis successfully (i.e., root cause entities are found), we calculate \textit{MTTD}, which measures how soon (in seconds) an agent performs diagnosis. Otherwise, \textit{MTTD} is set to infinite. 

\textbf{Mean Time to Repair.}
Similarly, for mitigation, we identify the scenarios where an agent executes an automated action to resolve the faults successfully (i.e., alerts are cleared). For these scenarios, we calculate \textit{MTTR} (in seconds), which measures how soon an agent performs mitigation. Otherwise, \textit{MTTR} is set to infinite. 

\subsubsection{Evaluation Results}

We present evaluation results for four LM-based agents across 42 SRE scenarios in the \bench framework.

\textbf{Overall agent results.}
gpt-4o shows the strongest performance, achieving a 13.81\% pass@1 in diagnosis and 11.43\% pass@1 in mitigation (\Cref{tab:sreagent-eval}), significantly higher than any other agent. Moreover, it also attains the best scores on the NTAM metrics (FL and FPC). 
Notably, in hard scenarios (\Cref{tab:appx:sre:traces}), gpt-4o is the only agent capable of performing multiple accurate diagnosis (granite only succeeded once), and \textit{none of the agents can repair the hard scenarios}.
Meanwhile, llama-3.1-8B, despite having fewer parameters, offers the fastest detection (lowest MTTD of 57.50s) and repair times (lowest MTTR of 245.13seconds) among successful attempts. 
Although granite-3.1-8B shares the same parameter size as llama-3.1-8B, it demonstrates slightly better diagnostic capabilities yet weaker mitigation ability. 
llama-3.3-70B performs second best overall, trailing behind gpt-4o on all the metrics we compute.

\begin{table*}[!ht]
\small
\centering
\caption{Diagnosis pass@1 (in \%).}
\label{tab:sre:diag_pass1}
\begin{tabular}{lccc}
\toprule
\textbf{Model} & \textbf{Easy} & \textbf{Medium} & \textbf{Hard} \\
\midrule
\textbf{gpt-4o}                             & 36.00 $\pm$ 4.73 & 7.73 $\pm$ 1.74 & 5.00 $\pm$ 2.24 \\
\textbf{granite-3.1-8B-instruct}    & 8.00 $\pm$ 2.68  & 2.73 $\pm$ 1.05 & 1.00 $\pm$ 1.03 \\
\textbf{llama-3.1-8B-instruct}      & 1.18 $\pm$ 1.20  & 1.36 $\pm$ 0.79 & 0.00 $\pm$ 0.00 \\
\textbf{llama-3.3-70B-instruct}      & 10.00 $\pm$ 2.93 & 1.36 $\pm$ 0.78 & 0.00 $\pm$ 0.00 \\
\bottomrule
\end{tabular}

\end{table*}

\begin{table*}[!ht]
\small
\centering
\caption{Repair pass@1 (in \%)}
\label{tab:sre:repair_pass1}
\begin{tabular}{lccc}
\toprule
\textbf{Model} & \textbf{Easy} & \textbf{Medium} & \textbf{Hard} \\
\midrule
\textbf{gpt-4o}                             & 21.00 $\pm$ 4.06 & 12.27 $\pm$ 2.19 & 0.00 $\pm$ 0.00 \\
\textbf{granite-3.1-8B-instruct}     & 1.00 $\pm$ 1.01  & 0.00 $\pm$ 0.00  & 0.00 $\pm$ 0.00 \\
\textbf{llama-3.1-8B-instruct}       & 5.88 $\pm$ 2.48  & 1.36 $\pm$ 0.80  & 0.00 $\pm$ 0.00 \\
\textbf{llama-3.3-70B-instruct}      & 7.00 $\pm$ 2.50  & 3.18 $\pm$ 1.16  & 0.00 $\pm$ 0.00 \\
\bottomrule
\end{tabular}

\end{table*}

\begin{figure}[ht]
    \centering
    \includegraphics[width=0.8\linewidth]{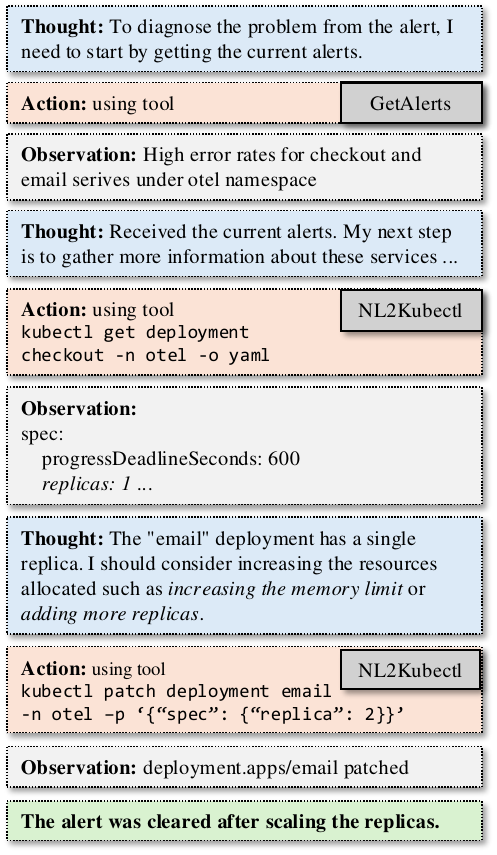}
    \caption{Sample Trajectory of llama-3.3-70b-instruct in Scenario 15} 
    \label{fig:incident-15}
\end{figure} 

\textbf{Result analysis by scenario complexity.} 
As we categorize the benchmark scenarios into Easy, Medium, and Hard levels based on the complexity described in \Cref{ss:bench-sre-eq-task-complexity}, a clear performance gap emerges as \textit{scneario\_complexity} increases. In particular, \Cref{tab:sre:diag_pass1} shows lower diagnosis accuracy (pass@1) in more complex scenarios, and \Cref{tab:sre:repair_pass1} reveals a corresponding drop in mitigation success (pass@1). Among the five hard scenarios, none can be resolved by any agent in any run. By contrast, for easy scenarios, over half (five out of eight) scenarios were successfully repaired by at least one agent, and six were diagnosed correctly. 
We use difference‐of‐proportions z‐test to compare success rates across different task levels (evaluating two levels each time). %
\textit{The agent performance consistently declines from Easy to Hard scenarios, validating our complexity model based on propagation chain length, resolution steps, and technology diversity.}

\textbf{Interdependence between diagnosis and mitigation.} Interestingly, diagnosis and mitigation are often intuitively assumed to be interdependent, with accurate diagnosis serving as a prerequisite for effective mitigation. However, \textit{our findings reveal that: in some scenarios, agents can successfully mitigate an incident despite misidentifying the root cause.} For example, in scenario 15 (\Cref{fig:incident-15}), an agent using the llama-3.3-70b model incorrectly identified the root cause as ``memory limit'' in the service, while the real root cause was HTTP request corruption fault; yet it still managed to resolve the issue by scaling up the email service pods to make it functional, essentially bypassing the handling of actual HTTP fault. Such cases illustrate how generic mitigation actions, such as restarting services or scaling replicas, can sometimes fix the system symptoms even without a fully accurate diagnosis. We observed similar behavior in real-world SRE incident analysis, where, despite the root cause remaining unidentified, SREs were able to mitigate the incident.

Conversely, some scenarios highlight the opposite issue: scenario 13, though labeled as ``easy'', cannot be fixed by any of the tested agents, even though they achieved high scores in diagnosing the root cause. Notably, gpt-4o attained roughly 80\% pass@1 and over 0.7 on both FPC and RC (NTAM) metrics. This implies that, \textit{although the agent cannot fully resolve certain issues, it can still offer near-accurate diagnostic insights, potentially assisting human operators in debugging.} 

\begin{figure}[ht]
    \centering
    \includegraphics[width=0.8\linewidth]{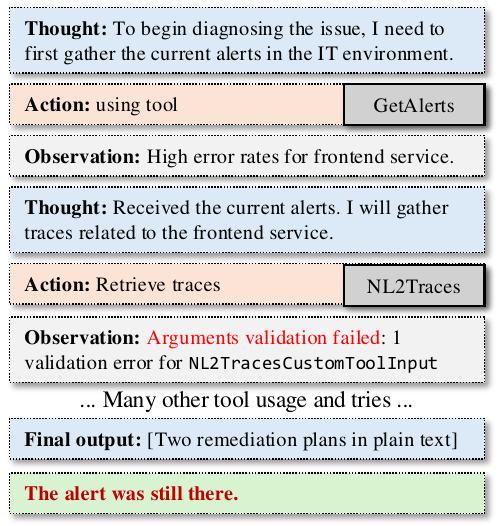}
    \caption{Sample Trajectory of gpt-4o in Scenario 19} 
    \label{fig:incident-19}
\end{figure}

Some scenarios yield even worse outcomes. For example, in scenario 19 (\Cref{fig:incident-19}), the agent fails to identify the root cause and cannot repair the system, offering only a mitigation plan at the end that is entirely ineffective.

\textbf{Inconsistency between runs.} Another crucial observation is the agents' inconsistency across repeated runs. For instance, llama-3.1-8b-instruct and mistral-large-2 in scenario 11 occasionally succeed in only a single run out of 10. Though gpt-4o can reliably repair the scenario 8, its running time can fluctuate between 100 and 800 seconds. \textit{These inconsistencies stem from real-time telemetry fluctuations, where minor changes (e.g., CPU utilization reported at 58\% in one run vs.\ 71\% in another) affect LM outputs, leading to varied diagnostic and mitigation results.}

\textbf{Impact of tracing on accuracy.}
Many benchmarks provide raw telemetry data, a key differentiator of \bench is its alert-driven workflow, which mirrors how SREs are notified of faults through golden-signal-alerts triggered from collected telemetry data. To further assess the importance of different telemetry sources, \textit{\bench also supports automated telemetry data masking.} 
As shown in \Cref{tab:appx:sre:traces} and \Cref{tab:appx:sre:disabled}, gpt-4o sees its diagnosis pass@1 drop from 18.10\% (with traces) to 9.52\% (without traces), and its mitigation pass@1 plummet to 2.86\%. Similarly, llama-3.3-70B experiences its diagnosis rate decline from 5.24\% to 0.95\%.
In fact, only three  scenarios were successfully resolved by gpt-4o once trace data was masked.
Take Scenario 13 (easy level) as an example. The agent is able to achieve an 80\% diagnosis rate in its all runs; however, when masking the traces, the rate drops to 0. Note that all of these telemetry masking and agent evaluation steps are integrated into \bench’s automated pipeline. Agents can be evaluated with different observability configurations in \bench.

\begin{table*}[h]
\small
\centering
\begin{threeparttable}
  \caption{Evaluation of SRE-agent only on scenarios with tracing enabled.}
  \label{tab:appx:sre:traces}
  \begin{tabular}{@{}lcccccc@{}}
    \toprule
    \multirow{2}{*}{\textbf{Models}}
      & \multicolumn{4}{c}{\textbf{Diagnosis}}
      & \multicolumn{2}{c}{\textbf{Mitigation}} \\
    \cmidrule(lr){2-5}\cmidrule(lr){6-7}
    & \textbf{pass@1 (\%)$\uparrow$}
    & \textbf{FL (NTAM)$\uparrow$}
    & \textbf{FPC (NTAM)$\uparrow$}
    & \textbf{MTTD (s)$\downarrow$}
    & \textbf{pass@1 (\%)$\uparrow$}
    & \textbf{MTTR (s)$\downarrow$}\\
    \midrule
    \textbf{granite-3.1-8B-instruct} &
    $3.33 \pm 1.20$ &
    $0.15 \pm 0.02$ &
    $0.16 \pm 0.01$ &
    $341.58 \pm 81.71$ &
    $0.48 \pm 0.50$ &
    $845.50 \pm $ \textemdash \\
    \textbf{llama-3.1-8B-instruct} &
    $0.50 \pm 0.51$ &
    $0.07 \pm 0.01$ &
    $0.08 \pm 0.01$ &
    $58.24 \pm $ \textemdash&
    $2.50 \pm 1.09$ &
    $245.39 \pm 49.45$ \\
    \textbf{llama-3.3-70B-instruct} &
    $5.24 \pm 1.59$ &
    $0.21 \pm 0.02$ &
    $0.22 \pm 0.02$ &
    $155.78 \pm 19.91$ &
    $5.71 \pm 1.60$ &
    $449.50 \pm 46.59$ \\
    \textbf{gpt-4o} &
    $18.10 \pm 2.58$ &
    $0.45 \pm 0.05$ &
    $0.37 \pm 0.03$ &
    $67.53 \pm 3.84$ &
    $20.00 \pm 2.75$ &
    $266.97 \pm 32.95$ \\
    \bottomrule
  \end{tabular}
  \begin{tablenotes}
      \scriptsize 
      \item 21 scenarios, 10 runs per scenario. 
  \end{tablenotes}
\end{threeparttable}
\end{table*}

\begin{table*}[h]
\small
\centering
\begin{threeparttable}
  \caption{Evaluation of SRE-agent only on scenarios in which tracing is disabled.}
  \label{tab:appx:sre:disabled}
  \begin{tabular}{@{}lcccccc@{}}
    \toprule
    \multirow{2}{*}{\textbf{Models}}
      & \multicolumn{4}{c}{\textbf{Diagnosis}}
      & \multicolumn{2}{c}{\textbf{Mitigation}} \\
    \cmidrule(lr){2-5}\cmidrule(lr){6-7}
    & \textbf{pass@1 (\%)$\uparrow$}
    & \textbf{FL (NTAM)$\uparrow$}
    & \textbf{FPC (NTAM)$\uparrow$}
    & \textbf{MTTD (s)$\downarrow$}
    & \textbf{pass@1 (\%)$\uparrow$}
    & \textbf{MTTR (s)$\downarrow$}\\
    \midrule
    \textbf{granite-3.1-8B-instruct} &
    $3.81 \pm 1.30$ &
    $0.18 \pm 0.02$ &
    $0.21 \pm 0.02$ &
    $160.97 \pm 51.06$ &
    $0.00 \pm 0.00$ &
    \textemdash \\
    \textbf{llama-3.1-8B-instruct} &
    $1.46 \pm 0.84$ &
    $0.06 \pm 0.01$ &
    $0.07 \pm 0.01$ &
    $57.26 \pm 2.88$ &
    $1.46 \pm 0.83$ &
    $244.96 \pm 68.53$ \\
    \textbf{llama-3.3-70B-instruct} &
    $0.95 \pm 0.68$ &
    $0.11 \pm 0.02$ &
    $0.10 \pm 0.02$ &
    $430.86 \pm $ \textemdash &
    $0.95 \pm 0.67$ &
    $1429.80 \pm 552.71$ \\
    \textbf{gpt-4o} &
    $9.52 \pm 2.15$ &
    $0.32 \pm 0.05$ &
    $0.31 \pm 0.04$ &
    $85.19 \pm 12.84$ &
    $2.86 \pm 1.12$ &
    $385.87 \pm 15.292$ \\
    \bottomrule
  \end{tabular}
  \begin{tablenotes}
    \footnotesize
    \item 21 scenarios, 10 runs per scenario.
  \end{tablenotes}
\end{threeparttable}
\end{table*}

\textbf{Flexibility and extensibility.} Beyond scenario design, \textit{\bench is designed with flexibility and extensibility as guiding principles, allowing for both the addition of new scenarios within existing tasks and the support of new tasks like resource management.} We have already integrated four distinct applications, including both microservice applications (OpenTelemetry-Demo and Hotel-Reservation from DeathStarBench), and non-microservice (TiDB application and Elasticsearch application). \bench makes it straightforward to incorporate custom applications by simply adding an Ansible playbook. It took around four human hours for the external collaborators to add an application to \bench. Moreover, by \bench introduces realistic faults at multiple system layers (e.g., application, virtualization), ensuring a comprehensive evaluation of agent performance across a wide range of failures.

\begin{figure*}
    \includegraphics[width=\textwidth]{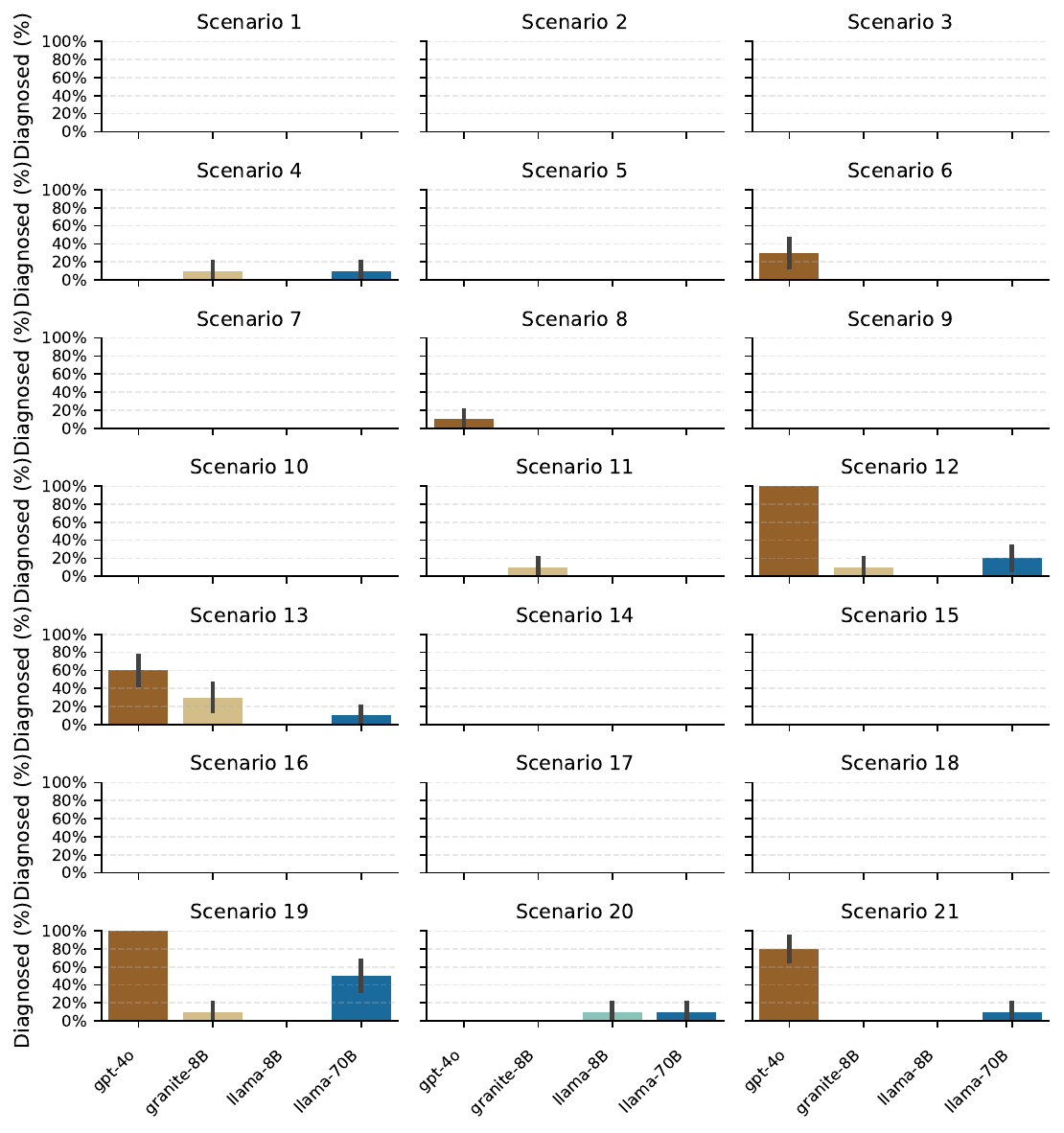}
    \caption{\label{fig:sre:trace_on_diagnosis_pass1} Percent diagnosed for each scenario with tracing enabled.}
\end{figure*}

\begin{figure*}
    \includegraphics[width=\textwidth]{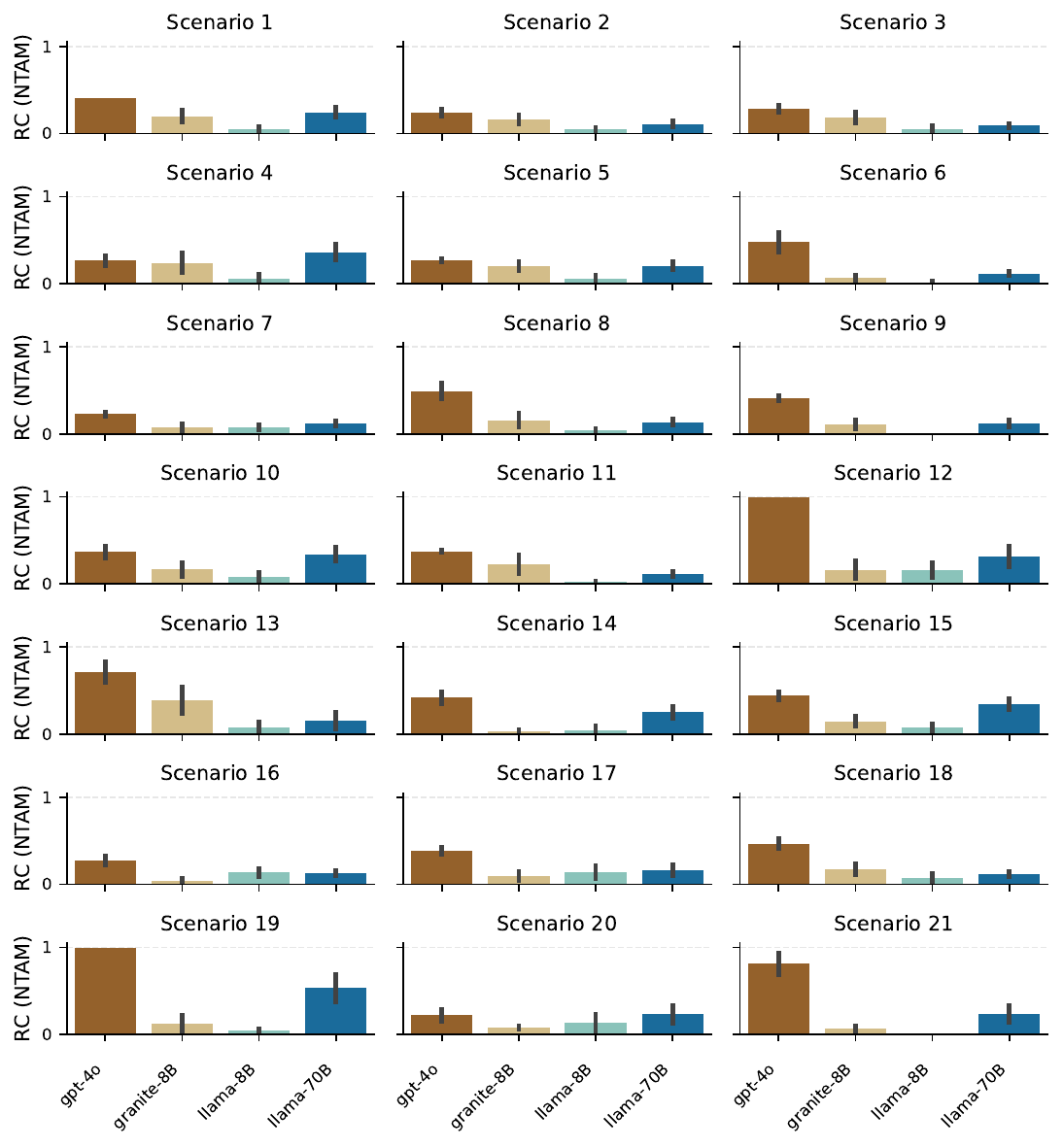}
    \caption{\label{fig:sre:trace_on_ntam_rc} Normalized topology-aware metric (NTAM) for root cause for scenarios with tracing enabled.}
\end{figure*}

\begin{figure*}
    \includegraphics[width=\textwidth]{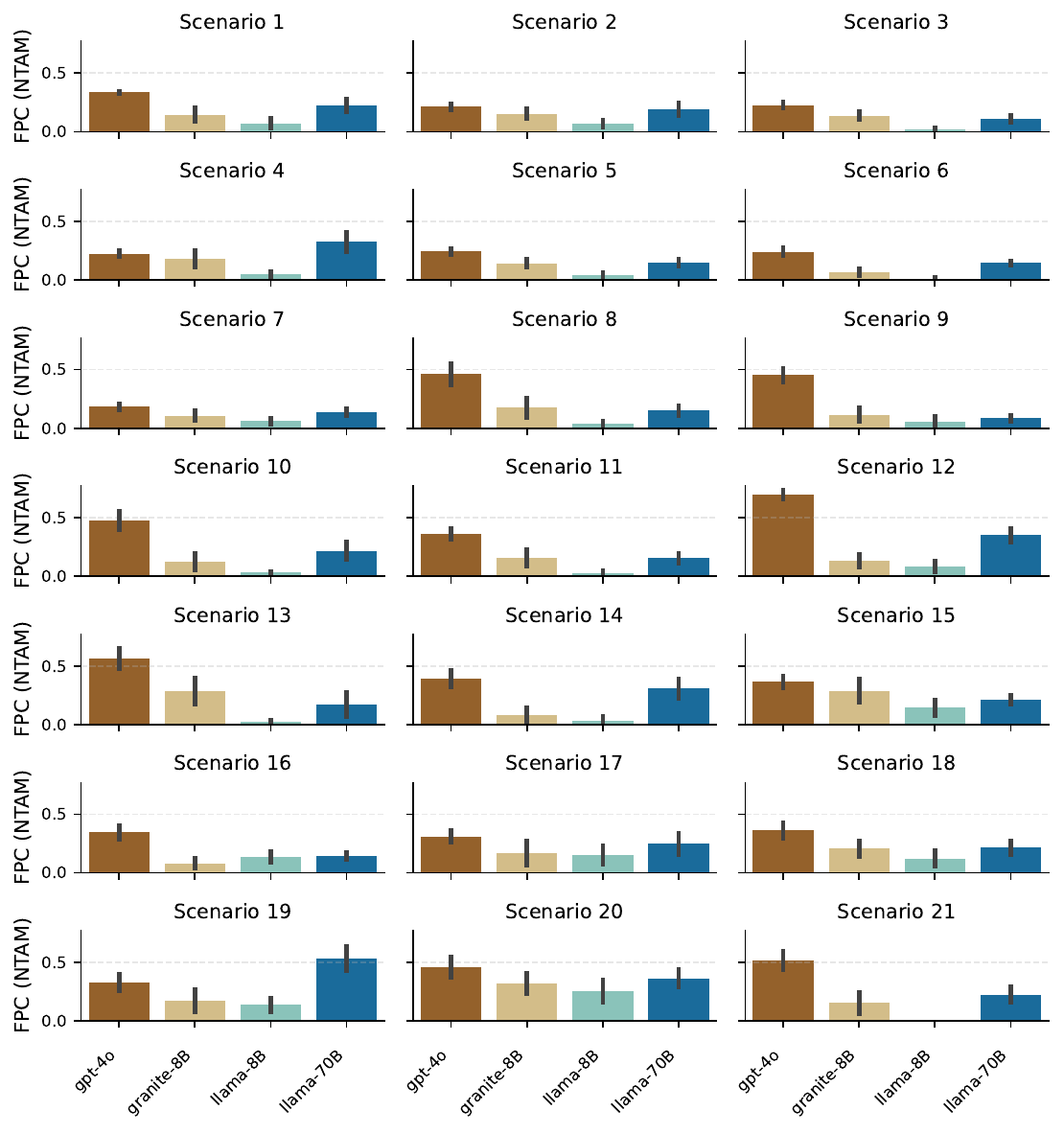}
    \caption{\label{fig:sre:trace_on_ntam_fpc} Normalized topology-aware metric (NTAM) for fault propagation chain (FPC) for each scenario with tracing enabled.}
\end{figure*}

\begin{figure*}
    \includegraphics[width=\textwidth]{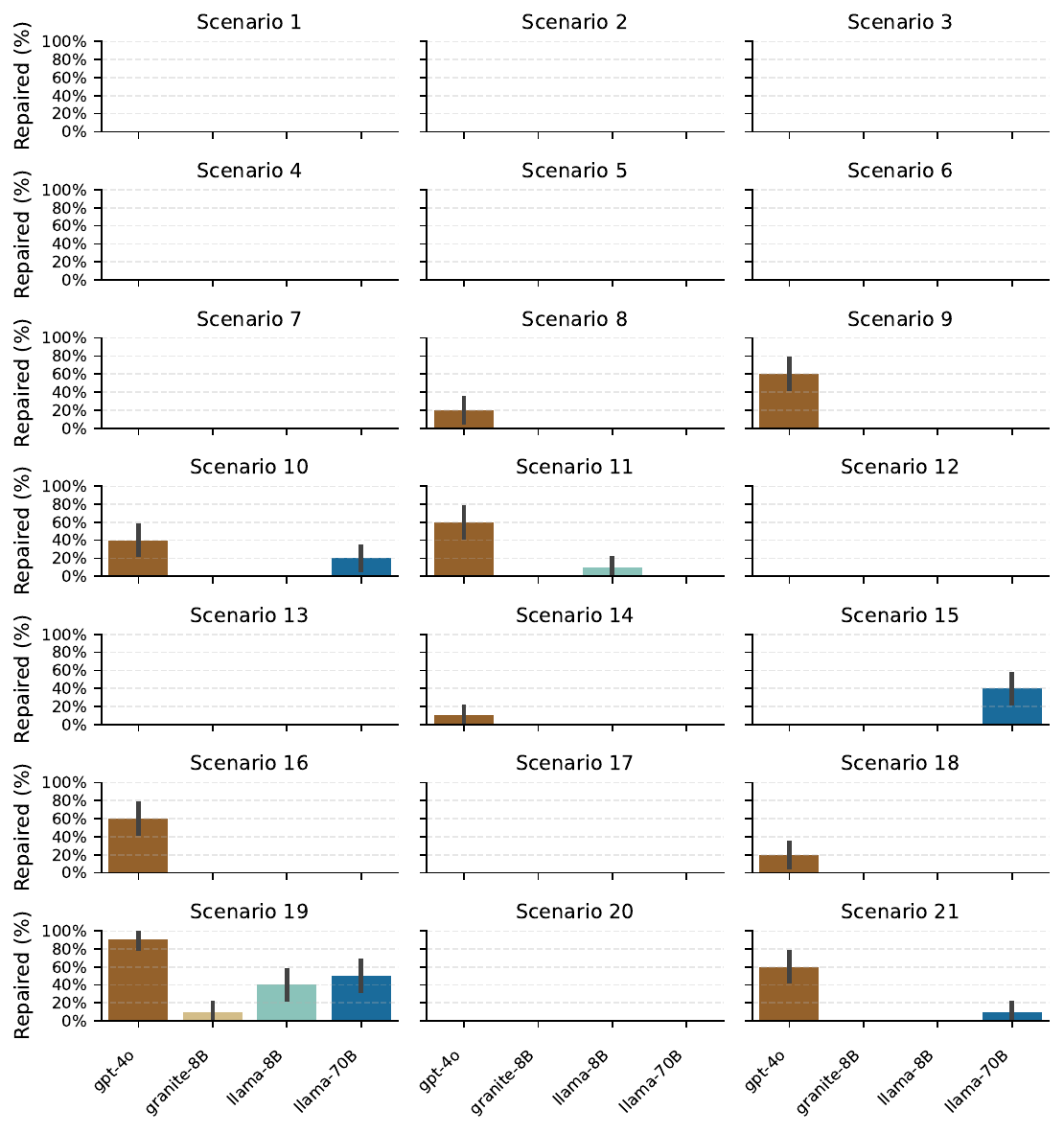}
    \caption{\label{fig:sre:trace_on_repair_pass1}Percent repaired for each scenario with tracing enabled.}
\end{figure*}

\input{appx/usecases/SRE/agent_reports}

%% file: appx/usecases/SRE/agent_reports.tex
\lstdefinelanguage{JSON}{
    basicstyle=\ttfamily\small,
    comment=[l]{//}, %
    morestring=[b]", %
    literate=
     *{:}{{\color{red}{:}}}{1} %
      {,}{{\color{red}{,}}}{1} %
      {\{}{{\color{blue}{\{}}}{1} %
      {\}}{{\color{blue}{\}}}}{1} %
      {[}{{\color{blue}{[}}}{1} %
      {]}{{\color{blue}{]}}}{1}, %
}

\subsubsection{Diagnosis and Mitigation Report}
Example of agent JSON report is also provided in Figure~\ref{fig:sre-report}. The JSON report contains information collected by the agent during its investigation. This includes a list of the faulty entities it encountered, as well as its best guess at the cause of the incident. Further, it contains a list of actions the agent took to try to mitigate the problem.

\begin{figure*}

    \centering
     \begin{mdframed}
\scriptsize
\begin{lstlisting}[language=JSON]
{
    "alert_start_time": "2025-01-25T14:54:24.976978",
    "entities": [
        {
            "id": "checkoutservice-779456f5fb-x5824",
            "root_cause": false
        },
        {
            "id": "emailservice-768cd9c799-m6wz9",
            "root_cause": true
        }
    ],
    "propagations": [
        {
            "source": "checkoutservice-779456f5fb-x5824",
            "target": "emailservice-768cd9c799-m6wz9",
            "condition": "improper configuration of emailservice to handle volume of requests from checkoutservice",
            "effect": "high error rate in checkoutservice due to emailservice not properly handling requests"
        }
    ],
    "mitigation": [
        [
            {
                "action": "Describe the deployment emailservice in the otel-demo namespace to understand its current configuration"
            },
            {
                "action": "Patch the deployment emailservice in the otel-demo namespace to increase the memory limit of the container emailservice to 200Mi"
            },
            {
                "action": "Patch the deployment emailservice in the otel-demo namespace to increase the number of replicas to 2"
            }
        ]
    ]
}
\end{lstlisting}
\end{mdframed}
\caption{Example agent output for Scenario 15.}
\label{fig:sre-report}
\end{figure*}

%% file: appx/usecases/compliance/main.tex
\section{Chief Information Security Officer (CISO) and Benchmarking the Compliance Assessment Agent}
\label{appx:ciso}
\subsection{Background}

\begin{figure*}[t!]
    \centering
    \includegraphics[width=\linewidth]{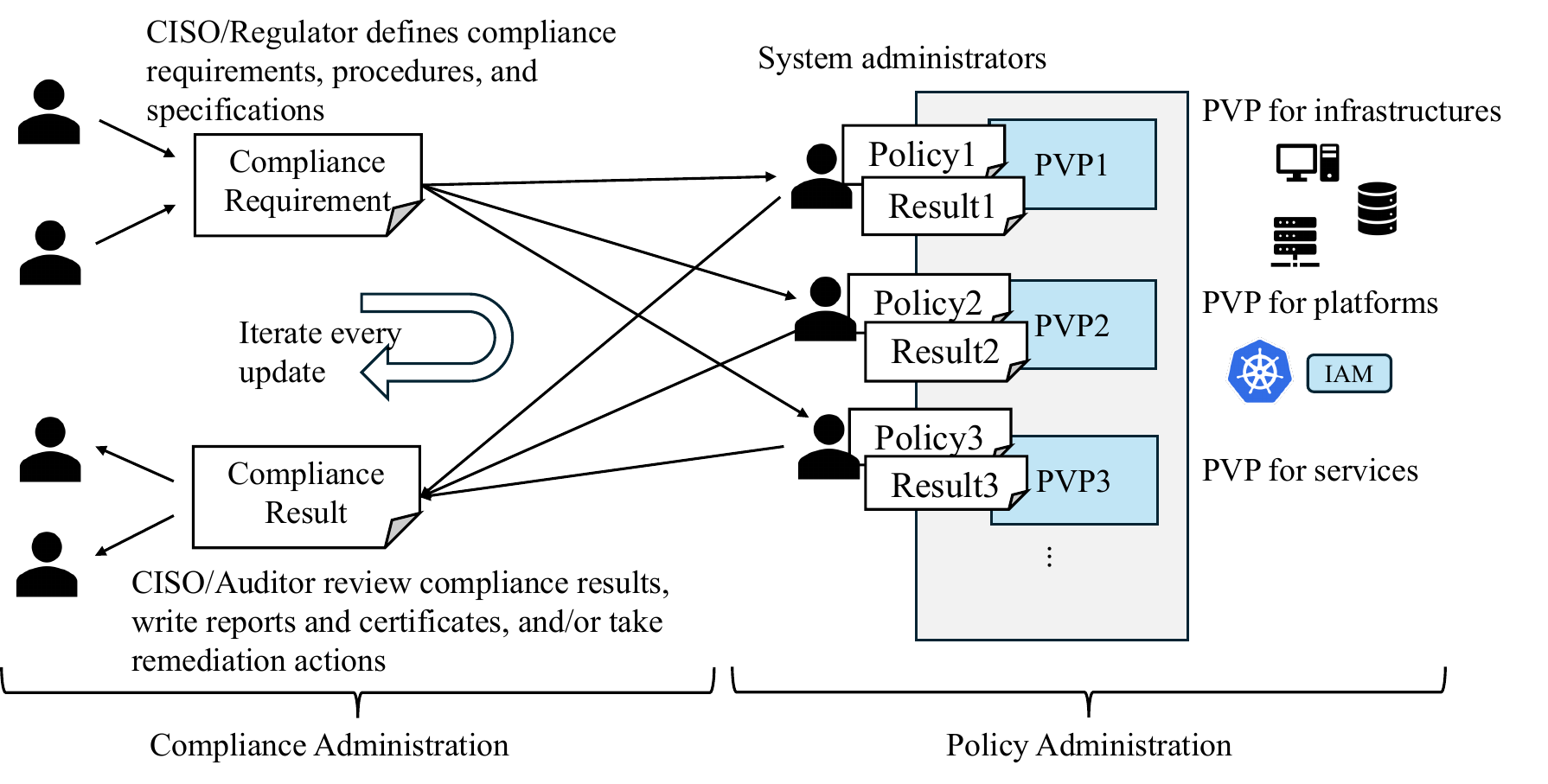}
    \caption{Compliance Authoring and Administration vs. Policy Validation Point Engines}
    \label{fig:CAA-background}
\end{figure*}
Advances in technology are increasing application and infrastructure complexity. As a result, traditional approaches that depend on a dedicated security and compliance team to identify vulnerabilities in production systems and mitigate them based on threat that they pose to the organization, are no longer working. Modern organizations rely on a Development, Security, and Operations (DevSecOps) practice, a process in which application security is verified before deployment, where security and regulatory controls are put in place at software development time. Then, post deployment, runtime checks take over. With these multiple layers of security and compliance checks owned by different teams, some of them with limited cybersecurity knowledge, not only it is no longer feasible to have manual compliance processes for the technical security controls, but the automation of those processes also needs an unprecedented acceleration to keep up the go-to-market pace and scale.

The overall process starts with CISOs, security administrators, or regulators establishing and authoring the relevant body of compliance recommendations, typically in natural language, for specific mission critical environments. Then they rely on dev teams or security focals to collect status as evidence across those environments, and to validate it against the recommendations in view of obtaining the authorization to operate or other certifications. The sought after benefit from automating the \textbf{evidence collection} and its \textbf{validation} is to enable scalability, both in handling complex environments and supporting frequent scans -daily or on demand per system fix or update-  for posture measurement and reporting. Examples of validation automation tools are policy engines such as Kyverno \cite{Intro:Kyverno}, OPA Gatekeeper \cite{Intro:GateKeeper} for Kubernetes , Ansible \cite{Intro:Ansible} for PaaS, or Cloud Security Posture Management (CSPM) solutions for the cloud. 
\Cref{fig:CAA-background} illustrates the dichotomy of compliance authoring on the left versus compliance validation via policy scripts and their diverse programmatic languages on the right. 

Automating the translation of natural-language recommendations into policy scripts requires an unprecedented level of trust and synchronization across domains and experts typically in different business units. Additionally, it also demands an unprecedented level of technical knowledge for the compliance teams typically focused on legal and IP matters. 

The rising popularity of AI agents and their projected ability to handle intricate tasks have increased the demand for AI agents managing IT systems (John, 2024; Miguel Carreon, 2024). Given the complexity of the compliance tasks, a major hurdle for this research is establishing systematic methods to assess the effectiveness of our AI agents prior to their production deployment. Consequently, there is an urgency to develop
methods for evaluation of AI agents based on real IT tasks and their corresponding environments.

We detailed our CISO compliance assessment agent deployment and execution in its git repo documentation available in open-source \cite{CISO-CAAagentrepo}.
We present below our CISO agent benchmarking performance using a well defined benchmarking methodology with real-world scenarios and environments. A sample of those scenarios with their environment executable automation packages is also available in open-source \cite{CISO-CAAscenariorepo}.

\subsection{Real-World Benchmarking }

Our CISO compliance assessment agent (CAA) and corresponding bench bring together the latest technology on compliance as code to enable the programmatic expression of regulatory controls and their posture assessment, using Gen AI generation of code to fulfill these tasks. Our agent aim is to empower a compliance team in accelerating the adoption and operation of new regulatory programs by automating the generation of code for the evidence collection and for its posture validation against the requirements, based on compliance requirements described in natural language. Our benchmarking experiments cover the end-to-end agentic workflow from the discovery of the policy assessment engine in the benchmark scenario, the generation of assessment policy as code and its real-time git PR management, deployment,  execution, and posture generation. Finally, the results are evaluated, rated, and reported in our \bench solution leaderboard.

\subsubsection{Terms and Notations}
 We define the following key terms used hereafter to describe the main aspects of the agent framework and benchmarking methodology:

\begin{enumerate}
    \item \emph{Agent:} An agent is an AI driven software that autonomously acts on behalf of a persona to solve a given task. We group the agents by Agent Types that reflect the IT operations personas, for example CISO, FInOps, or SRE type.   
    
    \item \emph{Task:} A task is a specific job corresponding to the role of a persona that the agents aim to automate. Typical tasks for CISO are to collect evidence and assess compliance controls posture.

    \item \emph{Scenario:} A scenario is a real-life occurrence of a task in a given setting. For CISO, for instance, \textit{each} Kubernetes CIS-benchmark requirement instantiated on OPA is a unique scenario. The scenarios can be grouped in classes. 
 
    \item \emph{Scenario Class:} A scenario class is a class of real-life scenarios that are grouped together expecting the same behavior and outcome from the corresponding persona. Examples of scenarios classes are the \textit{set} Kubernetes CIS-benchmarks on OPA engine, the \textit{set} of RHEL9 CIS-benchmarks on Ansible engine, or a \textit{set} of Kubernetes CIS-benchmarks updates on Kyverno engine.

    \item \emph{Scenario Environment:} A scenario environment is the part of the scenario that specifies the deployment settings.

    \item \emph{Environment State:} An environment has a countable set of states that we consider to mark a particular condition at a specific time. Example of states are an environment initial deployment state, an environment failure state after a fault or non-compliant configuration injection, or an environment remediated or compliant state after mitigation. 

    \item \emph{Goal:} A goal is the desired state for the environment known as the goal state.
\end{enumerate}
    
Agents are tasked to transition environments from their initial state to their goal state in the most efficient manner. At their disposal are environment actions, including requests for observations or actuation attempts to affect the state of the system. The agents first step towards moving the environment into the goal state is by reasoning over the outcomes of several observational actions to determine the next optimal step. With a strong hypothesis for what that is, agents seek to find and execute strategies to move towards the goal state in the most efficient way. 

Our CISO compliance assessment tasks are coupled with pre-defined scenarios for assessing the effectiveness of the agents delivering the automation in a standard manner. We detail in the next section these pre-defined CISO tasks before detailing our \bench and CISO scenarios. 

\subsubsection{CISO Compliance Assessment Tasks}

The compliance assessment tasks include various activities aimed at comparing the actual state of the systems with the desired state described in English in the policy. Based on this comparison, the system provides a "pass" or "fail" posture with respect to the policy.

\begin{itemize}
    \item \textbf{Identify Evidence Collector (IEC):} Acquiring evidence requires selecting collection mechanisms appropriate to the target system's characteristics. For instance, evidence about the state of an application in a Kubernetes cluster necessitates access to the Kubernetes API, often through tools like "kubectl" command. For host configuration evidence, tools like Ansible Playbooks are suitable. This IEC task and associated agent or tools identifies the collector used in the environment in view of generating the script for its corresponding language and interface.    
    
    \item \textbf{Identify Policy Assessment Tool (IPA):} Evaluating evidence against policies requires selecting a suitable policy engine. For general scenarios, the industry is using the open source policy engine Open Policy Agent (OPA)~\cite{Intro:OPA} with its specific programmatic language Rego. Alternatively, for Kubernetes-specific configurations, Kyverno policies prevalently used along OPA. This IPA task and associated agent or tools identifies the appropriate policy engine for the scenario's policy at hand, in view of generating policies code according to the policy engine's  programmatic language and interface.    
    
    \item \textbf{Collect Evidence (CE):} This CE task and associated agent or tools is responsible for the actual evidence collection, including the generation of code, management of code, deployment and execution of the evidence collection code to acquire the actual evidence from the environment. Proper placement and execution of the code are necessary to achieve this in a reliable and scalable manner.
    
    \item \textbf{Scan Assessment Posture (SAP):} This SAP task and associated agent or tools is responsible for generating the posture whether the evidence does or does not satisfy the scenario CIS-benchmark requirement. It includes the generation of validation code, management of code, deployment and execution of code on the policy engine to assess the evidence and produce the compliance posture.  
\end{itemize}

Table \ref{tab:bench_scenarios} summarizes the CISO tasks initially supported in our \bench, namely CE and SAP. The other will be covered in subsequent releases. 
These tasks are executed in \bench against predefined, standard scenarios and compared to the ground truth expected assessment posture "pass"/"fail" stored in the \bench under each scenario environment specification. 

\subsection{\bench Architecture for handling CISO Tasks}

\begin{figure*}[t!]
    \centering
    \includegraphics[width=0.8\linewidth]{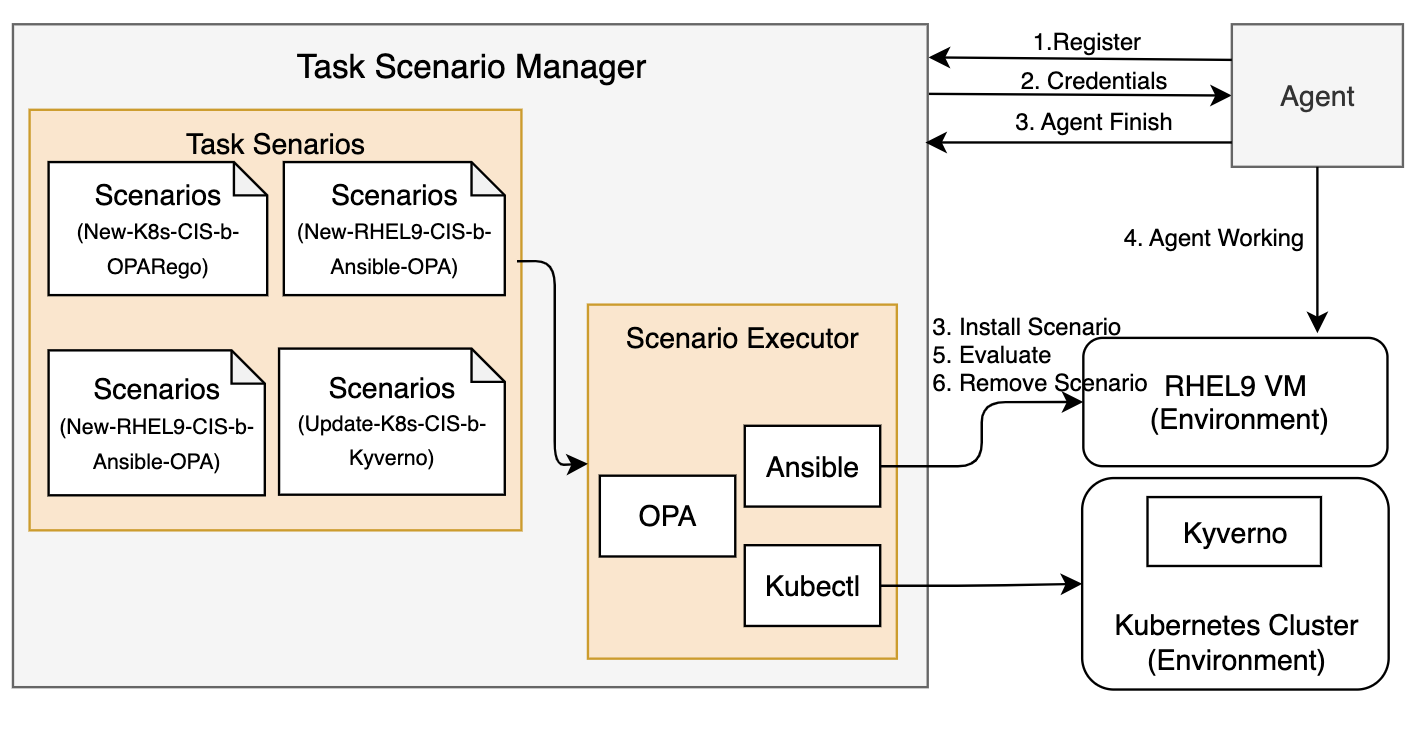}
    \caption{Architecture of \bench responsible for orchestrating CISO scenarios.}
    \label{fig:bench_design_CISO}
\end{figure*}

\bench uses open source technologies to create repeatable and reproducible scenarios and environments for the CISO tasks, on a Kubernetes cluster as shown in \Cref{fig:bench_design_CISO}.

\subsubsection{Principles}

\begin{figure}[t!]
        \centering
        \includegraphics[width=0.45\linewidth]{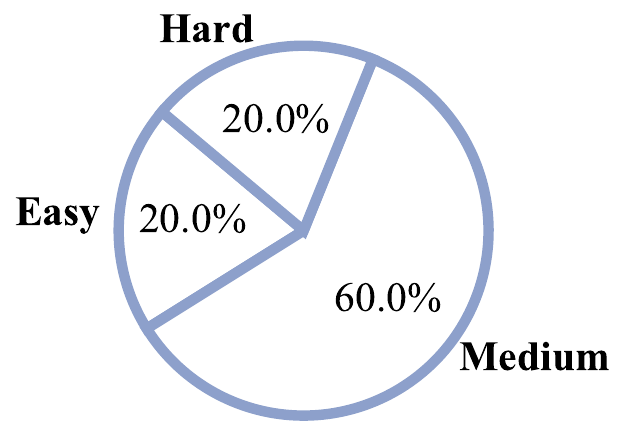}
        \caption{CISO scenario complexity.}
        \label{fig:ciso-task-difficulty2}
    \end{figure}

Following the bench principles indicated in the introduction, our \bench uses open-source technologies to construct completely repeatable and reproducible  scenarios that simulate real-world incidents. 
\begin{itemize}[left=0pt, topsep=0pt, partopsep=0pt, itemsep=0pt, parsep=0pt]
    \item \textbf{Mimic CISO best practices.} \bench follows the guidelines outlined by the National Institute of Standards and Technology (NIST) Cybersecurity Framework (CSF)~\cite{NIST-CSF} for CISOs and security teams to improve their organization's cybersecurity as follows: (1) \textit{Identify} critical data, systems, assets, and capabilities to protect; (2) \textit{Protect} via security measures that limit the impact of incidents; (3) Develop a strategy to \textit{detect} non-compliance with clear procedures and tools; (4) \textit{Respond} via plans to quickly eliminate threats and mitigate damage; (5) Design a \textit{recovery} policy to support timely recovery to normal operations. Our first \bench release covers the CISO's CSF first three activities by evaluating the following compliance tasks: (1) Identify and collect evidence from the systems in the selected scenarios; (2) Implement the policies recommended in the scenarios; (3) Assess the policies posture to detect failure into non-compliance. The remaining two CISO's CSF activities of \textit{respond} and \textit{recover} will make the topic of a future \bench release that will support agents collaboration, namely the leverage of the SRE agent for mitigation. 
   
    \item \textbf{Mimic CISO real-world problems.} 
    We studied and used in \bench the real-world Cloud Internet Security (CIS) Benchmarks~\cite{cis-b} which are a set of best practices for securing the IT cloud infrastructure. They are recognized worldwide as the cloud security standards.
    We used those CIS-benchmarks to create our CISO compliance policies scenarios of various complexity levels: 25\% Easy, 50\% Medium, and 25\% Hard policies (see Figure \ref{fig:ciso-task-difficulty2}). 

    \item \textbf{Provide observability.} 
    Cloud Native Compute Foundation (CNCF) recent Sandbox project (OSCAL-compass, 2024) released a compliance as code SDK to support the machine readable compliance as code standard (OSCAL, 2024) of the
National Institute of Standards and Technology for programmatic usage in compliance automation. \bench CISO automation leverages this methodology to represent the CIS-benchmarks requirements, detect the events of creation or update of requirements, and trigger the creation or update of evidence collection and validation code.

    \item \textbf{{Ensure} Determinism.} \bench enforces the scenarios and their environments are generated as per the specification, while the environment cleanup after each scenario ensures a clean slate for the next run. 
\end{itemize}

\subsubsection{\bench Architecture}

The environments for the benchmarking scenarios comprise a Kubernetes Cluster and virtual machines (VMs). Each CISO scenario pre-defined in the \bench as described in the section hearafter, is managed by a deployable stack, a software component responsible for handling the benchmarking process and environment in a real run-time environment. 

The deployable stack manages various tasks, including preparation ("deploy\_environment"), fault injection ("inject\_fault"), agent performance evaluation ("evaluate"), and environment cleanup ("delete\_environment") for each benchmark scenario run. Each deployable stack is specifically designed for a particular CISO compliance assessment scenario, ensuring the necessary tools, configurations, and policies are in place. Environment administrators define these scenarios deployable stacks and configure the required software, which can involve setting up policy engines, tools, or creating conditions that simulate violations of specific compliance requirements.

A deployable stack may for instance intentionally exhibit a misconfiguration settings to mimic a violation of a particular compliance standard.  In the execution of a scenario, the agent is presented initially with the natural language description of that scenario compliance requirement. Based on the description, the agent generates the necessary artifacts, including scripts for evidence collection and policies for evidence evaluation. The run-time scenario environment is deployed and made accessible to the agent. The agent accesses the environment to retrieve evidence, deploy policies, and work toward achieving the specified automation goal of assessing the compliance posture, in this case as "fail" or "not-satisfied".

\begin{figure*}[t!]
    \centering
    \includegraphics[width=0.85\linewidth]{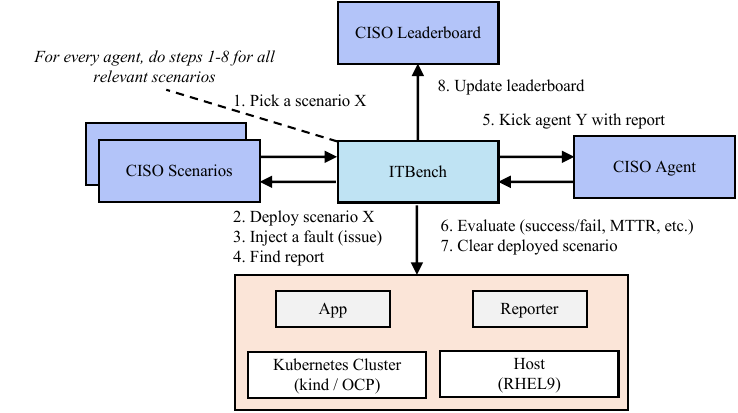}
    \caption{CISO Compliance Assessment Agent end-to-end Benchmarking Process.}
    \label{fig:CISO_CAA_ench}
\end{figure*}

During the scenario run process, the agent notifies the \bench of the start and completion of its tasks. Upon receiving the completion notification, the \bench accesses the environment to measure the benchmarking metrics. Once the metrics for all predefined scenarios are collected, they are aggregated and displayed on the \bench Leaderboard. Fig.~\ref{fig:CISO_CAA_ench} illustrates the end-to-end benchmarking process for the CISO scenarios.

In the context of our CISO compliance assessment benchmarking, the scenario environment, including Kubernetes Clusters and VMs, is prepared by the Agent Submitter. The setup and metric measurements are automated using a tool called Mini-Bench. Benchmarking results, including Task Metrics, are registered with a central Bench Server via API. These results are displayed on the Leaderboard on the Bench Server. This setup allows benchmarking to proceed uniformly, whether using the Agent Submitter's local environment or a remote environment, as the same API interactions are employed in both cases.

\subsection{\bench Real-World CISO Scenarios}
We used in \bench the real-world Cloud Internet Security (CIS) Benchmarks~\cite{cis-b} standard to create our CISO compliance assessment scenarios. The technologies that we have considered as playground for benchmarking our CISO agent are Kubernetes and RHEL9, however, any other technology available in CIS can be leveraged with their CIS-benchmark infusing the policies scenario in \bench.

\Cref{tab:Kube-CIS-b-bench} and \Cref{tab:CRHEL9-CIS-b} illustrate examples of the typical CIS-benchmarks recommendations. Each scenario rendered on \bench is designed to mirror the complexity of the recommendation. This ensures that \bench replicates real-world compliance requirements thus allowing for a realistic evaluation of agents.

\begin{table*}[htbp]
\centering
\small
\begin{threeparttable}
\caption{Kubernetes - Center for Internet Security Benchmarks (sample)}
\label{tab:Kube-CIS-b-bench}
\centering
\begin{tabular}{@{}llp{3cm}p{1.3cm}p{7.3cm}@{}}
\toprule
\textbf{Section \#} &
\textbf{Recommendation \#} &
\textbf{Title} &
\textbf{Assessment Status} &
\textbf{Description} \\
\midrule
1.1.1 & 1.1.1.1 & Ensure cramfs kernel module is not available & Automated &
The `cramfs` filesystem type is a compressed read-only Linux filesystem embedded in small footprint systems. A `cramfs` image can be used without having to first decompress the image.\\

1.1.1 & 1.1.1.2 & Ensure freevxfs kernel module is not available & Automated &
The `freevxfs` filesystem type is a free version of the Veritas type filesystem. This is the primary filesystem type for HP-UX operating systems.\\

1.1.1 & 1.1.1.3 & Ensure hfs kernel module is not available & Automated &
The `hfs` filesystem type is a hierarchical filesystem that allows you to mount Mac OS filesystems.\\

1.1.1 & 1.1.1.4 & Ensure hfsplus kernel module is not available & Automated &
The `hfsplus` filesystem type is a hierarchical filesystem designed to replace `hfs` that allows you to mount Mac OS filesystems.\\

1.1.1 & 1.1.1.5 & Ensure jffs2 kernel module is not available & Automated &
The `jffs2` (journaling flash filesystem 2) filesystem type is a log-structured filesystem used in flash memory devices.\\

1.1.1 & 1.1.1.8 & Ensure usb-storage kernel module is not available & Automated &
USB storage provides a means to transfer and store files ensuring persistence and availability of the files independent of network connection status. Its popularity and utility has led to USB-based malware being a simple and common means for network infiltration and a first step to establishing a persistent threat within a networked environment.\\

1.1.1 & 1.1.1.9 & Ensure unused filesystems kernel modules are not available & Manual &
Filesystem kernel modules are pieces of code that can be dynamically loaded into the Linux kernel to extend its filesystem capability, or so-called base kernel, of an operating system. Filesystem kernel modules are typically used to add support for new hardware (as device drivers), or for adding system calls.\\
\bottomrule
\end{tabular}
\end{threeparttable}
\end{table*}

\begin{table*}[htbp]
\small
\begin{threeparttable}
\caption{Red Hat Enterprise Linux - Center for Internet Security Benchmarks (sample).}
\label{tab:CRHEL9-CIS-b}
\centering
\begin{tabular}{@{}llp{3cm}p{1.3cm}p{7.45cm}@{}}
\toprule
\textbf{Section \#} & 
\textbf{Recommendation \#} & 
\textbf{Title} & 
\textbf{Assessment Status} & 
\textbf{Description} \\
\midrule
1.1.2 & 1.1.2.3 & Ensure noexec option set on \texttt{/tmp} partition & Automated &
The `noexec` mount option specifies that the filesystem cannot contain executable binaries. \\

1.1.2 & 1.1.2.4 & Ensure nosuid option set on \texttt{/tmp} partition & Automated &
The `nosuid` mount option specifies that the filesystem cannot contain `setuid` files. \\

1.1.3 & -- & Configure \texttt{/var} & -- &
The `/var` directory is used by daemons and other system services to temporarily store dynamic data. Some directories created by these processes may be world-writable. \\

1.1.3 & 1.1.3.2 & Ensure nodev option set on \texttt{/var} partition & Automated &
The `nodev` mount option specifies that the filesystem cannot contain special devices. \\

1.1.3 & 1.1.3.3 & Ensure nosuid option set on \texttt{/var} partition & Automated &
The `nosuid` mount option specifies that the filesystem cannot contain `setuid` files. \\

1.1.4 & -- & Configure \texttt{/var/tmp} & -- &
The `/var/tmp` directory is a world-writable directory used for temporary storage by all users and some applications. Temporary files residing in `/var/tmp` are to be preserved between reboots. \\

1.1.4 & 1.1.4.2 & Ensure noexec option set on \texttt{/var/tmp} partition & Automated &
The `noexec` mount option specifies that the filesystem cannot contain executable binaries. \\

1.1.4 & 1.1.4.3 & Ensure nosuid option set on \texttt{/var/tmp} partition & Automated &
The `nosuid` mount option specifies that the filesystem cannot contain `setuid` files. \\

1.1.4 & 1.1.4.4 & Ensure nodev option set on \texttt{/var/tmp} partition & Automated &
The `nodev` mount option specifies that the filesystem cannot contain special devices. \\

1.1.5 & -- & Configure \texttt{/var/log} & -- &
The `/var/log` directory is used by system services to store log data. \\

1.1.5 & 1.1.5.2 & Ensure nodev option set on \texttt{/var/log} partition & Automated &
The `nodev` mount option specifies that the filesystem cannot contain special devices. \\

1.1.5 & 1.1.5.3 & Ensure noexec option set on \texttt{/var/log} partition & Automated &
The `noexec` mount option specifies that the filesystem cannot contain executable binaries. \\

1.1.5 & 1.1.5.4 & Ensure nosuid option set on \texttt{/var/log} partition & Automated &
The `nosuid` mount option specifies that the filesystem cannot contain `setuid` files. \\

1.1.6 & -- & Configure \texttt{/var/log/audit} & -- &
The auditing daemon, `auditd`, stores log data in the `/var/log/audit` directory. \\

1.1.6 & 1.1.6.2 & Ensure noexec option set on \texttt{/var/log/audit} partition & Automated &
The `noexec` mount option specifies that the filesystem cannot contain executable binaries. \\

1.1.6 & 1.1.6.3 & Ensure nodev option set on \texttt{/var/log/audit} partition & Automated &
The `nodev` mount option specifies that the filesystem cannot contain special devices. \\

1.1.6 & 1.1.6.4 & Ensure nosuid option set on \texttt{/var/log/audit} partition & Automated &
The `nosuid` mount option specifies that the filesystem cannot contain `setuid` files. \\

1.1.7 & -- & Configure \texttt{/home} & -- &
Please note that home directories could be mounted anywhere and are not necessarily restricted to `/home`, nor restricted to a single location, nor is the name restricted in any way. Checks can be made by looking in `/etc/passwd`, looking over the mounted file systems with `mount` or querying the relevant database with `getent`. \\
\bottomrule
\end{tabular}
\end{threeparttable}
\end{table*}

\subsection{CISO Scenario Classes and their Complexity}
\subsubsection{New-K8s-CIS-b-Kyverno}

New-K8s-CIS-b-Kyverno represents the Easy scenario class in the \bench. The 10 scenarios in this class are prepared based on the CIS Benchmark for Kubernetes, specifically focusing on the Pod Security Policy. This scenario assumes a Kubernetes cluster with a pre-configured Kyverno policy engine. Within the cluster, certain misconfigurations related to Pod Security Policy are present, but the Agent is unaware of their exact locations.

The requirements for the misconfigurations that need to be addressed are communicated to the Agent. Based on these requirements, the Agent generates a Kyverno policy and deploys it to the cluster. Subsequently, the Agent collects the report from the cluster. The accuracy of this report is verified by checking whether it successfully identifies the misconfigurations. If the policy is correctly generated and deployed, the report should indicate the appropriate posture. Conversely, if errors occur, an incorrect posture will be reported.

In this scenario, the four Compliance Assessment Task are evaluated as follows:

\begin{itemize}
\item IEC: Assessed by verifying whether the correct configuration is reflected in the Kyverno policy.
\item IPA: Evaluated by confirming that the policy is successfully generated for Kyverno.
\item CE: Measured by verifying whether a Kyverno report is generated in the predefined location.
\item SAP: Determined by whether the posture reported in the Kyverno report matches the expected value.
\end{itemize}

\subsubsection{New-K8s-CIS-b-OPARego}

New-K8s-CIS-b-Kubectl-OPARego is categorized under Medium complexity scenarios. This benchmark comprises 10 scenarios derived from the CIS Benchmark for Kubernetes, specifically focusing on Pod Security Policies.

The foundational background for this scenario closely resembles that of New-K8s-CIS-b-Kyverno, sharing the assumption of a Kubernetes cluster as the operating environment and CIS Benchmark for Kubernetes. The key difference lies in the dual output for Open Policy Agent (OPA) policy engine: the generation of both an evidence fetcher script and a policy checker code. 

\begin{itemize}
\item A fetcher script is designed to gather the required evidence from the target cluster by executing kubectl commands.
\item A policy checker  verifies the collected evidence for compliance based on predefined rules. This is implemented using the Open Policy Agent (OPA) and defined through OPA Rego policies.
\end{itemize}

The goal of the agent in this scenario is to generate two outputs: 1) a script executing kubectl commands (fetcher), 2) an OPA Rego policy for compliance verification (checker).

The verification process evaluates the outputs generated by the agent as follows: The fetcher script, which consists of kubectl commands, is executed against a real Kubernetes cluster to collect evidence. The collected evidence is then assessed using the generated OPA Rego policy and the OPA policy engine to verify whether the results align with expected compliance outcomes.

The scenario is assessed on the following four Compliance Assessment Tasks:
\begin{itemize}
\item IEC: Determine whether a fetcher script, incorporating kubectl commands, is successfully generated.
\item IPA: Verify if the checker, implemented as an OPA Rego policy, is correctly generated.
\item CE: Check whether evidence can be successfully collected by executing the fetcher script against the cluster.
\item SAP: Verify that the OPA Rego policy evaluates the collected evidence as expected, producing the correct compliance assessment.
\end{itemize}

This approach provides a structured evaluation of the agent's capability to generate effective scripts and policies for Kubernetes cluster compliance assessment.

\subsubsection{New-RHEL9-CIS-b-Ansible-OPA}

New-RHEL9-CIS-b-Ansible-OPA belongs to the Medium complexity scenario class and comprises 20 scenarios based on the CIS benchmark for RHEL9 OS. This scenario shares common characteristics with New-K8s-CIS-b-Kubectl-OPARego, including the generation of two codes (a fetcher script and a checker policy), and the use of OPA Rego Policies for compliance verification. The primary distinction lies in the target system: unlike New-K8s-CIS-b-Kubectl-OPARego, which focuses on Kubernetes clusters, this scenario targets RHEL9 hosts. Consequently, instead of using kubectl as the fetcher script, New-RHEL9-CIS-b-Ansible-OPA employs Ansible playbooks. The objective of this scenario is to generate Ansible playbooks as fetcher scripts and OPA Rego Policies as checkers.

The verification process evaluates the outputs generated by the agent as follows: the fetcher script (Ansible playbook) is executed against a real RHEL9 host to collect evidence. The collected evidence is subsequently analyzed using the generated OPA Rego policy on the OPA policy engine, assessing whether the results align with the expected compliance outcomes.

In this scenario, the agent's performance is evaluated on the following four Compliance Assessment Task:

\begin{itemize}
\item IEC: Does the agent generate a fetcher script in the form of an Ansible playbook?
\item IPA: Does the agent generate an OPA Rego policy for the checker?
\item CE: Can the Ansible playbook successfully execute against an RHEL9 host and collect relevant evidence?
\item SAP: Can the generated OPA Rego policy evaluate the collected evidence and produce the expected compliance results?
\end{itemize}

This scenario is designed to assess the agent's performance in compliance evaluation tasks for host management environments other than Kubernetes, specifically focusing on RHEL9 systems.

\subsubsection{Update-K8s-CIS-b-Kyverno}

Update-K8s-CIS-b-Kyverno falls under the scenario class with a complexity level classified as Hard and currently includes 10 scenarios. Unlike New-K8s-CIS-b-Kyverno, which generates new Kyverno policies based on new requirements specified for that environment in the goal, this scenario involves a different objective. Specifically, it takes an existing Kyverno policy as input, along with instructions detailing modifications to the original requirements, and generates an \textit{updated} policy to meets the revised requirements. The updated policy is then deployed (or updated) as the final output.

The validation process for this scenario is consistent with the methodology used in New-K8s-CIS-b-Kyverno.

\subsection{CISO \bench Evaluation}
 
We conduct our experiments primarily on AWS EC2 instances (m4.xlarge), although \bench can also be readily deployed on a consumer-grade laptop using a pseudo-cluster, thus making it easier to develop AI agents.

We measure the efficacy of our CISO compliance assessment agent on a set of 50 scenarios across the four scenario classes introduced in \Cref{tab:bench_scenarios}. Each scenario class imposes a distinct set of CIS-benchmarks requirements (e.g., ``minimize the admission of containers wishing to share the host network namespace''), each class has a specific level of complexity (e.g., Easy, Medium, Hard), and generates scenario-specific code artifacts. 

In our evaluation we considered a variety of LLMs, such as GPT-4o, Llama-3.3-70B-instruct, Llama-3.1-8B-instruct, and Granite-3.1-8B-instruct for tasks that rely on natural language understanding and reasoning. For code-focused use cases, we additionally utilize GPT-4o-mini, Llama-3.1-405b-instruct, and Mixtral-8x7b-instruct. 
All models use a context window of 128K tokens, enabling them to process more extensive input sequences.

\subsubsection{Evaluation Metrics}

The efficacy of our CISO agents is measured based on the ability to detect artifact misconfigurations (aka non-compliance, e.g., no minimum count of containers sharing namespace, or the count is above the threshold), or confirm proper configurations (aka compliance), within the varied environments of the scenario classes randomly injected with misconfigurations. 

We evaluate how effectively the agent detects the (non)compliance using the following metrics:
\begin{itemize}[left=0pt, topsep=0pt, partopsep=0pt, itemsep=0pt, parsep=0pt]
    \item Success rate is quantified using pass@1.
    \item Efficiency is captured through Time to Process (TTP).
\end{itemize}

\subsubsection{Metric definitions}

\textbf{pass@1.}
We evaluate the agent proper assessment of the posture "pass"/"fail" using the pass@1 metric~\cite{chen2021evaluating}, which is defined as follows:
$$\text{pass@}k := \mathbb{E}_{\text{Scenarios}}\left[1 - \frac{\binom{n-c}{k}}{\binom{n}{k}}\right] $$
It is an unbiased estimator of correctness in \textit{k}=1 trials across all scenarios. 

\textbf{Time to Process.}
We identify the scenarios where an agent identifies the posture  successfully (i.e., misconfiguration results in a "fail" posture, while a compliant configuraiton results in a "pass" posture). For these task scenarios, we calculate \textit{TTP}, which measures how soon an agent performed the assessment. Otherwise, \textit{TTP} is set to infinite. 

\subsubsection{Evaluation Results}

\textbf{Overall agent results.}

\begin{figure*}
    \includegraphics[width=0.95\textwidth]{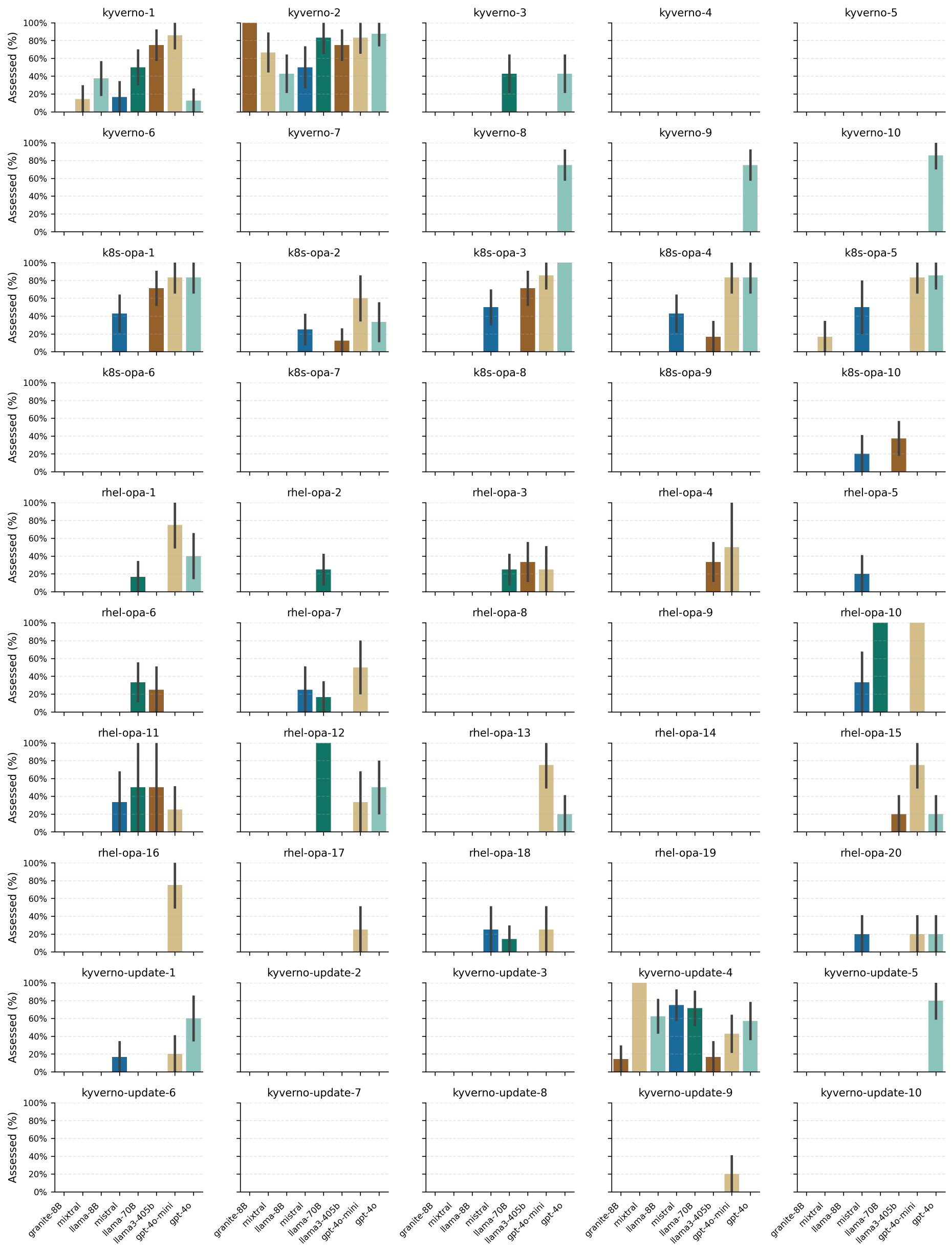}
    \caption{\label{fig:result:percent_repaired}Percent pass@1 for each scenario.}
    \label{fig:pass-percent-ciso}
\end{figure*}

Overall our results in \Cref{tab:cisoagent-eval} and \Cref{fig:pass-percent-ciso} show the GPT-based models dominate on both pass@1 and Time to Process metrics. The pass@1 is nearly 2x better than second-best models (alternating between llama-3.1-405b-instruct and mistral-large-2), while the TTP shows a handling of the scenarios in the minimal time across our scenario classes.

\textbf{Impact of Scenario Complexity}

The complexity of the CISO scenarios is directly mapped to scenario classes. For example, Kyverno scenarios are of Easy complexity, k8s-opa and rhel-opa are of Medium complexity, while Kyverno-update scenarios are of Hard complexity. 
As expected, all the models struggle, as expected, as the difficulty of the scenarios increases from the Easy \textit{kyverno} class to the Hard \textit{kyverno-upadate} class. 

\subsubsection{CISO Compliance Assessment Agent Trajectories}  

We examine two sample trajectories from our experiments to highlight the agent's thought patterns and actions. These include two scenarios, one successful and one unsuccessful, derived from the NEW-CIS-B-RHEL9-ANSIBLE-OPA experiments with LLaMa 405B, focusing on the CIS Benchmark for RHEL9 (control section 5.1.8 and 5.2.19).

In the successful scenario, the playbook is successfully generated, tested independently, leading to successful data collection, followed by the successful creation of an OPA, and ultimately detecting issues through evaluation in the bench.

In the unsuccessful scenario, the playbook is successfully generated and tested independently, leading to successful data collection, but the OPA cannot be properly created. This results in repeated syntax error corrections and retries, eventually reaching the maximum retry limit.

Following this, we present the trajectories for both cases. In the successful scenario the agent is tasked with performing a compliance assessment based on the following request:

\begin{mdframed}
Ensure cron is restricted to authorized users.
The system must:
  Allow only authorized users to schedule jobs using cron.
Specific requirements:

- The file /etc/cron.allow must exist and specify the list of authorized users.

- The file /etc/cron.deny must not exist.

- Unauthorized users (e.g., user1, user2) must not be listed in /etc/cron.allow.

To check the condition, do the following steps. 
- collect configuration from the RHEL host and save it locally. you can use ansible-playbook to do that. 
- chcek if the condition is met by using rego policy with the input given by the step above. 

for those steps, you need to create ansible playbook `playbook.yml` and OPA rego policy `policy.rego`.
If you can fix the generated code, do it and run the fixed code again.
Once you get a final answer, you can quit the work.
\end{mdframed}

The agent runs this task, provided with the target environment credentials and a working directory for saving the generated files to be submitted.
The Figure \ref{fig:rhelsuccess-1} shows the trajectory that the agent takes to achieve this request.  

\def\rhelsuccess{Success Scenario}
\def\rhelfailure{Failure Scenario}

\begin{figure}[hbtp]
    \centering
    \includegraphics[width=0.8\linewidth]{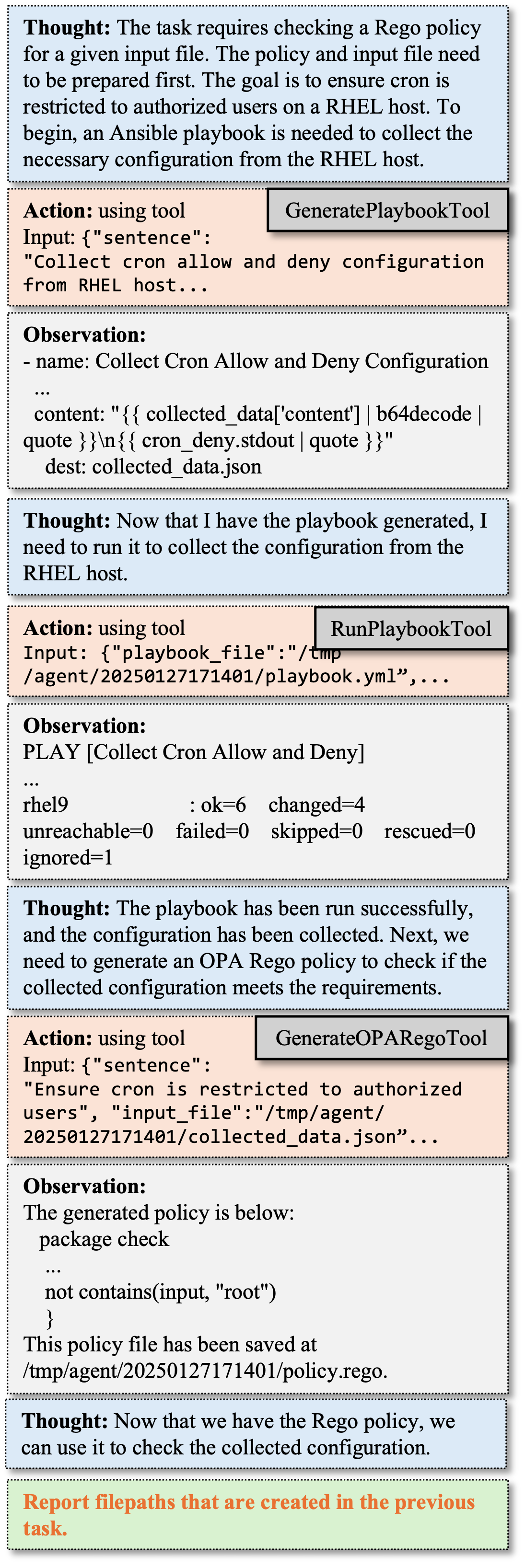}
    \caption{Sample Trajectory of LLaMa 405B in \rhelsuccess} 
    \label{fig:rhelsuccess-1}
\end{figure} 

\vspace{1em}

In the next scenario (CIS Benchmark control section 5.2.19), the agent is given a different compliance requirement.  
This time, the focus is on ensuring that SSH login sessions adhere to security best practices.
However, unlike the previous scenario, the agent encounters a critical issue: it fails to correctly write a syntactically valid Rego policy, causing repeated OPA errors during execution.  

The compliance requirement for Scenario 5.2.19 is as follows:

\begin{mdframed}
Ensure SSH LoginGraceTime is set to one minute or less 

To check the condition, do the following steps. 
- collect configuration from the RHEL host and save it locally. you can use ansible-playbook to do that. 
- chcek if the condition is met by using rego policy with the input given by the step above. 

for those steps, you need to create ansible playbook `playbook.yml` and OPA rego policy `policy.rego`.
If you can fix the generated code, do it and run the fixed code again.
Once you get a final answer, you can quit the work.

\end{mdframed}

\begin{figure}[hbtp]
    \centering
    \includegraphics[width=0.8\linewidth]{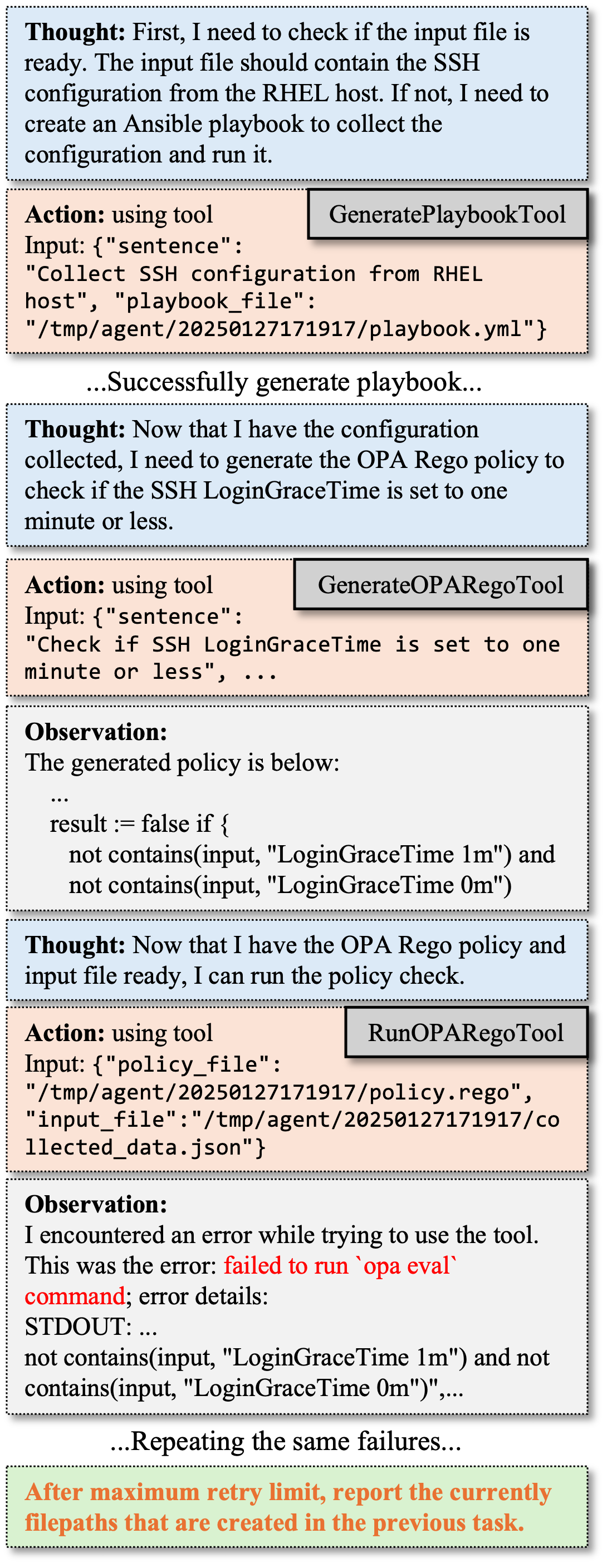}
    \caption{Sample Trajectory of LLaMa 405B in \rhelfailure} 
    \label{fig:rhelfailure-1}
\end{figure}

The agent run this task, provided with the target environment credentials and a working directory for saving the generated files to be submitted.
The Figure \ref{fig:rhelfailure-1} show the trajectory that the agent takes to achieve this request.  

\vspace{1em}

After repeating this process three times, the maximum retry limit was reached, and only the currently saved playbook.yml and policy.rego were used at the evaluation. At the evaluation, this policy.rego is syntactically wrong so the evaluation did not pass.

\begin{table*}[htp]
\small
\centering
\begin{threeparttable}
\caption{Experimental details}
\label{tab:exp-setup-ciso}
\begin{tabular}{@{}l c c c c c@{}}
\toprule
\multirow{2}{*}{\textbf{CISO/Agent: Models}} 
 & \multirow{2}{*}{\textbf{Scenarios}}
 & \multicolumn{4}{c}{\textbf{Experiment Setup}} \\
\cmidrule(lr){3-6}
 & 
 & \textbf{\#Repeats} 
 & \textbf{\#Total} 
 & \textbf{\%Exp. Completion}
 & \textbf{\%Agent Submission}\\
\midrule
\textbf{granite-3.1-8B-instruct}           & 50 & 8 & 400 & 91.41\% & 88.48\% \\
\textbf{mixtral-8x7B-instruct}           & 50 & 8 & 400 & 91.02\% & 94.67\% \\
\textbf{llama-3.1-8B-instruct}   & 50 & 8 & 400 & 93.21\% & 85.71\% \\
\textbf{llama-3.3-70B-instruct} & 50 & 8 & 400 & 92.82\% & 89.11\% \\
\textbf{mistral-large-2}           & 50 & 8 & 400 & 94.48\% & 85.23\% \\
\textbf{llama-3.1-405B-instruct}                & 50 & 8 & 400 & 92.43\% & 84.50\% \\
\textbf{gpt-4o-mini}                       & 50 & 8 & 400 & 95.25\% & 85.71\% \\
\textbf{gpt-4o}                       & 50 & 8 & 400 & 90.83\% & 90.54\% \\
\bottomrule
\end{tabular}
\begin{tablenotes}
\footnotesize
\item \textit{Note:} “\%Exp. Completion” is the percentage of experiments that were started and finished by the bench runner correctly. \\
\item \textit{Note:} “Trials Agent Submitted” is the percentage of all trials completed in which the agent returned results.\\
\end{tablenotes}
\end{threeparttable}
\end{table*}

\begin{table*}[h]
\small
\centering
\begin{threeparttable}
  \caption{CISO/Agent: Assessed Results}
  \label{tab:agent101}
  \begin{tabular}{@{}lcccccc@{}}
    \toprule
    \multirow{2}{*}{\textbf{Models}}
      & \multicolumn{4}{c}{\textbf{Scenario pass@1 (\%) $\uparrow$}}
      & \multirow{2}{*}{\textbf{Avg. pass@1 (\%) $\uparrow$ }} 
      & \multirow{2}{*}{\textbf{MPR (s) $\downarrow$}} \\
    \cmidrule(lr){2-5}
    & \textbf{kyverno}
    & \textbf{k8s-opa}
    & \textbf{rhel-opa}
    & \textbf{kyverno-upadate} \\
    \midrule
    \textbf{granite-3.1-8B-instruct} &
    $7.84 \pm 3.84$ &
    $0.00 \pm 0.00$ &
    $0.00 \pm 0.00$ &
    $1.59 \pm 1.58$ &
    $1.71 \pm 0.76$ &
    $197.03 \pm 2.52$ \\
    \textbf{mixtral-8x7B-instruct} &
    $7.35 \pm 3.19$ &
    $1.43 \pm 1.42$ &
    $0.00 \pm 0.00$ &
    $1.29 \pm 4.34$ &
    $3.94 \pm 1.03$ &
    $120.63 \pm 3.77$ \\
    \textbf{llama-3.1-8B-instruct} &
    $8.57 \pm 3.37$ &
    $0.00 \pm 0.00$ &
    $0.00 \pm 0.00$ &
    $7.46 \pm 3.23$ &
    $3.59 \pm 1.07$ &
    $121.49 \pm 3.00$ \\
    \textbf{llama-3.3-70B-instruct} &
    $18.46 \pm 4.94$ &
    $0.00 \pm 0.00$ &
    $1.43 \pm 2.88$ &
    $8.06 \pm 3.50$ &
    $9.32 \pm 1.67$ &
    $189.61 \pm 2.71$ \\
    \textbf{mistral-large-2} &
    $6.56 \pm 3.20$ &
    $22.73 \pm 5.32$ &
    $7.23 \pm 2.88$ &
    $10.45 \pm 3.77$ &
    $11.55 \pm 1.95$ &
    $167.98 \pm 3.42$ \\
    \textbf{llama-3.1-405B-instruct} &
    $16.22 \pm 4.32$ &
    $20.83 \pm 4.86$ &
    $8.75 \pm 3.26$ &
    $3.17 \pm 2.22$ &
    $12.46 \pm 1.98$ &
    $178.89 \pm 3.37$ \\
    \textbf{gpt-4o-mini} &
    $16.18 \pm 4.54$ &
    $43.10 \pm 6.99$ &
    $30.38 \pm 5.43$ &
    $9.43 \pm 4.08$ &
    $25.19 \pm 2.80$ &
    $102.40 \pm 3.70$ \\
    \textbf{gpt-4o} &
    $40.28 \pm 5.99$ &
    $39.34 \pm 6.55$ &
    $7.61 \pm 2.81$ &
    $17.74 \pm 4.92$ &
    $24.74 \pm 2.64$ &
    $101.29 \pm 3.81$ \\
    \bottomrule
  \end{tabular}
  \begin{tablenotes}
    \footnotesize
    \item ``pass@1'' values are in percent. pass@1 is calculated as defined in Codex~\cite{chen2021evaluating}\\
    \item ``MPR'' mean processing time\\
    \item \textbf{kyverno} = New K8s CIS-benchmarks on Kyverno, 
          \textbf{k8s-opa} = New K8s CIS-benchmarks on OPA, 
          \textbf{rhel-opa} = New RHEL9 CIS-benchmarks on Ansible-OPA,
          \textbf{kyverno-update} = Update K8s CIS-benchmarks on Kyverno.
  \end{tablenotes}
\end{threeparttable}
\end{table*}

%% file: appx/usecases/finops/main.tex
\section{Financial Operations}
\label{appx:finops}

\subsection{Background}
\label{ss:finops-background}

FinOps (Finance + Operations) is an operational framework and cultural practice which maximizes the business value of cloud and creates shared financial accountability. One of the primary objective is to enable timely data-driven decision making  by fostering collaboration between engineering, finance, and business teams.  FinOps comprises of three iterative phases - Inform (Visibility \& Allocation), Optimize (Rates \& Usage), and Operate (Continuous Improvement \& Usage). 
Importantly, success in FinOps hinges on making iterative changes using real-time insights with data gathered from Application Performance Management (APM), Application Resource Management (ARM), and Finance (Cloud Cost Management)~\cite{cloud_finops_2nd}.

Contrary to common belief, FinOps is about maximizing business value and not just about reducing operating costs.
The 2024 State of FinOps Report ~\cite{finops_data, finops_insights_2024} gathered data from 1,245  respondents with an average annual cloud spend per company of \$44M, with some companies reporting up to \$1B+ in annual cloud spend, total cloud spend \$55B. This study showed a shift from previous years, with reducing waste and managing commitments becoming new leading key priorities across the board. 
Management of cloud discount programs (such as Savings Plans and Reserved Instances), and accurate forecasting of spend remained high on the list. 
 Other areas such as increased Automation, AI costs, and sustainability registered growing interest.
\subsubsection{Key terms}
FinOps is performed by working iteratively on the Framework Capabilities through three phases \cite{finopsbenchkpis} namely, Inform, Optimize and Operate, which are described below:
\begin{enumerate}[labelwidth=*, labelindent=0pt, leftmargin=*, label=(\arabic*),itemsep=0mm]
\item The Inform phase involves identifying data sources for cloud cost, usage and efficiency data. This data is then used for budgeting, allocation, forecasting, analysis and reporting. 
\item The Optimize phase identifies opportunities to improve cloud efficiency using the data and capabilities developed in the Inform Phase.
\item The Operate phase implements operationalizes FinOps using the data and capabilities developed in the Inform and Optimize phase. 
\end{enumerate}
Some of the common KPIs in the FinOps that are amenable to optimization (full set is listed in https://www.finops.org/wg/finops-kpis/) include
\begin{enumerate}[labelwidth=*, labelindent=0pt, leftmargin=*, label=(\arabic*),itemsep=0mm]
    \item Percentage Resource Utilization - This is the amount of resources utilized as a percentage of the the total capacity allocated.
    \item Total Unpredicted Variance of Spend - Measures the unpredicted variance of cost associated with CSP Cloud usage recorded over a given period of time.
    \item Auto-scaling Efficiency Rate -  Maximum capacity cost of running workload to meet workload demand / Cost of running workload with auto-scaling to meet same workload demand.
    \item Forecast Accuracy Rate (Usage) - Compares forecasted vs. actual cloud usage (vCPUs, Memory, etc) over a specific period (e.g., day, month, quarter).
    \item Forecast Accuracy Rate (Spend)- This metric compares forecasted vs. actual cloud spend over a specific period (e.g., day, month, quarter).
    \item Percent of Compute Spend Covered by Commitment Discounts - Measures the percentage of compute cost (excluding Spot) covered by commitment discount for a specific time period.
    \item Percentage of Commitment Discount Waste - The percentage of commitments not applied to on-demand spend.
    \item Percent of Unused Resources - Measure of unused cloud resources, e.g., unattached/orphaned storage volumes, load balancers, EIPS, Network gateways, snapshots.
    \item Percentage of Unallocated Shared CSP Cloud Cost - This measurement refers to expenses that cannot be directly attributed to a specific project, team, or department within an organization.
    \item Percentage Variance of Budgeted vs. Forecasted CSP Cloud Spend - Measures the difference between budgeted costs and the forecasted costs for using CSP cloud services
    \item Effective Savings Rate Percentage - Actual Spend with Discounts / Equivalent Spend at On Demand Rate
    \item Percentage of CSP Cloud Costs that are Tagging Policy Compliant - Total Costs Associated with Tagging Policy Compliant CSP Cloud Resources During a Period of Time / Total CSP Cloud Costs During a Period of Time. 
    \item Percent Storage on Frequent Access Tier - Number of GB in Standard (or “frequently accessed” tiers vs. total GBs stored)
    \item Percentage of Carbon Associated with Untagged CSP Cloud Resources
\end{enumerate}
There are several analogous KPIs related to carbon footprints instead of dollar costs. 
In the current version of the bench our scenarios have used an alert based on variance in spend. The remaining KPIs offer a rich basis for formulating many additional scenarios.
\subsection{Motivating Example and FinOps Tasks}
\label{ss:finops-tasks}
\Cref{tab:optimization_use_case_demand} shows an exemplar budget overrun incident with the steps for diagnosis and resolution. 
In this incident, FinOps practitioners were notified of an ~\textemdash unusual cost increase (>20\% more than last week's average)~\textemdash alert by OpenCost.
The increase is mainly caused by the increase in replica counts as a result of an observed load spike.
The autoscaler increased the number of replicas to serve the demand as expected.
However, budget thresholds are not updated which causes false alerts. 
Agent finds out that the application is healthy and recommends updating budget alerts based on the new load level. 
Similarly, Table~\ref{tab:optimization_use_case_autoscaler} demonstrates a sample scenario where a budget overrun alert has been generated due to increased replicas. 
However, in this scenario application services are scaled up significantly despite low utilization in containers. 
Agent finds out autoscaler scaled up the application at low utilization thresholds and analyze the autoscaler configuration. 
It detects low thresholds for scale up policy and recommends updating autoscaler accordingly.

\begin{table*}[ht!]
    \small
    \centering
    \begin{threeparttable}
        \caption{Optimization Use Case: Increased Cost Alert - Increasing Demand}
        \label{tab:optimization_use_case_demand}
        \begin{tabular}{m{0.3\textwidth}m{0.65\textwidth}}
            \toprule
            \textbf{Optimization Use Case} & \textbf{Details} \\
            \midrule
            Triggering Alert & Cost increase alert for an application “Foo”.  \\
            Summary & Cost alert was seen on application “Foo”. The increase was 20\% higher than the expected budget. Investigations reveal that the application is healthy and cost increase is caused by load increase. Client budget is flexible and thus cost alert is updated accordingly to increase the threshold. Long term fix requires additional check in CI/CD pipeline and possible automation of budget adjustments.   \\
            Time to detection & 7 days (Current practices to observe utilization metrics for cost analysis \\
            Time to diagnose & 60 minutes \\
            Time to mitigate & 15 minutes \\
            Event Type & An alert is generated to show there is more than 20\% increase in expected cost.\\
            Cost Overrun & 20\% increase in cost\\
            Diagnosis Steps & 
            \begin{itemize}[left=0pt, topsep=0pt, partopsep=0pt, itemsep=0pt, parsep=0pt]
                \item Checked infrastructure size changes. e.g. Increase in replica counts in Kubernetes cluster.
                \item Found replica counts are greater than historical average.
                \item Checked whether there is a legit increase in application load.
                \item Found there is a stable increase in the application load and the cost increase is acceptable.
                \item Checked budget constraints of the application.
                \item Found application budget is flexible for scaling up.
            \end{itemize} \\
            Resolution Plan & \begin{itemize}[left=0pt, topsep=0pt, partopsep=0pt, itemsep=0pt, parsep=0pt]
                \item Calculated the new cost alert thresholds.
                \item  Update cost alert budget thresholds to accommodate new stable load level to prevent false alerts.
            \end{itemize} \\ 
            Long Term Improvements & Created playbooks to ensure such handling such adjustments are automated. \\
            \bottomrule
        \end{tabular}
    \end{threeparttable}
\end{table*}

\begin{table*}[ht!]
    \small
    \centering
    \begin{threeparttable}
        \caption{Optimization Use Case: Increased Cost Alert - Faulty Auto-scaler}
        \label{tab:optimization_use_case_autoscaler}
        \begin{tabular}{m{0.3\textwidth}m{0.65\textwidth}}
            \toprule
            \textbf{Optimization Use Case} & \textbf{Details} \\
            \midrule
            Triggering Alert & Cost increase alert for an application “Foo”.  \\
            Summary & Cost alert was seen on application “Foo”. The increase was 20\% higher than the expected budget. Investigations reveal that auto scaler was misconfigured. SREs manually updated the auto scaling configuration by changing the scale up policy. Long term fix requires additional check in CI/CD pipeline.  \\ \\
            Time to detection & 7 days (Current practices to observe utilization metrics for cost analysis  \\
            Time to diagnose & 60 minutes \\
            Time to mitigate & 15 minutes \\
            Event Type & An alert is generated to show there is more than 20\% increase in expected cost.\\
            Cost Overrun & 20\% increase in cost\\ \\
            Diagnosis Steps & 
            \begin{itemize}[left=0pt, topsep=0pt, partopsep=0pt, itemsep=0pt, parsep=0pt]
                \item Checked infrastructure size anomalies, e.g., increase in replica counts in Kubernetes cluster.
                \item Found replica counts are greater than historical average.
                \item Checked whether there is a legit increase in application load.
                \item Found an increase in load and pending containers.
                \item  Checked utilization of containers
                \item Found low utilization
                \item Checked autoscaler for scaling policy
                \item Found low threshold for scale up rules
            \end{itemize} \\
            Resolution Plan & \begin{itemize}[left=0pt, topsep=0pt, partopsep=0pt, itemsep=0pt, parsep=0pt]
                \item Configure auto scaler to update scaling policies.
                \item Manually delete extra replicas.
            \end{itemize} \\ 
            Long Term Improvements & Created playbooks to ensure such misconfigurations do not happen for future deployments. \\
            \bottomrule
        \end{tabular}
    \end{threeparttable}
\end{table*}

\subsection{\bench Architecture for Constructing FinOps Task Scenarios}
\label{ss:finops-bench}
We have extensively leveraged the set up that we established for the SRE scenarios. We employ OpenCost to monitor costs and raise an alert when the predefined budget and efficiency thresholds are crossed. 
OpenCost is an opensource tool which distributes the cost of a virtual machine/node to the Kubernetes deployments running on it based on the allocated resources. It selects higher of utilization or request numbers of each container and distributes the cost using a load distribution policy. 
In our experiments, we have forced a custom pricing model which mainly includes hourly CPU cost rate for a single core and memory cost rate per 1 GB memory. Other pricing components such as networking cost and spot instance pricing are not the main scope of our evaluated scenarios. Thus, we have included negligible costs for these components. 
We mimic real world scenarios as explained in SRE scenarios by including cost variation alerts.

\subsection{Evaluation}
We have used the same evaluation framework used in SRE-bench by including OpenCost and cost alerts in the scenarios. 

\subsubsection{Evaluation Metrics}
We evaluate each LLM-based agents on two primary tasks: (i) diagnosis and (ii) mitigation.

\textbf{Diagnosis.}
The agent is evaluated on diagnosis by its ability to provide accurate \textit{root cause} of budget alerts.
Diagnosis includes the analysis steps that the agent follows to identify the root cause of the budget alerts.
Agent performance in diagnosis is measured using pass@1 scores that indicate the accuracy of the root cause analysis provided to the SREs.

\textbf{Mitigation.}
Mitigation involves recommending resolution steps for incidents to clear the alerts and optimizing cost and efficiency of the application.  
We evaluate the agents with the success rate using pass@1 score and using proximity scores to analyze the performance of agents to achieve optimal cost and efficiency metrics defined in Section~\ref{s:finopsMetricDef}.
Both diagnosis and mitigation evaluations can be expanded in the future to automatically measure the performance of agents using LLM-as-a-judge similar to SRE-bench evaluations.

\subsubsection{Metric definitions}
\label{s:finopsMetricDef}
\textbf{pass@1.}
We use the same metric defined in Section~\ref{sss:eval-metrics} for both diagnosis and mitigation steps recommended by the agents. 

\textbf{Proximity metrics.} To analyze the performance of agents in achieving optimal cost and efficiency, we defined proximity scores for hourly CPU cost, hourly memory cost, workload CPU efficiency, and workload memory efficiency. Proximity scores indicates the performance by calculating the proportional absolute difference between observed and optimal values for the measured cost or efficiency metric. We subtract the proportional value from 1 such that having proximity score of 1 indicates achieving the optimal performance. Proximity scores are calculates as follows:

$$\text{proximity}\_i = 1 - \frac{| \text{observed}\_i - \text{optimal} |}{\text{optimal}}$$
where $i$ represents the experiment trial, $observed$ is the retrieved value for the measured metric, and $optimal$ is the value given in the ground truth for the same field.

\textbf{Hourly CPU cost:} The cost of allocating a single CPU core for an hour. We used custom pricing policy through opencost to ensure the ground truth values are not affected by changes in pricing between different cloud providers and on-premise deployments. \\

\textbf{Hourly Memory cost:} The cost of allocating 1 GB of memory. It also uses custom pricing policy. \\

\textbf{Workload CPU efficiency:} We take the average CPU efficiency of all containers of the application. We calculate CPU efficiency by dividing the CPU utilization of a container by the request amount. For instance, if the CPU request is 100m in the deployment configuration but the container uses 50m, the efficiency is calculated as 50.\\

\textbf{Workload memory efficiency:} It is calculated in the same way with CPU efficiency by dividing the utilization by requested memory amount. \\

We take the arithmetic mean of all runs for each metric and calculates the standard deviation. We present our results in Table~\ref{tab:finopsagent-eval}.

\subsection{Example Trajectories}
\label{ss:finops-trajectories}

\begin{figure}[hbtp]
    \centering
    \includegraphics[width=0.8\linewidth]{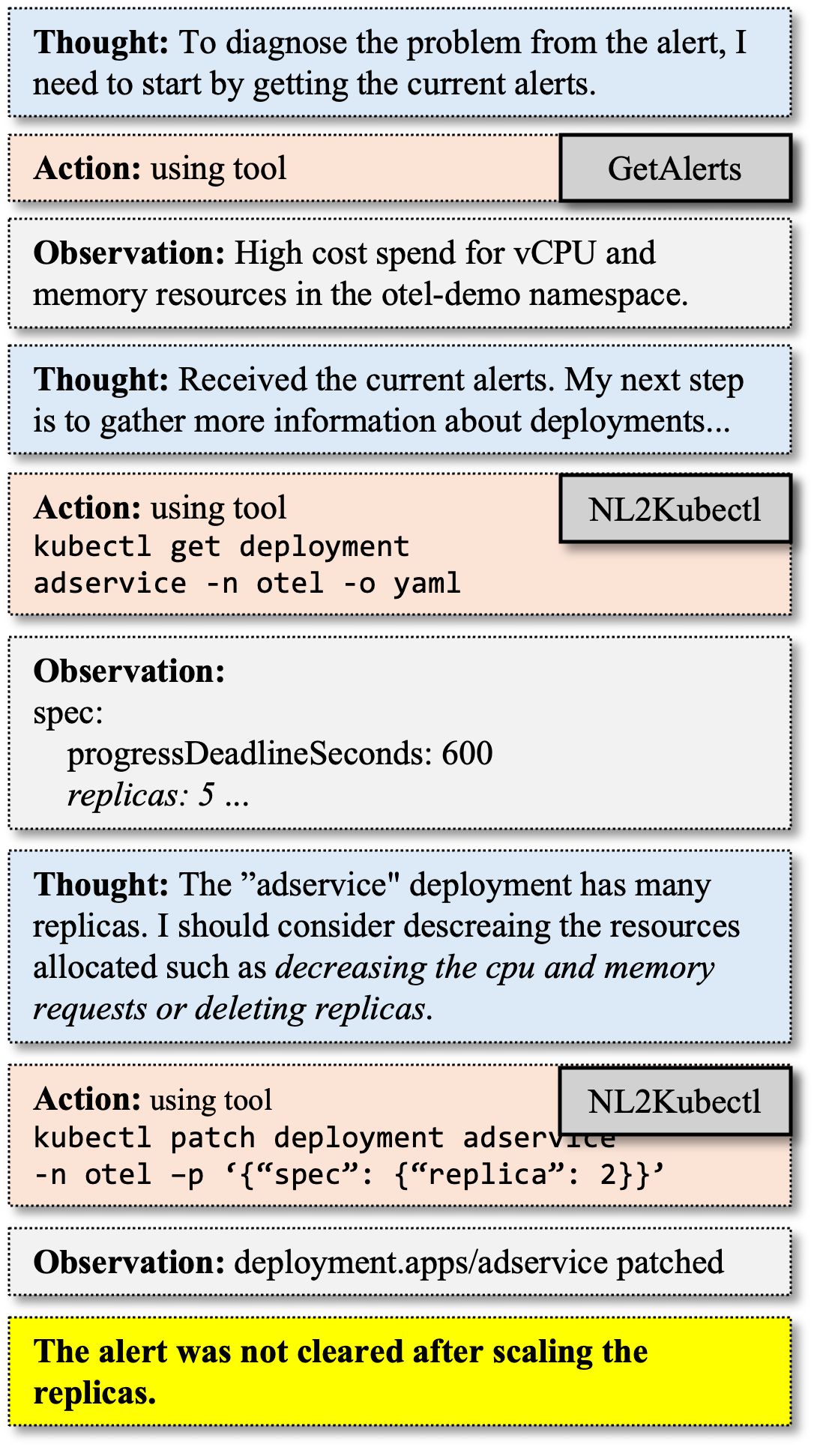}
    \caption{Sample Trajectory of unusual cost variation use case.} 
    \label{fig:finops-bad-trajectory}
\end{figure}

FinOps issues are highly connected to used infrastructure and deployment configurations and policies for Kubernetes deployments. 
Agents commonly need to use the same tools as SRE-agent to understand the status of underlying infrastructure and deployment configuration of the application to identify the root cause of a FinOps concern. 
Similarly, to investigate budget variations accurately, they need analyzing load and utilization patterns to avoid diverging to irrelevant resolution recommendation. 
Figure~\ref{fig:finops-bad-trajectory} shows a trajectory for Scenario 37 where agent starts retrieving the alerts and observing CPU hourly cost has exceeded the budget threshold.
It continues checking deployment details using NL2Kubectl tool and retrieve replica counts for each deployment. 
Due to high number of replicas in adservice, it considers scaling down the adservice deployment. 
Missing utilization checks and not analyzing the deployment details caused agent to an inaccurate diagnosis and led to scaling down recommendation despite the high load for the application. 
In a correct trajectory, an agent would analyze the utilization and decides the high load in the system and would recommend changing the budget alert thresholds if total budget allocation allows or analyze the deployments which could be scaled down without hurting the application performance. 

